\documentclass[sigconf]{acmart}

\AtBeginDocument{%
  \providecommand\BibTeX{{%
    \normalfont B\kern-0.5em{\scshape i\kern-0.25em b}\kern-0.8em\TeX}}}

\setcopyright{acmcopyright}
\copyrightyear{2023}
\acmYear{2023}
\acmDOI{XXXXXXX.XXXXXXX}

\acmConference[CHI '24]{Make sure to enter the correct
  conference title from your rights confirmation emai}{May 11--16,
  2024}{Honolulu, Hawaii}

\acmBooktitle{Hawaii' 24: ACM (Association of Computing Machinery) CHI conference on Human Factors in Computing Systems ,
 May 11--16,2024, Honolulu, Hawaii} 
\acmPrice{15.00}
\acmISBN{978-1-4503-XXXX-X/18/06}

\usepackage{multicol}
\usepackage{multirow}
\usepackage{array}
\usepackage{cleveref}
\usepackage{subcaption}
\definecolor{ForestGreen}{RGB}{34,139,34}
\usepackage{xcolor} % Consolidated xcolor package line
\usepackage{enumitem}

\begin{document}

\title{\textit{Art or Artifice?} Large Language Models and the False Promise of Creativity}

\author{Tuhin Chakrabarty}
\email{tuhin.chakr@cs.columbia.edu}
\affiliation{%
  \institution{Columbia University}
  \country{USA}
}

\author{Philippe Laban}
\affiliation{%
  \institution{Salesforce AI Research}
  \country{USA}
}

\author{Divyansh Agarwal}
\affiliation{%
  \institution{Salesforce AI Research}
  \country{USA}
}

\author{Smaranda Muresan}
\email{smara@cs.columbia.edu}
\affiliation{%
  \institution{Columbia University}
  \country{USA}
}

\author{Chien-Sheng Wu}
\affiliation{%
  \institution{Salesforce AI Research}
  \country{USA}
}

\renewcommand{\shortauthors}{Chakrabarty, et al.}

\begin{abstract}
Researchers have argued that large language models (LLMs) exhibit high-quality writing capabilities from blogs to stories. However, evaluating objectively the creativity of a piece of writing is challenging. Inspired by the Torrance Test of Creative Thinking (TTCT) \cite{torrance1966torrance}, which measures \textit{creativity as a process}, we use the Consensual Assessment Technique \cite{amabile1982social} and propose \textit{Torrance Test of Creative Writing} (TTCW) to evaluate \textit{creativity as product}. TTCW consists of 14 binary tests organized into the original dimensions of Fluency, Flexibility, Originality, and Elaboration. We recruit 10 creative writers and implement a human assessment of 48 stories written either by professional authors or LLMs using TTCW. Our analysis shows that LLM-generated stories pass 3-10X less TTCW tests than stories written by professionals. In addition, we explore the use of LLMs as assessors to automate the TTCW evaluation, revealing that none of the LLMs positively correlate with the expert assessments.\end{abstract}



% Author Keywords
% Print the classficiation codes

%     \small
%     \centering
%   \includegraphics[width=0.35\textwidth]{figures/sample-ai.png}
%   \caption{\href{https://www.newyorker.com/culture/cultural-comment/the-computers-are-getting-better-at-writing}{The Computers Are Getting Better at Writing (The NewYorker)\sm{We don't need this figure, I would remove, you got several comments on this. definitely remove for CHI submission}}}

\received{20 February 2007}
\received[revised]{12 March 2009}
\received[accepted]{5 June 2009}

\maketitle

% \input{macros}
%SM As I am making changes to intro, I am creating a new file to keep the old file just in case
\section{Introduction}
\label{sec:intro}

Recent work has explored the potential of using large language models (LLMs) as assistants in creative writing tasks, from stories to screenplays 
\cite{mirowski2023cowriting,yang2022re3,lee2022coauthor,ippolito2022creative,chen2023ambient}. An important aspect of this research is to investigate LLMs’ capabilities in terms of their ability to both \textit{generate} creative content and \textit{assess} whether a piece of writing is creative, with the ultimate goal of informing interaction design. However, evaluating objectively the creativity of a piece of writing is challenging. 

\begin{figure*}
\centering
\includegraphics[width=\textwidth]{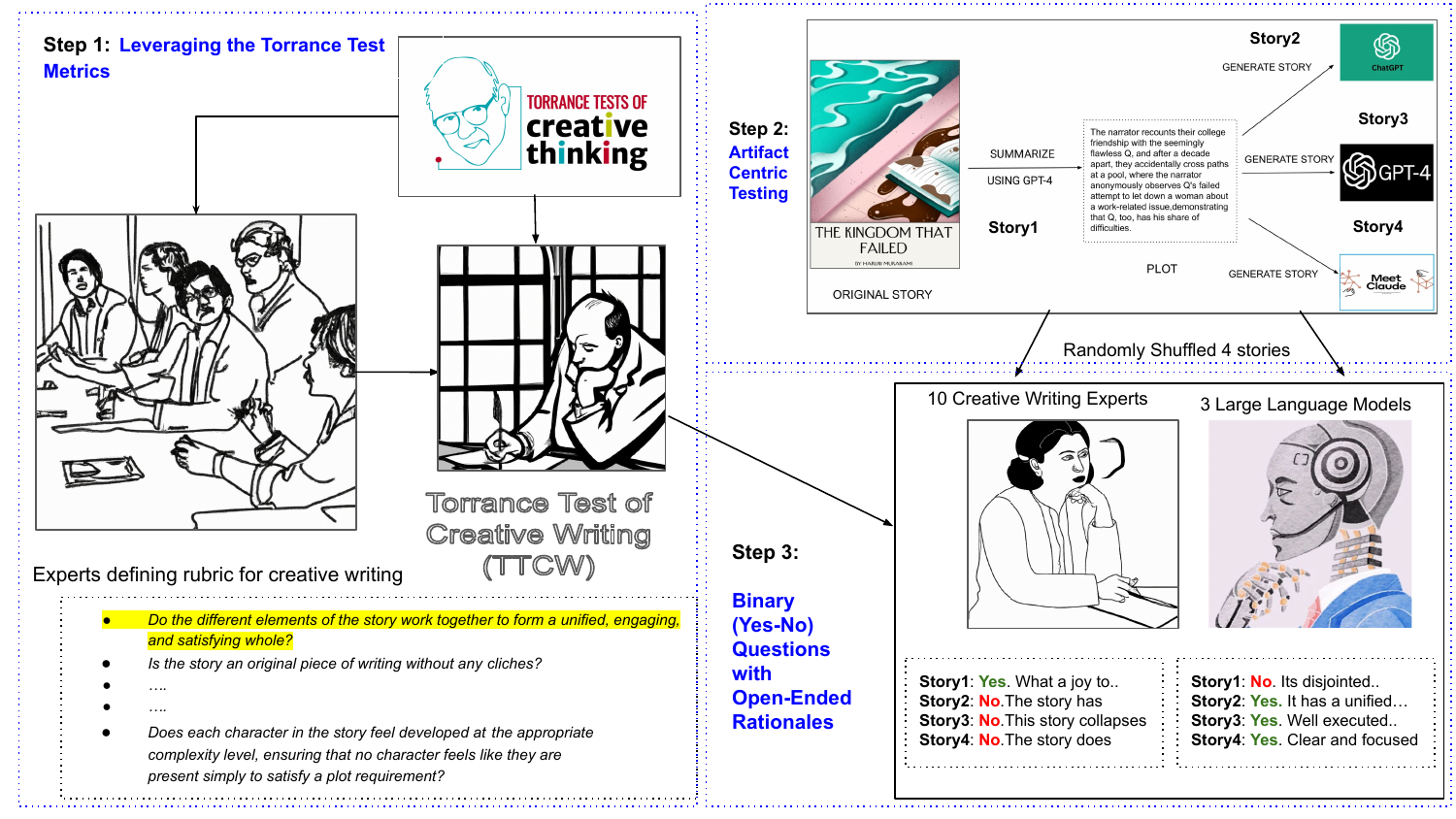}
\caption{\label{pipeline}Pipeline showing the construction of TTCW and evaluation of short stories using the TTCW framework where Step 1) shows how experts leverage the process-oriented Torrance Test of Creative Thinking to create 14 tests for evaluating creativity in short stories as a product. Step 2) demonstrates artifact-centric testing where 4 stories based on a single plot are used as a product of creativity evaluation Step3) shows an evaluation of stories using the TTCW framework by both expert humans and LLMs where they each provide Yes/No answers to individual tests followed by natural language rationales justifying their decision.}
\end{figure*}

We propose a protocol to evaluate \textit{creativity as product} grounded in a widely accepted protocol that evaluates \textit{creativity as process} --- the Torrance Tests of Creative Thinking (TTCT) \cite{torrance1966torrance}. Based on Guilford's work on divergent thinking \cite{guilford1967nature}, TTCT measures creativity as a process by testing participants’ abilities in dealing with unusual uses of objects, specific situations, or impossibilities. TTCT is centered around evaluating four dimensions of creativity: \textit{fluency} (the sheer volume of meaningful ideas produced in reaction to a given stimulus), \textit{flexibility} (the diversity of categories within the responses), \textit{originality} (the uniqueness or novelty of answers) and \textit{elaboration} (the depth or granularity of details within the responses). 
While the direct application of TTCT might not be possible across diverse creative domains \cite{amabile1982social,baer2009assessing}, its four fundamental dimensions have proven adaptable \cite{trisnayanti2019development,mcintyre2003individual,10.1145/3313831.3376495}. Thus, based on TTCT and using the Consensual Assessment Technique (CAT) \cite{amabile1982social} we design the \textit{Torrance Tests for Creative Writing (TTCW)} to evaluate \textit{creativity as product} (Section \ref{design} Design Principle 1 and 2; Figure \ref{pipeline} Step 1). %SM-rr I added Step 1
The Consensual Assessment Technique states that the most valid assessment of the creativity of an idea or creation in any field is the collective judgment of experts in that field. Thus, to design TTCW, we asked 8 creative writing experts in a formative study (Section~\ref{sec:approach}) to propose creativity measures for the evaluation of short fictional stories aligned with the four Torrance dimensions. This resulted in 14 binary tests organized across the four original Torrance dimensions of \textit{fluency, flexibility, originality} and \textit{elaboration} (Section~\ref{CreativityTest}). 

To empirically validate TTCW as an evaluation protocol for creativity as product (fictional short stories), we build a benchmark consisting of 48 short stories: 12 stories written by professionals, and 36 by three top performing LLMs (ChatGPT \cite{ChatGPT}, GPT4 \cite{OpenAI2023GPT4TR} and Claude 1.3 \cite{Claude}) with 1400 words on average per story (Section~\ref{sec:data}; Figure \ref{pipeline} Step 2). %SM-rr I added Step 2
We recruit a new set of 10 experts, again adhering to CAT, to administer the 14 tests of TTCW on each story, collecting 3 evaluations per story. In this \textit{TTCW implementation with experts as assessors} (Figure \ref{pipeline} Step 3) %SM-rr I added Step 3
we aim to answer three research questions: 
\begin{enumerate}[label=RQ\arabic*:,leftmargin=3em]
     \item \textit{Is the TTCW-based creative evaluation consistent and reproducible? In other words, is there agreement among expert annotators when they perform tests on the same stories?} Based on a total of 2,000+ expert-administered tests, our findings reveal that experts reach moderate agreement on average (Fleiss Kappa 0.41) across the 14 TTCW tests, and reach strong agreement when considering the aggregated tests (Pearson correlation $0.69$). This confirms the validity of TTCW for the evaluation of creativity in fictional short stories.\\
     
    \item \textit{Are the human-written stories more likely to pass individual TTCW than LLM-generated stories? If so, which tests demonstrate the most significant gaps?} The analysis of the conducted tests reveals that the 12 expert-written stories pass an average of 84.7\% of the tests, confirming that although an expert story must not pass all tests to be deemed creative, experienced writers typically produce artifacts that pass the majority of the tests. In comparison, LLM-generated stories pass many fewer tests on average, from 9\% for ChatGPT-generated stories, up to 30\% for Claude-generated stories. In other words, LLM-generated stories are three to ten times less likely to pass individual TTCW tests compared to expert-written stories, revealing a wide gap in the evaluated creativity of LLM-generated content.\\ 
    
    \item \textit{Which LLMs perform better in TTCW evaluation, and are there specializations observed, with some different LLMs performing better on different Torrance dimensions?} Besides the general gap, the granularity of the TTCW reveals that individual LLMs differ in abilities, with GPT4 more likely to pass tests associated with Originality, and Claude V1.3 more likely to pass tests in Fluency, Flexibility and Elaboration.
\end{enumerate}

Prior work \cite{ippolito2022creative} has argued that given the nature of LLMs where they regurgitate text seen during pre-training, they are often unable to directly generate a truly original and creative piece of writing (Also see Section \ref{expertvsai}). However, they might be utilized in providing feedback to authors during their writing process \cite{chakrabarty2023creativity}. Thus, we perform a study of \textit{TTCW implementation with LLMs as assessors} (Section ~\ref{sec:llmeval}; and Figure \ref{pipeline} Step 3) %SM-rr I added Step 3
to understand whether LLMs can be used to assess creative writing. 
We expanded each test into a detailed prompt, and measured whether LLM assessments correlate with collected expert judgments. Our analysis reveals that for the most part, LLMs are not capable of administering the TTCW tests, as the three LLMs we experiment with achieve correlations with experts that are close to zero.

In summary, our work makes the following contributions:
\begin{itemize}
    \item We adapt the Torrance Test for Creative Thinking (TTCT), a protocol for evaluating creativity as a process, and align it for the evaluation of creativity as a product particularly focusing on short stories. Using the Consensual Assessment Technique, we design 14 tests called the \textit{Torrance Test for Creative Writing (TTCW)} based on the four original Torrance dimensions of fluency, flexibility, originality, and elaboration,
    \item We experimentally validate the TTCW through an assessment of 48 stories involving 10 participants with expertise in creative writing, finding that they reach moderate agreement when administering individual tests, and strong agreement when evaluating all tests in aggregate.
    \item We study the abilities of LLMs to generate stories that pass/fail the TTCW tests and their ability to reliably assess the creativity of stories following the TTCW framework through correlation with human judgments. Our findings show that LLM-generated stories are three to ten times less likely to pass TTCW tests compared to expert-written stories, as well as the fact that current state-of-the-art LLMs are not yet capable of reproducing expert assessments when administering TTCW tests. To enable future research in this fast-evolving domain, we release the large-scale annotation of 2,000+ TTCW assessments, each accompanied with a natural language expert explanation. 
    \item Finally, we discuss how creative writing experts can distinguish between AI vs. human written stories and how future work can use our evaluation framework for building rich interactive writing support tools.
\end{itemize}

\section{Related Work}
\subsection{Creativity Evaluation}
In prior work, \citet{baer2014creativity} argued how divergent thinking remains the most frequently used indicator of creativity in both creativity research and educational practice, and divergent thinking theory has a strong hold on everyday conceptions of what it means to be creative. Along the same lines, \citet{kaufman2008essentials} further discusses and evaluates common creativity measures such as divergent thinking tests, consensual technique, peer/teacher assessment, and self-assessment. \citet{silvia2008assessing} examined the reliability and validity of different subjective scoring methods for divergent thinking tests and introduced a new Top 2 scoring method that involves participants selecting their most creative responses, and demonstrates that this method yields reliable scores with a small number of raters. \citet{plucker2010assessment} discuss key issues and methods in creativity assessment including reliability, validity, bias, and use of assessments. \citet{beaty2021automating} explored the use of automated scoring via semantic distance, using natural language processing in assessing the quality of ideas in creativity research, demonstrating its strong predictive abilities for human creativity and novelty ratings across various tasks, thus addressing the labor cost and subjectivity issues in traditional human-rating methods.Like many of the prior works our research centers around grounding creativity evaluation through divergent thinking. In particular, we rely on the Torrance Tests of Creative Thinking (TTCT) as a foundation for measuring creativity.
\subsection{Evaluating Creative Writing}
Rubrics are one of the major tools for assessing writing which incorporate a set of prominent characteristics relevant to a specific type of discourse \cite{weigle2002assessing}. \citet{vaezi2019development} developed a rubric for the evaluation of fiction writing fiction through nine elements, namely narrative voice, characterization, story, setting, mood and atmosphere, language and writing mechanics, dialogue, plot, and image. To ensure its validity, they further recruited a number of distinguished creative writing professors to review this assessment tool and comment on its appropriateness for measuring the intended construct. \citet{biggs1982psychological} proposed to evaluate creative writing through the lens of the structural complexity of the product by utilizing the SOLO (Structure of the Observed Learning Outcome) taxonomy which buckets creative writing into incoherent (prestructural), linear(unistructural), conventional (multistructural), integrated (relational) and metaphoric (extended abstract). In prior work \citet{rodriguez2008problem} argued that narrative theory is key in teaching and grading creative writing. They emphasized how breaking down narrative elements such as plot, discourse-time, character, setting, narration, and filter delineates the tools authors use to effectively write fiction. In the creativity evaluation space \citet{baer2009assessing,amabile1982social} proposed \textit{The Consensual Assessment Technique} as a method of assessing creative performance on a real-world task such as writing a poem or a story. Unlike tests from the Torrance Tests of Creative Thinking (TTCT), the CAT does not rely on any specific criteria or test scores. The method proposes that the most valid assessment of the creativity of an idea or creation in any field is the collective judgment of experts in that field. Unlike prior work, we aim to align the evaluation of creativity as a process to the evaluation of creativity as a product with feedback from experts building upon prior theoretical works such as the TTCT and the CAT. Unlike prior work from \citet{vaezi2019development} where the rubric was created by surveying existing literature our work involved experts for creating the rubric from scratch without biasing their opinion or thought-process. Finally our rubric was validated by 5X more experts and across 3X more stories compared to that of \citet{vaezi2019development}.
\subsection{Expert Evaluation of Language Model Generations}
A cogent argument posits that the engagement with artistic prose isn't confined solely to specialists, but non-experts also can be competent in assessing imaginative prowess. Inheriting from research standards for large-scale natural language processing tasks, the majority of studies assessing the quality of generations from LLMs evaluate model performance by collecting data from crowd workers \cite{10.1145/3424636.3426903,rashkin-etal-2020-plotmachines,goldfarb-tarrant-etal-2020-content,roemmele-gordon-2018-linguistic,10.1609/aaai.v33i01.33017378}. Nevertheless, it is critical to acknowledge that during narrative evaluation trials involving both teachers in English and participants from Amazon Mechanical Turk, the study by \citet{karpinska-etal-2021-perils} exhibited that AMT contributors, even when shortlisted via rigorous eligibility parameters (unlike teachers), struggle to discriminate between model generated text and human-crafted references. \citet{clark-etal-2021-thats} run a similar study assessing non-expert's ability to distinguish between human and machine-authored text (GPT2 and GPT3) in three domains (stories, news articles, and recipes) and find that, without training, evaluators distinguished between GPT3 and human-authored text at random chance level. \citet{mirowski2023cowriting} emphasized why crowd workers are not a good fit for evaluating AI-generated screenplays and instead engage 15 experts—theatre and film industry professionals—who have both experiences in using AI writing tools and who have worked in TV, film, or theatre in one of these capacities: writer, actor, director, or producer for evaluating LLM generated screenplays. Finally, a recent study by \citet{veselovsky2023artificial} highlighted the fact that approximately 33-46\% of crowd workers on such platforms currently utilize large language models (LLMs) to complete any assigned task. Taking account of the above factors, for our work on evaluating short stories, we recruit creative writing experts ranging from professors to literary agents as well as MFA Fiction candidates.
\section{Design Considerations}\label{design}

One of the main contributions of our work is the collection of 14 tests, referred to as the \textit{Torrance Test for Creative Writing (TTCW)}, to evaluate creativity in short fictional stories. These tests were formulated through collaboration with domain experts which we detail in Section \ref{sec:approach}, but in this Section, we first present the design principles that shaped the methodology and provide the desiderata for the tests, which are then empirically validated.

\paragraph{Design Principle 1: Leveraging the Torrance Test Metrics.} The 14 tests we propose are grounded in the Torrance Test for Creative Thinking (TTCT) \cite{torrance1966torrance}, which has been a cornerstone in the evaluation of creativity.TTCT provides measures to understand the creative process through tasks that encompass unusual uses of objects, scenarios, and out-of-the-box problems. TTCT is centered around evaluating four dimensions of creativity

\begin{itemize}
\item \textbf{Fluency.} The sheer volume of meaningful ideas produced in reaction to a given stimulus.
\item \textbf{Flexibility.} The diversity of categories within the responses. 
\item \textbf{Originality.} The uniqueness or novelty of answers.
\item \textbf{Elaboration.} The depth or granularity of details within the responses.
\end{itemize}

While the direct applications of TTCT might exhibit limitations in terms of applicability across diverse creative domains \cite{amabile1982social,baer2009assessing}, its fundamental dimensions have proven adaptable. Researchers have repurposed these dimensions effectively in diverse sectors like science education \cite{trisnayanti2019development}, content strategies in marketing \cite{mcintyre2003individual}, and even in human-computer interaction, particularly interface design \cite{10.1145/3313831.3376495}.In Section \ref{sec:approach}, we show how using creative writing experts we design domain-specific tests grounded in each TTCT dimension. Feedback from these specialists further underscores the pertinence of these dimensions in assessing creative writing.

\paragraph{Design Principle 2: Artifact-centric Testing.}\footnote{In the context of creative writing we will refer to the "product" as artifact}  A key consideration when designing tests to evaluate creativity is whether to center the evaluation on the \textit{cognitive process} that leads to creativity or whether to evaluate the final artifact, which is a byproduct of the process \cite{mayers2007re}.
Much prior work -- including the TTCT -- takes a design-centric approach, as it includes richer observation of the evaluated individual, which might not be captured in the final artifact. However, process-oriented evaluation is limited in several ways. First, process-oriented evaluation is inherently limited by the quality of the observation of the individual's process. For instance, internal thoughts of the individual and other unrecorded activities can bias and lower the quality of the evaluation. Second, prior work has argued that neatly separating a process from an artifact is challenging, as the two are ``tightly integrated'' \cite{mayers2007re}, with ``the creative process leaving traces within the artifact'' \cite{murray2012craft}. Finally, observing the process is not always possible, particularly when evaluating the creativity of a preexisting artifact (e.g., a short story written years ago), or evaluating black-box agents such as LLMs, whose process cannot be observed in an interpretable way. We, therefore, follow prior work \citet{vaezi2019development,rodriguez2008problem} and design our creative writing evaluation to be artifact-centric.

\paragraph{Design Principle 3: Binary (Yes-No) Questions with Open-Ended Rationales.} In accordance with findings from prior work \cite{doi:10.1177/001316447103100307} which showed that reliability and validity are independent of the number of scale points
used for Likert-type items, we stick to a binary scale. The evaluation within each Torrance dimension follows a similar procedure. Each dimension is associated with multiple binary questions, that represent individual tests. Each binary question is formulated as having a Yes/No answer such that an artifact receiving a ``Yes'' answer to a question corresponds to the artifact passing the test. Additionally, the Yes/No answer should be accompanied by a free-text rationale written by the evaluator which justifies the chosen binary label, with a length expectation of at least 1-3 sentences. For each test, the combination of a structured binary assessment and an open-ended rationale are complementary. The binary assessment can be used for quantitative assessment, such as measuring agreement amongst evaluators, or comparative evaluation of a story collection (such as the one we perform in Section ~\ref{ref:humanresult}), whereas the rationale can be used for qualitative assessment, such as understanding concrete reasons for the passing or failing of a test, such as the analysis we perform in Section \ref{sec:analysis} centering around the most common themes that lead to the passing or failing of a given test.
\paragraph{Design Principle 4: Additive Nature of Tests.} Each question is intended to be independent of other questions (i.e., no question is a prerequisite to another question), but the creative assessment of a given artifact requires completing all the TTCW. 
The final assessment of a given artifact is the number of tests passed by the artifact, with the general expectation that passing more tests is directly proportional to the creativity of the artifact. In other words, the passing or failing of any single test cannot be interpreted as a final assessment of the creativity of an artifact, but rather the number of tests passed can paint a more complete picture of the creativity of the artifact. Analysis in Section ~\ref{ref:humanresult} confirms that our experts achieve on average moderate agreement on individual tests, but strong agreement when considering all tests in aggregate, confirming empirically the additive nature of the tests. In Section \ref{sec:approach}, we conduct a formative study with experts to formulate the fourteen TTCW that satisfy our design principles, which we then use in Section ~\ref{ref:humanresult} to run a evaluation of short stories using TTCW with experts as assessors. 

\section{Formative Study: Formulating the Torrance Tests For Creative Writing} \label{sec:approach}
We restricted the involvement of participants in our formative study to only those possessing either a structured educational background in creative writing (for instance, a Master of Fine Arts in Creative Writing), traditionally published authors \footnote{We do not recruit self-published authors}, or lecturers/professors instructing Fiction Writing at the university level. We specifically chose this filtering criterion to restrict our selection pool to experts in the field thereby aligning with the \textit{Consensual Assessment Technique}. Our recruitment resulted in participants who have published novels with leading publishing houses, students enrolled in top MFA programs in the United States, University professors teaching Fiction Writing, and screenwriters from prime-time networks. Participants were recruited through \textit{UserInterviews}~\footnote{\url{https://www.userinterviews.com}}, a professional freelancing website, and were paid \$70 for taking part in the hour-long survey. Table \ref{surveyprof} shows the background of the recruited participants. Our recruited participants span across different age groups, gender and professional expertise.

The formative study was structured in three parts. Initially, over video conferencing, experts were briefed on the study's primary objective, which aimed at devising actionable metrics for assessing creative writing, emphasizing fiction. They were also introduced to the Torrance Tests and the four specific dimensions encompassed by it. In the subsequent phase, participants were emailed the URL to a web app where they were asked to input their measures in text. They were instructed to allocate a 20-minute window to articulate up to five distinct measures corresponding to each of the Torrance dimensions first. This task was conducted without a sample fiction to maintain abstraction. The final phase involved presenting the participants with a sample fiction piece \footnote{\url{https://www.newyorker.com/books/flash-fiction/the-mirror}} on the same web app for evaluation, retrieved from The New Yorker. Participants were encouraged to use this as a tangible example to refine and augment their initial measures, ensuring they were grounded and practical. As an outcome of our study, we received 126 measures from the participants across the four Torrance dimensions. \footnote{The research was conducted at an institution that does not have an IRB approval process in place, but an Ethical Practices team reviewed the work and study protocols. We did not collect or share any PII during data collection, and participants could choose not to complete the survey and still receive a payment.}

\begin{table}[!ht]
\centering
\small
\begin{tabular}{|l|l|l|l|}
\hline
ID & Profession  & Gender & Age                \\ \hline
W1 & Professor of Creative Writing & Female & 45 \\ \hline
W2 & Professor of Creative Writing & Female & 56 \\ \hline
W3 & Lecturer in Creative Writing & Male & 40  \\ \hline
W4 & MFA Fiction Student & Male & 35        \\ \hline
W5 & MFA Fiction Student & Male & 31      \\ \hline
W6 & MFA Fiction Student & Female & 48         \\ \hline
W7 & Young Adult Fiction Writer  & Non-Binary & 39                      \\ \hline
W8 & ScreenWriter & Non-Binary & 34                  \\ \hline
\end{tabular}
\vspace{2ex}
\caption{\label{surveyprof}Background of Participants recruited for collecting judgments about Creativity across the dimensions of Torrance Test}
\end{table}

\begin{table}[!ht]
\centering
\small
\begin{tabular}{|l|l|l|}
\hline
Dimension & Expert Measure & Participants                  \\ \hline
\multirow{7}{*}{Fluency} & Narrative Pacing & W4,W6,W7 \\ \cline{2-3}
&\begin{tabular}[c]{@{}l@{}}Understandability\\ \& Coherence\end{tabular}  & W3,W7 \\ \cline{2-3}
&\begin{tabular}[c]{@{}l@{}}Language Proficiency\\ \& Literary Devices\end{tabular}  & W4,W2  \\ \cline{2-3}
&Narrative Ending & W3           \\ \cline{2-3}
&Scene vs Summary & W2,W5,W8          \\ \hline
\multirow{3}{*}{Flexibility} &Structural Flexibility & W1,W3,W8           \\ \cline{2-3}
& Perspective \& Voice Flexibility & W3,W6                        \\ \cline{2-3}
& Emotional Flexibility & W3                  \\ \hline
\multirow{5}{*}{Originality} & Originality in Theme/Content & W3,W6                 \\ \cline{2-3}
&Originality in Thought & \begin{tabular}[c]{@{}l@{}}W1,W2,W3,\\W5,W7\end{tabular}                  \\ \cline{2-3}
&Originality in Form & W2,W3,W4                  \\ \hline
\multirow{5}{*}{Elaboration} & World Building \& Setting & W2,W6                  \\ \cline{2-3}
&Rhetorical Complexity & W3,W4                  \\ \cline{2-3}
&Character Development & \begin{tabular}[c]{@{}l@{}}W2,W3,W4\\W5,W7,W8\end{tabular}                  \\ \hline
\end{tabular}
\vspace{2ex}
\caption{\label{testsource}Tests proposed and mapped by experts for evaluating story writing across TTCW dimensions}
\end{table}

\subsection{From Measures to Actionable Tests}
The measures derived from the participants exhibited a considerable degree of semantic congruence. For example, W2 proposed one way to measure Originality in Creative Writing as \textit{``Shows an innovative use of form/structure.''} while W4 proposed a similar measure \textit{Formal or stylistic novelty}.
W6 proposed one way to measure Elaboration depending on whether \textit{``The story has developed 3D characters.''} while W3 proposed an exactly similar measure but phrased it as \textit{``Does the piece make a flat character complex?''}. To consolidate these measures and develop a framework of the underlying tests for measuring creativity, we use a general inductive approach for analyzing qualitative data \cite{thomas2006general}. Following this method, three authors independently read all of the measures and assigned each measure an initial potential low-level group. Then, through repeated discussion, we reduced category overlap and created shared low-level groups associated. Finally, these low-level groups were collected into high-level groups, and a name was proposed for each group that encapsulates a generalized representation of the measures within the group. During the later meetings, an American Novelist and Creative Writing Professor were present to give further insights into the data.

In total, the tagging process yielded 14 distinct groups, 5 in the Fluency dimension, and 3 in Flexibility, Originality, and Elaboration. Table~\ref{testsource} presents the name we assigned to each group and the study participants that proposed a measure tagged within this group. For every group, we had a list of expert-suggested questions speaking about the same artifact. To choose a representative measure for every group, we selected the most well-articulated measure (in terms of word count). If the measure was already suggested as a question, we keep it intact; otherwise, we used the GPT4 model to convert a measure to a Yes/No question. For example, \textit{Originality suggests that the piece isn’t cliche $\rightarrow$ Is the story an original piece of writing without any
cliches?}
Next, we list each TTCW and provide the necessary background to contextualize the test. It should be noted that our tests are additive in nature. Failing a particular test should not be interpreted as the fact that the story doesn’t present any creative elements. Instead, it should be considered, by counting the number of distinct tests passed by a given story to obtain a more calibrated understanding of the creativity in a given piece.

\subsection{The Torrance Test for Creative Writing} \label{CreativityTest}

\paragraph{\textbf{{Fluency}}}: Compared with both reading and speaking fluency, writing fluency has always been traditionally harder to define \cite{abdel2013we}. Our 5 measures across this dimension each look at individual aspects of creative writing. 

\subsubsection{\textbf{{\color{blue} Narrative Pacing (TTCW Fluency1)}: Does the manipulation of time in terms of compression or stretching feel appropriate and balanced?}}This measure refers to the manipulation of time in storytelling for dramatic effect. Essentially, it is about controlling the perceived speed and rhythm at which a story unfolds. A skilled writer can manipulate the relationship between these two to affect the pacing of the narrative, either speeding it up (compression) or slowing it down (stretching). This technique plays a crucial role in shaping the reader's experience and engagement with the story. To assess narrative pacing, W4 suggested looking  ``\textit{Compression/stretching of time (story time vs. real world time)}'' while W6 and W7 advised to ``\textit{control the speed at which a story unfolds
\footnote{\url{https://www.writingclasses.com/toolbox/articles/stretching-and-shrinking-time}}.}''

\subsubsection{\textbf{{\color{blue}Scene vs Exposition (TTCW Fluency2)}: Does the story display awareness and insight into the balance between scene and summary/exposition?}} A `Scene' is a moment in the story that is dramatized in real-time, often featuring character interaction, dialogue, and action, while `Exposition', on the other hand, involves summarizing events or providing information like character history, setting details, or prior events. The right balance between scene and summary/exposition can vary depending on the story, but in general, it's essential for maintaining a good pace, keeping the reader engaged, and delivering necessary information \citet{burroway2019writing} 
\footnote{\url{https://creativenonfiction.org/syllabus/scene-summary/}}.
W2 strongly felt that fluent writing needs to ``\textit{display awareness and insight into the balance between scene and summary/exposition in the story.}'' while W5 and W8 emphasized the need for ``\textit{Enough dialogue to compensate for backstory.}''

\subsubsection{\textbf{{\color{blue}Language Proficiency \& Literary Devices (TTCW Fluency3)}: Does the story make sophisticated use of idiom or metaphor or literary allusion?}} Eminent novelist Milan Kundera said ``\textit{Metaphors are not to be trifled with. A single metaphor can give birth to love.}''. Sophisticated use of literary allusion or figurative language such as metaphor/idioms often add depth, interest, and nuanced meaning to any creative writing. It allows for a richer reading experience, where the literal events are imbued with deeper symbolic or thematic significance. W4 and W2 both emphasized the presence of ``\textit{Sophisticated use of idiom, metaphor, and literary allusion} or \textit{Surprising, skilled, and complex use of metaphor/simile/allusion} as a way to measure Fluency in creative writing.
\subsubsection{\textbf{{\color{blue}Narrative Ending (TTCW Fluency4)}: Does the end of the story feel natural and earned, as opposed to
arbitrary or abrupt?}} In her New Yorker essay ``On Bad Endings'' \cite{BadEndings} Accocela writes ``Another possibility is that the author just gets tired. I review a lot of books, many of them non-fiction. Again and again, the last chapters are hasty and dull. `I’ve worked hard enough,' the author seems to be saying. `My advance wasn’t much. I already have an idea for my next book. Get me out of here'.''
If the writer ends the piece simply because they are ``tired of writing'', the conclusion might feel abrupt, disjointed, or unfulfilling to the reader. This is one of the important factors of creative writing fluency. A strong ending offers a sense of closure, ties up the central conflicts or questions of the story, and generally leaves the reader feeling that the narrative journey was worthwhile and complete. For this measure of Fluency, W3 asked  ``\textit{Does the writer know how to end the piece not because they're tired of writing, but because they have come to the moment the entire piece has been leading us towards?}''
\subsubsection{\textbf{{\color{blue}Understandability \& Coherence (TTCW Fluency5)}: Do the different elements of the story work together to form a unified, engaging, and satisfying whole?}} Narrative coherence is the degree to which a story makes sense  \footnote{https://en.wikipedia.org/wiki/Narrative\_paradigm}. A well-crafted story usually follows a logical path, where the events in the beginning set up the middle, which then logically leads to the end. Every scene, character action, and piece of dialogue should serve the story and propel it forward. Well-written stories have an underlying unity that binds the elements together. W3 strongly advocated for this measure by saying ``\textit{Does the piece hold together? In other words, does the beginning lead through the middle to the end in a way that feels deliberate and intentional? This is the difference between several pages of writing and a PIECE of writing. Great writing errs on the side of unity over disorder.}'' while W7 suggested the importance of this measure through ``\textit{a logical flow.''}

\paragraph{\textbf{Flexibility}} Flexibility is often referred to as the ability to look at something from a different angle or point of view. In the context of creative writing, our participants agreed on 3 distinct measures of Flexibility.
\subsubsection{\textbf{{\color{blue}Perspective \& Voice Flexibility (TTCW Flexibility1)}: Does the story provide diverse perspectives, and if there are unlikeable characters, are their perspectives presented convincingly and accurately?}} An \textit{omniscient} narrator is the all-knowing voice in a story that can convincingly and accurately depict a wide range of character viewpoints, including those of characters who may be morally ambiguous, difficult, or otherwise unappealing. As stated in \citet{friedman1955point} an omniscient narrator enhances a sense of reliability or truth within literary works since readers are given deeper insights into many characters. The multiple viewpoints feel more objective because readers have access to multiple interpretations of events and can thus decide how they feel about each character’s perspective. W3 wanted ``\textit{a writer to be able to inhabit various perspectives, even unlikable ones.}'' while W6 suggested ``\textit{Flexibility of voice. Can the author inhabit the consciousness of different characters and not just likable ones?}"
\subsubsection{\textbf{{\color{blue}Emotional Flexibility (TTCW Flexibility2)} : Does the story achieve a good balance between interiority and exteriority, in a way that feels emotionally flexible?}}
Emotional flexibility is asking whether the piece of writing effectively balances action and introspection, and if it portrays a broad and realistic spectrum of emotions as corroborated by W3. \textit{Exteriority} refers to the observable actions, behaviors, or dialogue of a character, and the physical or visible aspects of the setting, plot, and conflicts.\textit{Interiority}, on the other hand, pertains to the inner life of a character — their thoughts, feelings, memories, and subjective experiences. A balance between these two aspects is crucial in creating well-rounded characters and compelling narratives. As stated in \citet{campe2014rethinking} if a story is too heavy on exteriority, it may feel shallow or lack emotional depth. If it leans too much on interiority, it could become overly introspective and potentially lose the momentum of the plot.
\subsubsection{\textbf{{\color{blue} Structural Flexibility(TTCW Flexibility3)}: Does the story contain turns that are both surprising and appropriate?}} A good piece of creative writing often has plot twists, character developments, or thematic revelations that surprise the reader, subverting their expectations in a thrilling way. However, despite the surprises and twists, the turns in the story must also make sense within the established context of the story's universe, its characters, and its themes. It shouldn't feel like the writer has broken the rules they've set up, or made a character behave inconsistently without reason, simply for the sake of shock value. In order for any writing to be structurally flexible W3 wanted to ensure that ``\textit{the writer capable of making turns in the work that are both surprising and appropriate} while W1 required the presence of ``\textit{New twist in the story that are believable} to measure structural flexibility
\paragraph{\textbf{{Originality}}} Creative writing requires originality, or the ability to generate unique ideas \cite{ward1999creative}.
Our participants suggested three unique ways in which they look for originality in creative writing.

\subsubsection{\textbf{{\color{blue}Originality in Theme and Content (TTCW Originality1)}: Will an average reader of this story obtain a unique and original idea from reading it?}} In his book ``Literature and the Brain'' well-known literary critic and scholar Norman Holland discusses how stories stimulate the mind and impact readers \citet{holland2009literature}. A good story that offers a deeper understanding of human nature, cultural insights, unique viewpoints, or even the exploration of new ideas and themes has a lasting impact on its reader and society. In ``Poetic Justice'', prominent philosophers Martha Nussbaum explores how the literary imagination is an essential ingredient of public discourse and a democratic society \citet{nussbaum1997poetic}. As such originality in theme and content is an important measure of creative writing. In the words of W6 originality in theme meant  ``\textit{New brilliant ideas about the future and humanity (mostly in speculative fiction). Does the writing have an original message?} while W3 mentioned ``\textit{Do I feel I am learning something new from the piece? What is the purpose of putting it into the world?}''
\subsubsection{\textbf{{\color{blue} Originality in Thought (TTCW Originality2)}: Is the story an original piece of writing without any cliches?}} A cliche is an idea, expression, character, or plot that has been overused to the point of losing its original meaning or impact \citet{fountain2012cliches}. They often become predictable and uninteresting for the reader. In his book \citet{clark2008writing} eminent American writer, editor, and writing coach: Roy Peter Clark advised writers to strictly avoid cliches because they often indicate a lack of original thought or laziness in language use. Originality suggests that the piece isn't cliche. Several experts agreed on this measure with W3 saying ``\textit{Originality suggests that the piece isn't cliche.} while W4 required ``\textit{Plot that is surprising rather than cliche}''. W6 emphasized originality in thought through ``\textit{writing that is not cliche, not recycled tropes, and does not contain stereotyped one-dimensional characters}
\subsubsection{\textbf{{\color{blue}TTCW Originality in Form \& Structure (Originality3)} : Does the story show originality in its form?}} In his book \citet{boardman1992narrative} Frederic Jameson highlighted the complexities of postmodern literature, where the blurring of genres and innovation in form was a key characteristic \citet{jameson1991postmodernism}. Originality in form has also been accomplished by the unconventional use of format, genre, or narrative structure or arc. For instance, the Pulitzer-winning book \textit{The Color Purple} by Alice Walker is told through a series of letters written by the protagonist. Neil Gaiman's \textit{American Gods} on the other hand combines elements of fantasy, mystery, and mythic fiction in unexpected ways. \textit{The Sound and the Fury} by William Faulkner deviates from the traditional plot structure by presenting a narrative that unfolds through the stream of consciousness of different characters. The goal of originality in form or structure is often to provide a fresh reader experience, challenge conventional reading expectations, or to create a deeper or more complex exploration of the story's themes. This was corroborated by W2 and W5 who wanted to see ``\textit{formal or stylistic novelty}. W3 also wanted to see if \textit{A piece demonstrates original ways to work in sometimes tired forms (such as the short story)}
\paragraph{\textbf{{{Elaboration}}}}
Elaboration in creative writing is the process of adding details and information to a story to make it more interesting and engaging. This can be done by describing the setting, characters, and action in more detail, or by adding dialogue, thoughts, and feelings to the story.

\subsubsection{\textbf{{\color{blue} World Building and Setting (TTCW Elaboration1)}: Does the writer make the fictional world believable at the sensory level?}} American poet and memoirist Mark Doty discuss the importance of creating a vivid, immersive reality at the sensory level through the use of detailed, evocative description \cite{doty2014art}. An effective writer often uses sensory details to paint a detailed picture of the story's environment, making it feel tangible and real to the reader. This level of detail contributes to the believability of the world, even if it is a completely fictional or fantastical setting. It helps the reader to suspend disbelief and become more deeply invested in the narrative. W2 suggested measuring elaboration through ``\textit{Use of setting to inform the action and atmosphere of the story without succumbing to pathetic fallacy}''. W6 further confirmed the same by saying ``\textit{There is a setting. The short story is rooted in a time and a place. If it's science fiction or fantasy there will be some world building.}''. W5 further mentioned ``\textit{the presence of a self-consistent world}'' as a measure of elaboration while W3 asked ``\textit{How does the writer make the fictional world believable at the sensory level?}
\subsubsection{\textbf{{\color{blue}Character Development (TTCW Elaboration2)}: Does each character in the story feel developed at the appropriate complexity level, ensuring that no character feels like they are present simply to satisfy a plot requirement?}} A 'flat character' is typically a minor character who is not thoroughly developed or who does not undergo significant change or growth throughout the story. They often embody or represent a single trait or idea, and they're only used to advance the plot or highlight certain qualities in other characters. A 'complex character' on the other hand also known as a round character, has depth in feelings and passions, has a variety of traits of a real human being, and evolves over time. \citet{forster1927aspects,fishelov1990types,currie1990nature} highlights that any creative piece of fiction or non-fiction tends to be more engaging to the reader when authors can take a character who initially appears to be one-dimensional or stereotypical (flat) and add depth to them, as it mirrors the complexity of real people. Multiple experts emphasized on the importance of character development. For instance, W6 thought ``\textit{The story should have developed 3D characters.} while W2 and W8 asked for ``\textit{Complex character development that avoids stereotype, generalization, trope, etc}''. W3 wanted to see ``\textit{Does the piece make a flat character complex?}
\subsubsection{\textbf{{\color{blue}Rhetorical Complexity (TTCW Elaboration3)}: Does the story operate at multiple ``levels'' of meaning (surface and subtext)?}}In Ernest Hemingway's short story ``Hills Like White Elephants,'' the couple's conversation about seemingly unrelated topics implies a much deeper and more serious discussion about abortion. Their actual dialogue never directly addresses this issue, but it's heavily suggested through what's left unsaid — the subtext. Effective writing often operates on both surface and subtext levels. The surface text keeps the reader engaged with the plot and characters, while the subtext provides depth, complexity, and additional layers of interpretation, contributing to a richer and more rewarding reading experience \citet{kochis2007baxter,phelan1996narrative}. W3 asked for rhetorical complexity by saying ``\textit{Is there a sense of the piece being complex in such a way that it needed to be written as a whole, not simply paraphrased or summed up?} while W4 mentioned that ``\textit{Text should operate at multiple levels of meaning (surface and subtext)}.''

\section{TTCW Implementation with Experts as Assessors} \label{sec:data}
In this Section, we detail our implementation of creative writing evaluation using the TTCW framework derived previously. We first carefully select the 48 short stories included in the evaluation, then go over the 2.5-hour study design protocol that our 10 experiment participants followed, and finally analyze the results based on the 2,000+ individual tests administered by the experts. We leverage the collected annotations to answer the following research questions:
\begin{enumerate}[label=RQ\arabic*:,leftmargin=2.5em]
    \item \textit{Are the human-written New Yorker stories more likely to pass individual TTCW than LLM-generated stories? If so, which tests demonstrate the most significant gaps?}
    \item \textit{Is the TTCW-based creative evaluation consistent and reproducible? In other words, is there agreement among expert annotators when they perform tests for similar stories?}
    \item \textit{Which LLMs perform better in TTCW evaluation, and are there specializations observed, with some different LLMs performing better on different Torrance dimensions?}
\end{enumerate}

\subsection{Data Selection}

\begin{figure*}
\centering
\includegraphics[width=\textwidth]{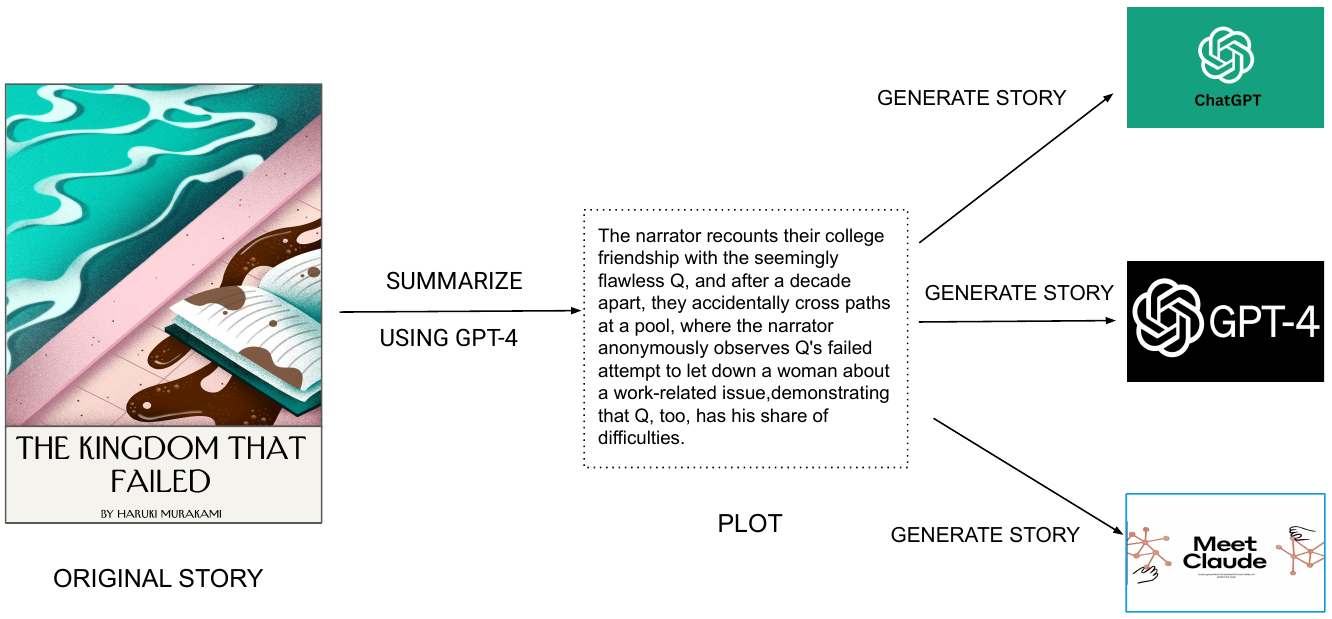}
\caption{\label{datapipeline}Pipeline showing how our test set is created for evaluation. For each human-written original New Yorker story, we generate 3 stories from one LLM each, based on the plot of the original story. The plot is a single-sentence summary of the original story automatically generated by GPT-4 and verified by humans.}
\end{figure*}

Prior studies in the creativity of model-generated text have shown that technologies such as LLMs are capable of generative long and coherent stories \cite{yang2022doc,yang2022re3}, as well as act as assist and collaborate with creative writers \cite{yuan2022wordcraft,ippolito2022creative,mirowski2023cowriting}. For practical considerations, these studies limit their evaluation solely to model-generated content, typically only claiming relative improvements from one system to another, and do not establish whether a gap remains between high-quality human-written stories and model-generated text. Our research employs a rigorous evaluation protocol that juxtaposes human-written short stories against those constructed by LLMs to discern the potential gap, if any, in their creative quality. The dataset selection process is visually summarized in Figure~\ref{datapipeline}. We first collect 12 short stories from The New Yorker collection of short stories \footnote{https://www.newyorker.com/tag/short-stories}. Our selection criterion involved choosing short stories with diverse authors and plots. Spanning from August 13, 2020, to May 8, 2023, these stories (ranging between 1000 to 2,400 words as delineated in Figure \ref{lengthdist} in the Appendix) include compositions from acclaimed authors such as Haruki Murakami to Nobel laureate Annie Ernaux. The titles and a one-sentence plot summary (generated by GPT-4 and verified by humans) of included New Yorker short stories are listed in Table ~\ref{teststories} in Appendix.

We prompt three top-performing LLMs: GPT3.5, GPT4, and Claude V1.3 to generate a story of similar length to each New Yorker story, based on the one-sentence plot summary. We note that the choice to condition model-generated stories on the plot of the New Yorker story is an important design consideration of our evaluation. First, LLMs are known to have limited ability in devising original plotlines, as highlighted in previous research \cite{ippolito2022creative}. Second, it creates groups of stories that center around a common plot, allowing to dissociating evaluation of a story's form (i.e., creative writing), from the evaluation of plot-line creativity. 

Although LLMs were prompted to generate stories of a given length, initial experimentation by the authors of the paper revealed that the LLMs typically generate stories that can be 20-50\% more concise than intended. Length is a known confounding factor in text generation evaluation, for example with work showing that evaluators systematically prefer longer summaries as they tend to be more informative \cite{stiennon2020learning}. To address this limitation, we employed an iterative mechanism that prompted the LLM to iteratively expand on its initial story until the divergence in word count between the AI-generated and its paired human-written story was less than 200 ( See Prompt in Table \ref{promptstory} in Appendix). In our processing, all LLMs were able to converge to the desired length in at most 20 iterations. The procedure yields a total of 12 story groups, each consisting of one New Yorker story, and three LLM-generated stories, all following a common plot-line and having very similar length, for a total of 48 short stories.
We also experimented with a few other choices in prompt design such as adding \textit{You are an expert of creative fiction writing} to the beginning of the prompt or demonstrating an example New Yorker story in the prompt but these did not lead to any appreciable difference in the quality of the stories based on preliminary evaluation.

In our experiments, the stories either originate from experts or LLMs, and we do not include stories written by non-experts or `amateurs'. Although this was a consideration in the initial phases of the project, collecting amateur-written stories turned out to be a challenging and expensive process. Given the way our experiments are designed, it would require recruiting 12 non-experts to write stories given a plot since each expert-written story comes from a distinct author. The definition of an `expert' or `amateur' creative writer is difficult in a field that has unclear professional delineations. Collecting expert-written stories is straightforward as we rely on publication venue as a signal but identifying amateur writers is in itself a challenging task and might entail certain selection biases. A story written by any individual otherwise deemed as `amateur' can still be of considerable high quality. Another alternative we considered is resorting to crowd-working platforms to recruit amateurs for writing stories but a recent study by \citet{veselovsky2023artificial} highlighted the fact that approximately 33-46\% of crowd workers on such platforms currently utilize large language models (LLMs) to complete any assigned task, which would negatively impact our experiments' reproducibility. Finally adding amateur-written stories would require increased budgets for evaluation. These difficulties govern our decision not to include amateur-written stories.
\subsection{Evaluation Protocol}

\subsubsection{Expanded Expert Measure and Prompt Design} \label{sec:prompting}

\begin{table*}[!ht]
\centering
\small
\begin{tabular}{|l|l|}
\hline
\begin{tabular}[c]{@{}l@{}}Expert \\ Measure\end{tabular}               & Does the writer make the fictional world believable at the sensory level?                                                                                                                                     \\ \hline
\begin{tabular}[c]{@{}l@{}}Expanded\\ Expert\\ Measure (M)\end{tabular} & \begin{tabular}[c]{@{}l@{}}Sensory details pertain to the five senses - sight, sound, touch, taste, and smell. An effective \\ writer can use these elements to paint a detailed picture of the story's environment, making\\ it feel tangible and real to the reader.\\ \\ For example, describing the specific colors and shapes in a scene, the sounds that fill a space,the \\ textures and temperatures that a character comes into contact with, the flavors of the food they \\ eat, or the scents that fill the air, can all contribute to creating a sensory-rich and believable world.\\ \\ By stimulating the reader's senses, the writer can make the reader feel as though they're \\ experiencing the events of the story firsthand.This level of detail contributes to the believability of\\ the world, even if it's a completely fictional or fantastical setting. It helps the reader to suspend\\ disbelief and become more deeply invested in the narrative.\end{tabular} \\ \hline
\begin{tabular}[c]{@{}l@{}}Human\\ Instruction\end{tabular}             & \begin{tabular}[c]{@{}l@{}}\{\{M\}\}\\ \\ Based on the story that you just read, answer the following question.\\ \textit{\color{blue}Does the writer make the fictional world believable at the sensory level?}\\ -Yes \\ -No \\\\ Reasoning : \end{tabular}                                                                       \\ \hline
\begin{tabular}[c]{@{}l@{}}LLM\\ Instruction\end{tabular}               & \begin{tabular}[c]{@{}l@{}}\{\{M\}\}\\ \\ Given the story above, list out the elements in the story that call to each of the\\ five senses. Then overall, give your reasoning about the question below and give\\ an answer to it between 'Yes' or 'No' only\\ \\ \textit{\color{blue} Q) Does the writer make the fictional world believable at the sensory level?}\end{tabular}                                                                                                                                                                                                                                 \\ \hline
\end{tabular}
\vspace{2ex}
\caption{\label{prompting1}Expert suggested question for World Building and setting (Row1) ; Expanded Expert Measure (Row2); Elucidated prompt designed for other expert humans
(Row3); Elucidated quantifiable prompt designed for Large Language Models that elicit Chain of Thought Reasoning(Row4) }
\vspace{-5ex}
\end{table*}

We want these tests described above to be understandable by both other creative writing experts or even LLMs, such that they can be used for evaluation purposes. An expert suggested questions for empirically evaluating creative writing might frequently elicit ambiguity in Large Language Models or even other creative writing experts. In order for LLMs or other experts to comprehend the suggested questions in Section
\ref{CreativityTest}, we attempt to expand them by adding more details.
Recent pre-trained LLMs (e.g., GPT-4 \cite{OpenAI2023GPT4TR} GPT3.5 \cite{ChatGPT}) can engage in fluent, multi-turn conversations out of the box, substantially lowering the data and programming-skill barriers to creating passable conversational user experiences. People can improve LLM outputs by prepending prompts—textual instructions and examples of their desired interactions—to LLM inputs. The prompts steer the model towards generating the desired outputs, raising the ceiling of what conversational UX is achievable for non-AI experts. To elucidate these questions we prompt GPT4 with the following instruction: \textit{What do creative experts mean when they say the following: \{\{expert question\}\}}. Once GPT4 gives a response 3 domain experts carefully verify the response and edit it where required. Table \ref{prompting1} (Row2) 
shows the human-verified GPT4 expanded expert measure in response to the input prompt. Table \ref{prompting1} (Row3; Human Instruction) shows the final instruction given to human experts during the evaluation of our stories that contains the expanded expert measure \{\{M\}\} in addition to the original yes/no question.More examples of Human Instructions for the remaining 13 tests are provided in Section \ref{allprompts} in the Appendix.

\begin{table}[!ht]
\centering
\small
\begin{tabular}{|l|l|l|l|}
\hline
ID & Profession  & Gender & Age                \\ \hline
E1 & Lecturer of Creative Writing & Male & 42 \\ \hline
E2 & Lecturer of Creative Writing & Male & 32 \\ \hline
E3 & Professor of Creative Writing & Male & 46  \\ \hline
E4 & Professor of Creative Writing & Female & 43 \\ \hline
E5 & Literary Agent & Male & 29 \\ \hline
E6 & Literary Agent & Female & 30  \\ \hline
E7 & Writer with an MFA in Fiction & Non-Binary &  25       \\ \hline
E8 & Writer with an MFA in Fiction & Male & 24      \\ \hline
E9 & Writer with an MFA in Fiction & Male & 28      \\ \hline
E10 & Writer with an MFA in Poetry  & Male & 30                      \\ \hline
\end{tabular}
\vspace{2ex}
\caption{\label{creativeexperts}Background of creative writing experts recruited for evaluating the stories from our test set}
\vspace{-3ex}
\end{table}

\begin{table*}[!ht]
\small
\centering
\begin{tabular}{|l|l|l|l|l|l|l|}
\hline
Dimension & Test & GPT3.5 & GPT4 & Claudev1.3 & NewYorker & Expert Agreement \\ \hline
\multirow{5}{*}{Fluency} & Understandability \& Coherence & 22.2 & 33.3 & 55.6 & \textbf{91.7} & 0.27 \\ \cline{2-7}
& Narrative Pacing & 8.3 & 52.8 & 61.1 & \textbf{94.4} & 0.39 \\ \cline{2-7}
& Scene vs Exposition & 8.3 & 50.0 & 58.3 & \textbf{91.7} & 0.27 \\ \cline{2-7}
& Literary Devices \& Language Proficiency & 5.6 & 36.1 & 13.9 & \textbf{88.9} & 0.37 \\ \cline{2-7}
& Narrative Ending & 8.3 & 19.4 & 33.3 & \textbf{91.7} & 0.48 \\ \hline\hline
\multirow{3}{*}{Flexibility} & Emotional Flexibility & 16.7 & 19.4 & 36.1 & \textbf{91.7} & 0.32 \\ \cline{2-7}
& Perspective \& Voice Flexibility & 8.3 & 16.7 & 19.4 & \textbf{72.2} & 0.44 \\ \cline{2-7}
& Structural Flexibility & 11.1 & 19.4 & 30.6 & \textbf{88.9} & 0.39 \\ \hline\hline
\multirow{3}{*}{Originality} & Originality in Form & 2.8 & 8.3 & 0.0 & \textbf{63.9} & 0.41 \\ \cline{2-7}
& Originality in Thought & 2.8 & 44.4 & 19.4 & \textbf{91.7} & 0.40 \\ \cline{2-7}
& Originality in Theme \& Content & 0 & 19.4 & 11.1 & \textbf{75.0} & 0.66 \\ \hline\hline
\multirow{3}{*}{Elaboration} & World Building \& Setting & 16.7 & 41.7 & 58.3 & \textbf{94.4} & 0.33 \\ \cline{2-7}
& Character Development & 8.3 & 16.7 & 16.7 & \textbf{61.1} & 0.31 \\ \cline{2-7}
& Rhetorical Complexity & 2.8 & 11.1 & 5.6 & \textbf{88.9} & 0.66 \\ \hline\hline
\multicolumn{2}{|c|}{Average} & 8.7 &27.9 & 30.0 & \textbf{84.7} & 0.41 \\ \hline
\end{tabular}
\vspace{2ex}
\caption{\label{absolutehumaneval} Average passing rate on individual TTCW, based on annotations of 10 creative writing experts across the 48 stories in our collection, authored by GPT3.5, GPT4, Claude, and expert human writers published in the New Yorker along with agreement measures (Fleiss Kappa) on individual test.}
\end{table*}

\subsubsection{Expert Evaluation Protocol}
\label{experteval}
We developed an evaluation protocol tailored specifically for experts in the domain of creative writing. The protocol, designed to be completed in approximately 2 to 2.5 hours, centered around a rigorous assessment of tuples of four distinct stories (one New Yorker story and the associated LLM-generated stories by the three LLMs) using the TTCW. The study was structured as follows:

\begin{enumerate}
    \item The four stories in a group were shuffled, and anonymized (i.e., the author of the story was not visible to the evaluator).
    \item The expert evaluator read the first story in its entirety and then administered the fourteen TTCW tests by assigning a Yes/No label and providing a justification for the label.
    \item Upon completing the evaluation of the first story in the group, the evaluator proceeded to read and evaluate the second, third, and fourth stories in the group respectively. The evaluators were also allowed to edit their responses at any point of time during the entire process.
    \item Once the evaluator had completed the TTCW evaluation of the four stories within a group, they were asked to rank all four stories in terms of subjective preference, and were asked to make an estimated guess of each story's origin: choosing from ``An experienced writer'', ``An amateur writer'', or ``Written by AI.'' The exact formulation of each question is given in Figure~\ref{relev} in the Appendix.
\end{enumerate}

In preliminary trials conducted by the authors of the paper, the entire task completion was observed to range from 2-2.5 hours. Consequently, an \$80 remuneration was determined to appropriately acknowledge the expertise of participants and encourage them to provide detailed justifications in their TTCW assessments. We note that participants were not provided details on the process used to create the groups of four stories, and were not told that each group consisted of one human-written story and 3 LLM-generated stories. Participants were explicitly instructed to avoid using search engines, which might reveal the origin of the New Yorker story. We also ensured beforehand that the participants were not familiar with any of the stories within a given group. Finally, participants were permitted to take breaks during the study but were encouraged to complete the entire task within a 24-hour window, so they would clearly remember each story when completing the final comparative task.

\subsection{Participant Recruitment}
\label{participant}

To test the robustness and validity of TTCW-based evaluation, we chose to recruit a new set of experts to conduct the evaluation and not re-hire the ones from our formative study that played a role in the creation of the tests. We posit that such a choice demonstrates the fact that the TTCW can be administered by any expert provided solely with the tests and their expanded explanations. We recruited 10 participants on the \textit{User Interviews} platform and listed their background in Table~\ref{creativeexperts}. Four of these participants are associated with the creative writing departments at leading American academic institutions, with considerable experience conducting undergraduate and graduate-level courses. Two participants function as literary agents at a top-tier, full-service US literary agency representing well-recognized authors, and four are professional writers with a Master of Fine Arts in Fiction or Poetry.

To experimentally analyze the reproducibility and validity of the TTCW, we randomly assigned each story group to 3 distinct experts. This allows us to study the agreement levels between experts on individual tests as well as in aggregate. After each task, an expert participant was given the option to receive another group of stories, and our participants completed on average 3.6 tasks over 3 weeks, for a total of 36 assessments (i.e., 3 for each of the twelve groups). Because each assessment contains four stories, and each story was evaluated using the 14 TTCW, we collected a total of 2,016 binary labels and expert-written justifications for these labels.
\subsection{Results} \label{ref:humanresult}
\subsubsection{Analyzing Pass Rates of Stories}
\begin{figure*}
    \centering
    \includegraphics[width=0.95\textwidth]{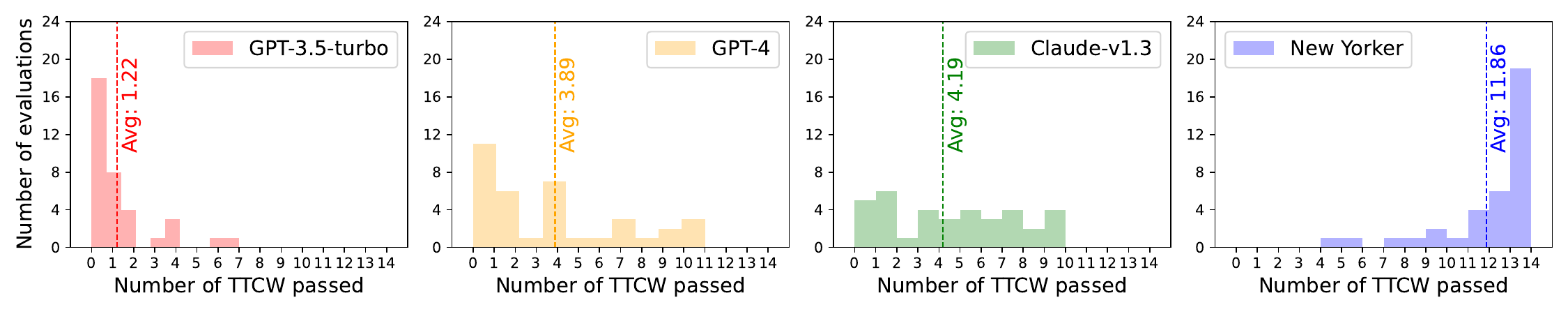}
    \caption{Distribution of aggregate TTCW results, in which only the number of tests passed is retained. }
    \label{fig:ttcw_aggregate_histogram}
\end{figure*}

\begin{figure*}[!ht]
    \subfloat[Likert plot showing ranking of stories within an individual group based on all expert preferences]{\frame{\includegraphics[width=0.44\textwidth]{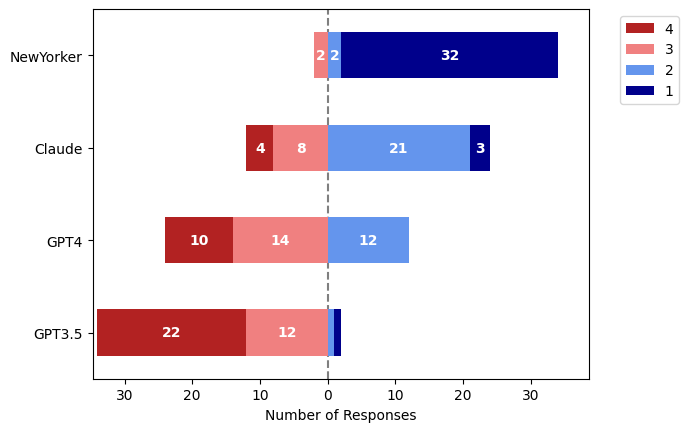}}} \quad
    \subfloat[Likert plot showing how often authors attributed the source of the stories correctly]{\frame{\includegraphics[width=0.53\textwidth]{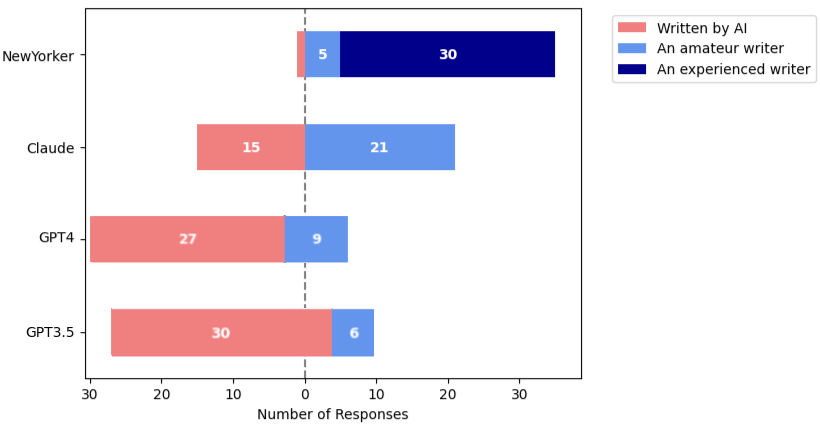}}} 
    \caption{\label{relativehumaneval1} \textbf{Relative Evaluation} Left figure showing ranking preference assigned to each story within a group. Right figure showing how creative experts attributed any given story from The NewYorker or 3 LLMs to one of the options between An experienced writer, An amateur writer, or An AI }
\end{figure*}

Table ~\ref{absolutehumaneval} summarizes the average \textit{passing rate} on the 14 TTCW for each of the four story types (GPT3.5, GPT4, Claude-v1.3, and New Yorker). Passing rate here corresponds to the percentage of time expert participants answer `Yes' to an individual TTCW for any given story. The New Yorker stories widely achieve the highest passing rate on all fourteen tests, with an overall pass rate of 84.7\%. In other words, individual New Yorker stories are assessed to pass 11.9 of the 14 TTCW on average. When examining performance on individual tests, no test receives a pass rate of 100\%, confirming that no test is an absolute requirement in high-quality creative writing, and experimentally justifying the need to conduct the TTCW tests as a set (Design Principle 4).

Moving to the performance of the LLM-generated stories, passing rates are much lower, with GPT3.5 stories passing less than 10\% of TTCW, while GPT4 and Claude v1.3 are closer to 30.0\%. In other words, LLM-generated stories pass between a third and a tenth of the TTCW compared to human-written New Yorker stories. When breaking down LLM-story performance across the Torrance dimensions, all models achieve their highest pass rate on the Fluency dimension, and Claude v1.3 achieves the highest performance on average across Fluency, Flexibility, and Elaboration, while GPT4 scores highest on the Originality dimension. This experimental finding is surprising as Claude v1.3 is an LLM that is smaller in size(52B) than GPT4 \footnote{https://the-decoder.com/gpt-4-architecture-datasets-costs-and-more-leaked/}. 

\subsubsection{Reproducibility of TTCW}

Since we collected three independent TTCW evaluations for each story, we can report the agreement levels of experts when they conduct the tests individually, and their assessment in aggregate. We compute the Fleiss $\kappa$ agreement across all annotations and report the interrater agreement level of each test in Table ~\ref{absolutehumaneval}. Individual test agreement ranges from 0.27 to 0.66, and averages at 0.41, suggesting moderate agreement on most of the individual TTCW.

Since the TTCW are designed to be additive, we further compute an aggregate score for each story by counting the number of TTCW tests a story passes. We visualize the results of this aggregate measure in Figure ~\ref{fig:ttcw_aggregate_histogram}. Since the aggregate measure is numerical (ranges from 0 to 14), we use Pearson correlation to measure agreement among experts. At this aggregate level, we obtain a correlation($\rho$) of 0.69, showing strong agreement among experts on the number of tests a story passes. In other words, even though experts reach slightly lower agreement on which exact TTCW a story passes or fails, they achieve strong agreement on the number of tests a story passes overall. This experimental finding confirms the importance of Design Principle 4, and the need for the tests to be performed as a set to achieve a reproducible evaluation of creativity for a given short story. When the objective is to evaluate the broad creativity in a short story, we recommend that all fourteen tests be administered as a set by one expert annotator, rather than by different experts or administering only an individual test, as this increases the reproducibility of the results.
\subsubsection{Comparative Evaluation Results}

The final portion of the evaluation protocol asks expert participants to rank the four shuffled stories in terms of subjective preference, as well as guess each story's origin between ``An experienced writer'', ``An amateur writer'', or ``Written by AI''. Figure~\ref{relativehumaneval1} summarizes the results from this final portion of the study.

Looking at the ranking results, human-written New Yorker stories were ranked as the most preferred story 89\% of the time, while the GPT-3.5-generated stories ranked as least preferred roughly two-thirds of the time. When comparing GPT-4 and Claude, Claude is almost twice as likely to rank as second (behind the human-written story) and was the most preferred on three of the four assessments in which the New Yorker story was not chosen as the most favored. These ranking results confirm and accentuate the observation from the test passing rates analysis that Claude V1.3 generates higher-quality short stories than models in the GPT family.

The attribution results paint a similar picture, with New Yorker stories predominantly attributed to an experienced writer, while LLM-generated stories get attributed to AI or an amateur writer. Interestingly, Claude V1.3 is more likely to be attributed to an amateur writer than an AI, whereas GPT3.5 and GPT4 stories are 80\%+ attributed to AI. One hypothesis for such behavior could be that the participants in our study might be more familiar with text written by OpenAI models, as these models are commercially more successful, providing an element of surprise to Claude-generated text.

\subsection{What can we infer from expert explanations of administering the TTCW?} \label{sec:analysis}
\begin{table*}[ht]
\small
\centering
\def\arraystretch{1.2}
\begin{tabular}{|l|l|lll|}
\hline
\multirow{4}{*}{\begin{tabular}[c]{@{}l@{}}Originality\\ in Thought\end{tabular}} & NewYorker & \multicolumn{3}{l|}{\begin{tabular}[c]{@{}l@{}}The ideas in this piece are unique, and expressed with original language. The metaphorical \\ language referenced above is a list of good examples. Others include the moment when she\\ slides her sunglasses down and everything goes darker; Rabbi Adler's monotonous drone \\ rendered as--son...his...own...flesh; Barbara rocking like the overloaded boat she's become.\\ This piece is practically bursting with new, exciting ways of expressing familiar things.\end{tabular}} \\ \cline{2-5} 
 & Claude & \multicolumn{3}{l|}{While the piece avoids overused expressions, its ideas and themes are hackneyed.} \\ \cline{2-5} 
 & GPT4 & \multicolumn{3}{l|}{\begin{tabular}[c]{@{}l@{}}The characters in this piece are so defined by their religion and culture as to be flattened by \\ stereotype. The events of this piece feel arbitrary, almost random. While that does grant it an\\ unpredictability and a vague form of originality, it feels thoughtless.\end{tabular}} \\ \cline{2-5} 
 & GPT3.5 & \multicolumn{3}{l|}{\begin{tabular}[c]{@{}l@{}}The piece relies on cliched turns of phrase to express actions and thoughts. Reality hits \\ Barbara like a tidal wave; days turn to weeks (and weeks?) and months; she uses her \\ experience to "bridge divides" and "heal wounds".\end{tabular}} \\ \hline
\end{tabular} 
\vspace{2ex}
\caption{\label{expertfeedbackcluster}Expert explanations on stories from NewYorker, Claude, GPT3.5 and GPT4 for one of the Originality test.}
\vspace{-4ex}
\end{table*}
\begin{table*}[!ht]
\small
\centering
\begin{tabular}{|l|l|l|}
\hline
Test & Passing Stories & Failing Stories \\ \hline
\begin{tabular}[c]{@{}l@{}}Understandability \\ \& Coherence\end{tabular} & \begin{tabular}[c]{@{}l@{}}Logical sequences, Interconnected details,\\ Rich character development, consistent\\ tone, Surprising yet earned conclusions, \\ Internal consistency in atmosphere and\\ symbolism\end{tabular} & \begin{tabular}[c]{@{}l@{}}Jumpy timelines, Unsatisfying endings, Undeveloped \\ plot points, Contradictory details, Illogical events, \\ Rambling prose, No overarching focus, Disjointed \\ Transition, Unclear Narration, Clichéd elements,\\ Inconsistent characterization\end{tabular} \\ \hline
\begin{tabular}[c]{@{}l@{}}Language \\ Proficiency \\ \& Literary \\Devices\end{tabular} & \begin{tabular}[c]{@{}l@{}}Utilize literary devices sparingly and directly,\\ enhancing the narratives that primarily draw \\ their strength from characterization and plot \\ rather than from highly figurative language.\end{tabular} & \begin{tabular}[c]{@{}l@{}}Attempts at figurative language come across as \\ overwrought, confusing, nonsensical. Subtlety and\\ restraint seem lacking, and the language choices \\ more often obscure rather than illuminate.\end{tabular} \\ \hline
\begin{tabular}[c]{@{}l@{}}Narrative\\ Pacing\end{tabular} & \begin{tabular}[c]{@{}l@{}}Adeptly manage time and pacing, varying \\ narrative rhythm, employing economic \\ language, skillfully transitioning between past\\ and present, creating suspense, mood, and interest\end{tabular} & \begin{tabular}[c]{@{}l@{}}Abrupt, jarring shifts in time or scene.Pacing that's\\ too slow/fast without purpose. Overly long descri-\\ ptions or passages. Events feeling randomly \\ ordered or disconnected\end{tabular} \\ \hline
\begin{tabular}[c]{@{}l@{}}Narrative\\ Ending\end{tabular} & \begin{tabular}[c]{@{}l@{}}Resolutions that weave earlier themes and\\ character perspectives into surprising yet \\ inevitable endings, leaving room for \\ interpretation\end{tabular} & \begin{tabular}[c]{@{}l@{}}Abrupt or unearned resolutions, overly moralistic \\ conclusions, unrealistic happy endings, unresolved \\ story threads, forced theme statements, narrative \\ disconnect, and unexpected surreal elements.\end{tabular} \\ \hline
\begin{tabular}[c]{@{}l@{}}Scene vs\\ Exposition\end{tabular} & \begin{tabular}[c]{@{}l@{}}Strategic balance of vivid scenes and concise \\ summaries, complementing each other to \\ enhance the narrative flow and reader experience\end{tabular} & \begin{tabular}[c]{@{}l@{}}Overreliance on summary leaves the narrative \\ ungrounded.Abrupt shifts between the two modes \\ prove confusing. Brief, underdeveloped scenes\\ fail to impact the reader.Overwrought scene \\ description obfuscates rather than illuminates\end{tabular} \\ \hline\hline
\begin{tabular}[c]{@{}l@{}}Emotional\\ Flexibility\end{tabular} & \begin{tabular}[c]{@{}l@{}}Harmonize interiority and exteriority, using \\ well-crafted transitions and narrative techniques\\ to convey character's inner states and external\\ actions, thereby fostering a deep connection\\ with the protagonist\end{tabular} & \begin{tabular}[c]{@{}l@{}}Unclear distinction between inner and outer realities, \\ Unrefined interior thoughts, insufficient external \\ grounding, cliched abstract language, an indistinct \\ mix of memory and metaphor with the present,\\ and underdeveloped character inner lives.\end{tabular} \\ \hline
\begin{tabular}[c]{@{}l@{}}Perspective\\ \& Voice\\ Flexibility\end{tabular} & \begin{tabular}[c]{@{}l@{}}Convincing \& diverse perspectives where \\ authors employ techniques such as nuanced \\ character development, authentic voice \\ crafting, detail-oriented storytelling, \&\\ efficient implication of perspectives, ensuring\\ a delicate balance between relatable \& \\ unlikeable traits\end{tabular} & \begin{tabular}[c]{@{}l@{}}Narrator's perspective dominates, and the other \\ characters are not portrayed convincingly as full human\\ beings with distinct viewpoints. While there are\\ some interesting moments, overall the story lacks truly\\ diverse perspectives that feel authentic and human.\end{tabular} \\ \hline
\begin{tabular}[c]{@{}l@{}}Structural\\ Flexibility\end{tabular} & \begin{tabular}[c]{@{}l@{}}Effective narrative turns that were surprising yet \\ believable, enhancing story depth, matching \\ themes and characters, increasing complexity\\ and stakes, aligning with the story’s style and \\ voice, and providing answers to previously raised\\ questions.\end{tabular} & \begin{tabular}[c]{@{}l@{}}Turns and surprises are ineffective, random, or\\ nonsensical rather than purposeful or appropriate.\\ While some stories sparked interest with an unusual\\ premise or relationship, they often failed to deliver \\ on the potential\end{tabular} \\ \hline
\end{tabular}
\vspace{2ex}
\caption{\label{expertexpl1}Common themes and issues found in expert explanations for tests focusing on TTCW-Fluency and TTCW-Flexibility}
\vspace{-5ex}
\end{table*}

Our annotation effort in Section~\ref{experteval} required experts to not only annotate for binary labels but also provide a justification paragraph accounting for their assessment. In Table~\ref{expertfeedbackcluster}, we provide examples of such justification for one of the TTCW tests for Originality. To gain insights into the main justifications experts provide for a story to pass or fail a TTCW test, we performed a manual thematic analysis of the expert justifications. We organized the results into a set of minimal phrases that often appear when a story passes or fails each TTCW test. Recent work has shown the utility of LLMs in clustering \cite{viswanathan2023large}. Based on these findings we asked the GPT4 model to cluster explanations across a given TTCW dimension into recurrent and broader representative themes. Three authors of the paper then manually verified these themes to ensure correctness. The outcome is summarized in Table~\ref{expertexpl1} for Fluency and Flexibility tests, and Table~\ref{expertexpl2} in the Appendix for Originality and Elaboration tests. The underlying themes found across the explanations reaffirm prior findings from \citet{ippolito2022creative} where writers found LLM-generated stories experiencing difficulty in maintaining a style/voice and easily reverting to tropes and repetition as well as those from \citet{mirowski2023cowriting} where screenwriters complained about the lack of subtext and character motivation.

\begin{table*}[!ht]
\small
\def\arraystretch{1.15}
\centering
\begin{tabular}{|l|l|}
\hline
E5 & \begin{tabular}[c]{@{}l@{}}\textbf{\color{blue}The AI rarely knew how to end a story} -- it would get bigger \& bigger in scope, leaving the characters behind \& talking\\ about their legacy and the community and the world at large. \textbf{\color{ForestGreen}Whenever the AI attempted metaphor or comparison, it}\\ \textbf{\color{ForestGreen}typically fell flat, with nonsensical analogies}. And it often didn't have a strong grasp on narrative structure; \textbf{\color{orange}characters} \\  \textbf{\color{orange}would sometimes appear without warning and then disappear without having any impact on the story itself}. The \\AI clearly knows the different elements of a story but doesn't have a grasp on how to merge them into a satisfying narrative.\end{tabular}                                                                                        \\ \hline
E3 & \begin{tabular}[c]{@{}l@{}}The story tends to use vague or cliched language, often repetitively. \textbf{\color{purple}Some words or phrases tended to be used again}\\ \textbf{\color{purple}and again over multiple stories for example ``\textit{inky sky},”  repeated a lot, as did ``\textit{etched}"}. There were parts that did not \\make sense over the course of the story. \textbf{\color{orange}A character might do something or think something in the beginning,}\\ \textbf{\color{orange}and then do something later that was contradictory or didn’t make sense}. Some of them were wildly trippy, like some\\ kind of surreal dream, in which nothing made sense.\textbf{\color{brown}The stories would spiral into a repetitive pattern, where the same}\\\textbf{\color{brown} things happened}, and the actions were described in similar ways. \textbf{\color{brown}It was as if the writer got stuck and just kept }\\ \textbf{\color{brown}repeating the same things}. This happened often when long periods of time passed.\end{tabular}                                                                                                                                                                      \\ \hline\hline
E4 & \begin{tabular}[c]{@{}l@{}}\textbf{\color{ForestGreen}Stories have over-modified descriptions but when AI does it, it's often followed by otherwise heightened diction} \\ in a way that human writing isn't. Sometimes, human writers will experiment with different syntax within their stories, but \\ I'd say, in my experience, it's somewhat rare. AI writing on the other hand demonstrated unusual/inconsistent syntax.Some\\ of the stories I believed were \textbf{\color{blue}AI-written had this weird forestalling of the ending}, which *could* happen with a novice \\ writer, for sure, but it didn't feel to me like a narrative choice a human writer would make. After the climax of the  story, \\\textbf{\color{blue}the falling action falls...and falls...and then just plateaus until the thing finally ends}\end{tabular}                                                                                                  \\ \hline\hline
E1 & \begin{tabular}[c]{@{}l@{}}\textbf{\color{ForestGreen}AI written sentences would be a series of words, positioned in a grammatically-correct fashion, with superficial}  \\ \textbf{\color{ForestGreen} shape that we associate with figurative language - but it just doesn't mean anything}. Figurative language adds \\meaning and context. It might be confusing, or poorly done, but there is always some idea being expressed in a non-literal\\ fashion. I've noticed that we, as real people, often answer questions in roundabout ways or ways that don't directly address \\the initial query. This brings a level of implied meaning or subtext to our conversations that hint at emotions and relationship\\ dynamics, something is often seen in high-quality fiction. However, \textbf{\color{red}I find AI-written dialogues disappointingly lacking}\\ \textbf{\color{red} in subtext}.They rely heavily on direct \& expositional dialogue, which feels decidedly unnatural \& less human. AI \textbf{\color{orange}characters}\\\textbf{\color{orange} frequently } \textbf{\color{orange} over-explain situations and emotions, resulting in an unrealistic portrayal of human interactions.}\\ Expert storytelling requires a nuanced understanding of reader engagement and artistic skill that amateur writers often \\lack, typically due to limited exposure to literature. This results in stories heavy in exposition and \textbf{\color{red}lacking in subtext}. AI\\-written stories take these issues to another level, often including random, incoherent elements and repeated pet phrases or\\ words.\textbf{\color{blue}These quirks become more} \textbf{\color{blue}noticeable over time, especially when AI attempts multiple disparate endings.}\\
Then consider the literal coherence of the stories. The definitely-AI ones would do random things \textbf{\color{orange}like introduce 'terrible }\\\textbf{\color{orange}beasts" who exist in the story for half a}\textbf{\color{orange} paragraph for some reason, and then just disappear.}
\textbf{\color{purple}There are also little}\\\textbf{\color{purple} pet phrases and words that repeat in the}\textbf{\color{purple}stories I am positive came from AI : a lot of "tendrils" and "inky}\\ \textbf{\color{purple}darkness."} And especially, a specific sentence construction of [time-orienting clause] [comma] [exposition].``\textit{In the days}\\\textit{that followed, X became a legend because Y.}" The same way authors have identifiable style and voice, so does AI.

\end{tabular} \\ \hline\hline
E2 & \begin{tabular}[c]{@{}l@{}}It seemed like there were two AI models that were generating stories for each batch. I feel like I began to notice and read\\ them as I would read or notice works by the same writer. \textbf{\color{brown} One would rapidly accelerate through time after the first}\\ \textbf{\color{brown} scene or so, probably once it was loose from the parameters it had been fed.}These accelerations were broadly pretty\\ absurd and hysterical. The other was a bit more subtle, but would \textbf{\color{blue}consistently end its stories  abruptly and employ}\\\textbf{\color{blue} somewhat bizarre logic}. In both cases, the \textbf{\color{red} AI seemed entirely unable to use implication or subtext}  \textbf{\color{ForestGreen}and its use of,}\\ \textbf{\color{ForestGreen} images, metaphors, etc was always very simple.}
\end{tabular}\\\hline
\end{tabular}
\vspace{2ex}
\caption{\label{expertfeedback}How do experts differentiate between AI vs. human-written stories?}
\end{table*}

\subsection{How do experts differentiate between human-written and LLM-generated stories?}
\label{expertvsai}

In an optional exchange with the expert participants (Section ~\ref{participant}) who participated in the annotation (Section ~\ref{experteval}), they were given the opportunity to describe how they differentiated between AI-generated and human-written stories. Collected responses showed that expert evaluators did not make a decision on which stories were AI-generated based on factors such as grammatical characteristics. Table \ref{expertfeedback} lists the replies of a few experts, which we color-coded to highlight the recurrent issues that led them to believe that a story is AI-generated. Feedback was often aligned with individual TTCW, demonstrating that experts discriminated AI vs human written stories based on creative execution rather than spurious cues.

In particular, E5, E4, E1 thought AI struggles at \textbf{\color{blue}\underline{Narrative Ending}}. E5 and E4 highlighted that AI-generated stories would forestall the ending by getting bigger in scope. E1 highlighted that AI-generated stories would have multiple disparate endings. E5, E4, E1, and E2 all highlighted that AI-generated stories would often contain abstruse and incoherent metaphors that do not add meaning or extremely cliched or simple metaphors thereby demonstrating poor \textbf{\color{ForestGreen}\underline{Language Proficiency and use of Literary Devices}}. E1 highlighted one such example in a story - \textit{However, she managed to laugh louder and louder until her laughter transformed into an embrace of the sun's atmosphere}.

E1 and E2 further highlighted that characters in AI-generated stories have poor \textbf{\color{red}\underline{Rhetorical Complexity}} and are often lacking in subtext. E1 further added that AI-generated stories operate in a nearly opposite and Drax-like fashion in which there is only literal meaning, and that literal meaning is often nonsensical, or at least presented without any of the context that might make it seem like something a human would say or do.
E1 highlighted an exchange below from an AI-generated story:

\begin{center}
    Sarah: \textit{``We've been avoiding the inevitable, Max. During our time here we've grown closer, and now that it's almost over, we can't just pretend like it never happened.''} \\
    Max:  \textit{``I understand, Sarah. But how do we move forward? How do we navigate this complexity without unraveling everything we've built?''}
\end{center}

where he exclaimed that these statements make hackish sense as clumsy exposition directed at the reader, and no sense at all as sentences spoken from one alleged human being to another.
Both E5, E3, and E1 agreed on poor \textbf{\color{orange}\underline{Character Development}} in AI-generated stories where a character would appear and then disappear without having any impact. E3 and E2 also highlighted issues in \textbf{\color{brown}\underline{Narrative Pacing}} where stories would either spiral into a repetitive pattern or rapidly accelerate through time after the first scene. E1 and E4 also highlighted \textbf{\color{purple}\underline{Unusual Syntax}} in sentence structure in AI-generated stories and repetition of certain words and phrases across stories.

\section{TTCW Implementation with LLMs as Assessors}
\label{sec:llmeval}
Expert annotation such as the one we perform in Section~\ref{sec:data} is costly: based on our evaluation protocol, evaluating a 1500-2500 word short story with a qualified expert costs \$20, and requires roughly 30 minutes of the expert's time. Prior work has shown the promise of using LLMs in text evaluation. For instance, GPT3.5 and GPT4 are effective at evaluating the factual consistency of a summary to its document \cite{laban2023llms}, or measuring the coherence of a summary \cite{gao2023human}. GPTEval \cite{liu2023gpteval} employs the framework of using large language models with chain-of-thoughts (CoT) \cite{wei2022chain} to assess the quality of NLG outputs. Recent work has also applied LLM-based evaluation to the creative domain \cite{rajani2023llm_labels}, claiming that GPT4 can achieve a high correlation with humans when evaluating brainstorming or creative generation tasks. In this section, we describe our TTCW implementation with LLMs as assessors to understand LLM's ability to assess creative writing.

We apply a similar evaluation protocol to the one described in Section \ref{sec:data}. We use the same data selection and the same three LLMs:
GPT3.5, GPT4, and Claude, and prompt them to answer the 14 individual TTCW tests for the 48 stories in our collection. 

Prior work \citet{wei2022chain} has shown how generating a \textit{chain of thought} -- a series of intermediate reasoning steps -- significantly improves the ability of large language models to perform complex reasoning. Taking advantage of this we design the prompts/instructions for large language models in a slightly different fashion than for the human experts as can be seen in Table \ref{prompting1}(Row 3 (Human Instruction) vs Row 4 LLM Instruction). To help the model make an informed decision we first ask it to list out elements specific to any given test such as ``elements in the story that call to each of the five senses" for the World Building and setting test followed by asking it to decide overall and then provide its reasoning before choosing an answer between `Yes' or `No'.
The exact prompt contains (1) the story, (2) the expanded TTCW context, (3) the TTCW question, and (4) an LLM-specific instruction guiding the model to perform the task in a chain-of-though manner. We then measure model agreement with the \textit{majority vote} of the three experts that conducted the test, using Cohen's Kappa.

\begin{table*}[!ht]
\small
\centering
\begin{tabular}{ll|lll|}
\hline
\multicolumn{1}{|l|}{Dimension}                    & Test                                                                               & \multicolumn{1}{l|}{GPT3.5} & \multicolumn{1}{l|}{GPT4}  & Claude \\ \hline
\multicolumn{1}{|l|}{\multirow{5}{*}{Fluency}}     & Understandability \& Coherence                                                     & \multicolumn{1}{l|}{-0.01}   & \multicolumn{1}{l|}{-0.01} & -0.17                                                      \\ \cline{2-5} 
\multicolumn{1}{|l|}{}                             & Narrative Pacing                                                                   & \multicolumn{1}{l|}{0.05}   & \multicolumn{1}{l|}{0.0}   &  -0.22                                                    \\ \cline{2-5} 
\multicolumn{1}{|l|}{}                             & Scene vs Exposition                                                                & \multicolumn{1}{l|}{-0.03}  & \multicolumn{1}{l|}{-0.08}  &  -0.23                                                          \\ \cline{2-5} 
\multicolumn{1}{|l|}{}                             & \begin{tabular}[c]{@{}l@{}}Literary Devices \& Language Proficiency\end{tabular} & \multicolumn{1}{l|}{0.04}   & \multicolumn{1}{l|}{-0.09} &  -0.11                                                         \\ \cline{2-5} 
\multicolumn{1}{|l|}{}                             & Narrative Ending                                                                   & \multicolumn{1}{l|}{-0.02}  & \multicolumn{1}{l|}{0.02} &  0.02                                                          \\ \hline\hline
\multicolumn{1}{|l|}{\multirow{3}{*}{Flexibility}} & Emotional Flexibility                                                              & \multicolumn{1}{l|}{-0.04}  & \multicolumn{1}{l|}{0.0}   & 0.09                                                          \\ \cline{2-5} 
\multicolumn{1}{|l|}{}                             & Perspective \& Voice Flexibility                                                   & \multicolumn{1}{l|}{0.0}    & \multicolumn{1}{l|}{0.26}  & 0.14                                                       \\ \cline{2-5} 
\multicolumn{1}{|l|}{}                             & Structural Flexibility                                                             & \multicolumn{1}{l|}{-0.04}  & \multicolumn{1}{l|}{0.0}   &  -0.07                                                   \\ \hline\hline
\multicolumn{1}{|l|}{\multirow{3}{*}{Originality}}                             & Originality in Form                                                             & \multicolumn{1}{l|}{0.08}   & \multicolumn{1}{l|}{0.09}  &  0.03                                                         \\ \cline{2-5} 
\multicolumn{1}{|l|}{}                             & Originality in Thought                                                             & \multicolumn{1}{l|}{0.19}   & \multicolumn{1}{l|}{0.31}  &  0.15                                                          \\ \cline{2-5} 
\multicolumn{1}{|l|}{}                             & Originality in Theme \& Content                                                    & \multicolumn{1}{l|}{0.06}   & \multicolumn{1}{l|}{-0.01} &  0.18                                                        \\ \hline\hline
\multicolumn{1}{|l|}{\multirow{3}{*}{Elaboration}} & World Building \& Setting                                                          & \multicolumn{1}{l|}{0.0}   & \multicolumn{1}{l|}{0.00}  &  0.09                                                            \\ \cline{2-5} 
\multicolumn{1}{|l|}{}                             & Character Development                                                              & \multicolumn{1}{l|}{-0.08}  & \multicolumn{1}{l|}{0.02}  &  0.00                                                          \\ \cline{2-5} 
\multicolumn{1}{|l|}{}                             & Rhetorical Complexity                                                              & \multicolumn{1}{l|}{0.0}    & \multicolumn{1}{l|}{0.0}   &  0.02                                                          \\ \hline\hline
\multicolumn{2}{|c|}{Average}                                                                                                           & \multicolumn{1}{l|}{0.016}   & \multicolumn{1}{l|}{0.035}  & -0.006                                                     \\ \hline
\end{tabular}
\vspace{2ex}
\caption{\label{llmhumancor} Correlation between LLM-administered TTCW and expert annotations (Cohen's Kappa) on all 48 stories}
\vspace{-3ex}
\end{table*}

\subsection{Results}

Table~\ref{llmhumancor} summarizes correlation results between LLM-based and expert-based TTCW assessments. On average, we find that none of the LLMs produce assessments that correlate positively with expert assessments, with correlation averages close to zero. GPT4 is the only model to obtain correlations above 0.2 on two of the fourteen tests, yet this still does not qualify as moderate agreement.This empirical result contrasts with prior work: even though GPT4 has been shown to have some ability to evaluate creativity in short-form tasks (such as responses with less than 100 words) \cite{rajani2023llm_labels}, our work shows that this result does not extend to longer-form evaluation. We note that although the prompts we utilized were zero-shot in nature (i.e., these prompts did not include example binary labels and justifications from experts), we experimented with few-shot prompts for a couple of the TTCW tests and did not obtain any significant correlation gains.

Yet LLM-administered TTCW would be a crucial building block in improving model-generated stories. Assuming that an automated method could produce reliable TTCW outcomes, it could be used in iterative algorithms such as Self-Refine \cite{madaan2023self} to iteratively edit a draft story until it passes a large proportion of tests. With this in mind, we release the TTCW benchmark which contains all binary judgments we collected and expert justifications, with the hope that the community can use it as a tool to track progress in the evaluation of the creative capabilities of LLMs.

To get a deeper understanding of how expert and LLM explanations differ we take a closer look at them. We explicitly prompted LLMs to do step-by-step reasoning before arriving at any verdict and this was often reflected in the explanations. The LLM-generated explanations were procedural and typically lengthier than expert explanations. While there were not any specific instructions given to experts about the length of the explanations we asked them to provide necessary details justifying their decision. Table~\ref{llmvsexpertexpl} in the Appendix shows such an example. 
\section{Discussion}

\subsection{Are experts good at detecting AI written texts or are they simply good at detecting lower quality texts?}
During our evaluation, expert annotators were tasked with predicting a story's author from three categories (Written by an Expert, Written by an Amateur, or AI-generated), even though none of the stories they annotated were selected from amateur authors. This mismatch was intended to reduce the likelihood of annotators tracing back the creation of the dataset and provide a more granular scale for their prediction.

Our goal is to design a rubric for creative writing evaluation that rates higher-quality text as better compared to lower-quality text. It is by no means designed to penalize AI-generated text. The primary reason for this was that it is really hard to detect AI-generated text because current large language models are good at generating fluent and human-like text. Certain traits are common in fiction written by an AI and an amateur writer which makes it difficult for the task of authorship attribution. One of our experts remarked \textit{There is, at best, little subtext in an amateur-written story, and likely none at all. This is the aspect that sometimes made me hesitate as to whether a story was written by AI or an amateur writer.} While both AI and amateur-written fiction gravitate towards clichés, several experts mentioned AI \textit{writing tics} that led them to believe it is not written by a human. Unlike text written by a mediocre writer, AI-written text would often contain predictable formations: strong topic sentences at the top of paragraphs; and summary sentences at the end of paragraphs; Some experts also mentioned peculiar sentence construction that was common in AI writing such as [time-orienting clause][comma][exposition]. Certain LLMs such as GPT4 also conflate good writing with ornamental use of language. This leads to generated text that is full of lofty superficial figurative language unlike text written by amateurs. Finally, AI's voice in writing tended to be overly moralizing, unlike amateur writers. This evidence somewhat makes us believe that experts might have an easier time discriminating when lower-quality text is produced by a human vs a machine.

\subsection{LLMs: from study subject to HCI research tool}

This work leveraged LLMs as a tool to facilitate research, such as generating the first pass of the ``expanded expert measure'' text, or clustering expert-written explanations. Recent work has defined researchers and model developers as a main user category of LLMs \cite{chen2023next}, with recent efforts for example making use of models to generate synthetic HCI research data \cite{hamalainen2023evaluating}. We reflect on the utility and limitations of using LLMs as a research tool in our work. We used an LLM in Section~\ref{sec:prompting} to assist in writing the ``expanded expert measure'' descriptions, filling in the missing detail and context in expert-written descriptions, by adding definitions of technical terms or providing context or an example. LLMs were also useful in situations where we had to process a large volume of textual data, such as the clustering of the 2016 expert-written explanations we performed in Section~\ref{sec:analysis}. Manually clustering such explanations would have been prohibitively time-consuming, and LLMs enabled us to swiftly extract initial insights, which we discussed and refined. In either situation, we approach the use of LLMs in a human-AI collaborative framework, where the output of the LLM is validated by humans (in this case, the authors of the paper) and serves as an initial step in the research process rather than the final product.

\subsection{Towards interactive LLM-based creative co-writing} 
Our experimental results and the analysis of the expert explanations highlight the limitations of current LLMs in generating both high-quality fictional stories as well as assessing the creativity of such existing stories. With the rapid progression in LLM development, we make available a corpus containing expert evaluations of TTCW assessments. We believe such a contribution will facilitate the evaluation of the upcoming model's abilities for creative writing assessment. If LLMs are capable of producing TTCW assessments that correlate positively with expert judgments, future work can explore new opportunities for creative LLM-based co-writing interfaces.

In particular, we envision LLMs assisting \textit{Planning} and \textit{Reviewing}, crucial phases in the cognitive process theory of writing \cite{flower1981cognitive}. In prior work \citet{gero2023social} states that writers expressed the importance of specificity in the feedback instead of generic feedback like \textit{this might be a bit boring}. We hope that the TTCW tests can provide the structure in future work looking to provide targeted feedback on writing. Further \citet{ippolito2022creative} recently pointed out that professional writers constantly felt that LLM-generated text is rife with cliches and overused tropes. Metrics that quantify elements like originality in theme, structural flexibility, or rhetorical complexity could guide creative writing support tools\footnote{\url{https://www.sudowrite.com/}} built using current LLMs, thereby improving \textit{planning} and \textit{translation} \cite{10.1145/3461778.3462050}.

\subsection{Is TTCW universal in nature?}

The development of TTCW was informed by the expertise of 8 field specialists, with each test reflecting the insights of 1-3 experts. However, TTCW cannot be considered a universal benchmark for creative writing due to its reliance on a narrow expert base, potentially echoing Western literary biases due to the panel's background in "highbrow" literary fiction. This might marginalize other cultural narratives and styles. Specifically, TTCW evaluates creative writing through various lenses: TTCW Flexibility1 values diverse narrative perspectives, potentially disadvantaging stories focused on a singular viewpoint. TTCW Fluency2 and TTCW Flexibility2 assess the balance in storytelling elements, which might not favor experimental works that blend narrative techniques. Despite focusing on short stories, TTCW also critiques story closure in TTCW Fluency4, contrasting with plays that seek non-cathartic endings for specific impacts, as seen in the Brechtian tradition. Moreover, TTCW Originality3 rewards formal innovation, possibly underrating stories that creatively navigate within strict formal boundaries, like those adhering to traditional mythic structures. Thus, while TTCW offers a structured approach to evaluating creative writing, its scope and applicability are affected by certain choices and specific literary conventions it prioritizes.
\section{Limitations and Future Work}

\paragraph{Response Diversity and Temperature.} In our experimental design, we employed the default generation parameters for the models, specifically a temperature setting, ($T=1.0$), aiming to evaluate the model's capabilities in a non-optimized setting. Notably, variations in parameters such as temperature have been documented to potentially enhance the originality of generated content \cite{roemmele2018automated}. Consequently, there exists a possibility that utilizing alternative generation parameters might lead to outputs that surpass a greater fraction of the TTCW. As the associated costs of conducting TTCW evaluations decrease, subsequent research endeavors could provide a more comprehensive insight into the influence of generation parameters on creativity within the context of the TTCW framework.

\paragraph{Coverage of the TTCW}
When selecting which expert measures would yield a TTCW test, we attempted to maximize coverage while minimizing overlap between the tests. Yet it is very likely that the passing of a test is correlated with another test, as the tests touch on common story elements (e.g., characters, prose). We did not investigate the degree of overlap between pairs of tests. We encourage future work to further refine the TTCW tests and propose additional tests.

\paragraph{Prompt Engineering and Optimization.}
We open-source the prompts used to instruct the LLMs to generate the stories as well as prompts to administer the TTCW tests, which we iterated on and reviewed carefully. While there is no upper bound on engineering the `best' prompt for a particular task, the output of an LLM is still dependent on the input prompt \cite{zhou2022large}. The LLMs that we evaluate in the study are closed-source models and the quality of their generation has been observed to change over time \cite{chen2023chatgpt}. We hope future work can explore further refining of the prompts and parameters used in our work, and exploring their impact on the results.

\paragraph{Generalization beyond short fictional stories.}
Our tests were explicitly designed for short fiction and we do not study the generalization of TTCW to other forms of creative writing. Empirical evaluation would need to be conducted to verify whether the TTCW is adequate and comprehensive to evaluate other forms of creative writing, including scripts, novels, or marketing material such as slogans. We posit there are specific metrics for each specific type of creative writing, which our current tests do not cover. For example, evaluating or critiquing poetry might require different fine-grained evaluation metrics compared to short stories. 

\paragraph{Objectivity in the Evaluation of Creativity.}
Creativity emerges over time in a complex interplay of factors. Human judgments of creativity are often biased by personal tastes, expectations, and hindsight. The definition of an `expert' or `amateur' creative writer is not clear-cut in a field that has unclear professional delineations \cite{gero2023social}. Many successful writers retain full-time jobs as teachers, editors, or in unrelated professions, as few are able to make a living from their writing alone. Although we emphasize the validity of our results by computing agreement levels among recruited experts, we note that work in creativity should not solely aim to maximize agreement, since subjective differences are an inherent property of divergent thinking, which is central to creativity. Future work looking to expand our line of work should seek to select experts with diverse backgrounds, offering a more comprehensive view of the creative process. 
\section{Conclusion}

In this work, we utilize the widely accepted Torrance Test of Creative Thinking originally designed for the evaluation of creativity as a process, and align it towards the evaluation of creativity as a product. In particular, we focus on short fiction writing and formulate the Torrance Test for Creative Writing (TTCW). Our collaborative process, involving creative writing experts in both the development and validation stages of the TTCW, has established the test as a robust instrument for assessing creativity in fictional short stories. The experimental data derived from the evaluation of both human-authored and LLM-generated stories offers a rich comparative analysis. It underscores the proficiency of seasoned writers in evoking creativity, outperforming LLMs by a considerable margin. Our analysis also highlights the disparity in creative prowess among different LLMs, revealing that while certain LLMs might exhibit proficiency in some dimensions of creativity, there exists a wide chasm between them and human expertise when assessed holistically. Notably, the TTCW tests also provided a granular perspective on the areas where LLMs falter the most in terms of creativity. Our secondary investigation into the feasibility of LLMs in reproducing expert assessments yielded that, at least in their current state, LLMs are not yet adept at administering TTCW tests. This indicates a dual challenge: not only are LLMs lagging in producing inherently creative content, but they also lack the finesse to evaluate creativity as experts do. We hope that our evaluation framework using TTCW, and findings on the creative capabilities of LLMs coupled with our dataset will steer future work and innovations in creativity research.

%TC:ignore
\bibliographystyle{ACM-Reference-Format}
\bibliography{sample-base}


\begin{thebibliography}{76}

%%% the Web address.

\ifx \showCODEN    \undefined \def \showCODEN     #1{\unskip}     \fi
\ifx \showDOI      \undefined \def \showDOI       #1{#1}\fi
\ifx \showISBNx    \undefined \def \showISBNx     #1{\unskip}     \fi
\ifx \showISBNxiii \undefined \def \showISBNxiii  #1{\unskip}     \fi
\ifx \showISSN     \undefined \def \showISSN      #1{\unskip}     \fi
\ifx \showLCCN     \undefined \def \showLCCN      #1{\unskip}     \fi
\ifx \shownote     \undefined \def \shownote      #1{#1}          \fi
\ifx \showarticletitle \undefined \def \showarticletitle #1{#1}   \fi
\ifx \showURL      \undefined \def \showURL       {\relax}        \fi
% invisible to TeX
\providecommand\bibfield[2]{#2}
\providecommand\bibinfo[2]{#2}
\providecommand\natexlab[1]{#1}
\providecommand\showeprint[2][]{arXiv:#2}

\bibitem[Abdel~Latif(2013)]%
        {abdel2013we}
\bibfield{author}{\bibinfo{person}{Muhammad M~Mahmoud Abdel~Latif}.} \bibinfo{year}{2013}\natexlab{}.
\newblock \showarticletitle{What do we mean by writing fluency and how can it be validly measured?}
\newblock \bibinfo{journal}{\emph{Applied linguistics}} \bibinfo{volume}{34}, \bibinfo{number}{1} (\bibinfo{year}{2013}), \bibinfo{pages}{99--105}.
\newblock


\bibitem[Accocela(2012)]%
        {BadEndings}
\bibfield{author}{\bibinfo{person}{Joan Accocela}.} \bibinfo{year}{2012}\natexlab{}.
\newblock \showarticletitle{On Bad Endings}.
\newblock \bibinfo{journal}{\emph{The NewYorker}} (\bibinfo{year}{2012}).
\newblock
\urldef\tempurl%
\url{https://www.newyorker.com/books/page-turner/on-bad-endings}
\showURL{%
\tempurl}


\bibitem[Amabile(1982)]%
        {amabile1982social}
\bibfield{author}{\bibinfo{person}{Teresa~M Amabile}.} \bibinfo{year}{1982}\natexlab{}.
\newblock \showarticletitle{Social psychology of creativity: A consensual assessment technique.}
\newblock \bibinfo{journal}{\emph{Journal of personality and social psychology}} \bibinfo{volume}{43}, \bibinfo{number}{5} (\bibinfo{year}{1982}), \bibinfo{pages}{997}.
\newblock


\bibitem[Anthropic(2022)]%
        {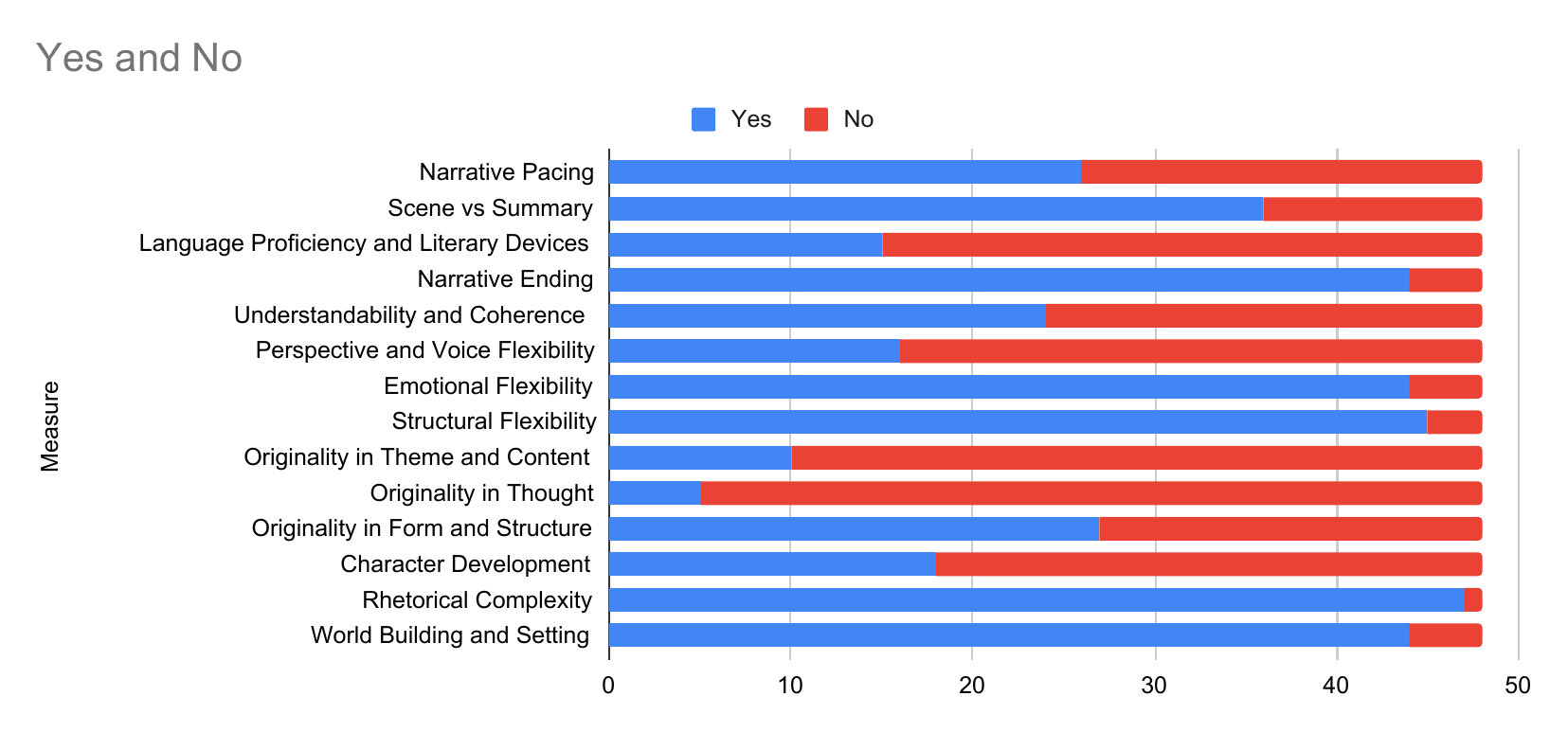}
\bibfield{author}{\bibinfo{person}{Anthropic}.} \bibinfo{year}{2022}\natexlab{}.
\newblock \showarticletitle{Introducing Claude}.
\newblock  (\bibinfo{year}{2022}).
\newblock
\urldef\tempurl%
\url{https://www.anthropic.com/index/introducing-claude}
\showURL{%
\tempurl}


\bibitem[Baer(2014)]%
        {baer2014creativity}
\bibfield{author}{\bibinfo{person}{John Baer}.} \bibinfo{year}{2014}\natexlab{}.
\newblock \bibinfo{booktitle}{\emph{Creativity and divergent thinking: A task-specific approach}}.
\newblock \bibinfo{publisher}{Psychology Press}.
\newblock


\bibitem[Baer and McKool(2009)]%
        {baer2009assessing}
\bibfield{author}{\bibinfo{person}{John Baer} {and} \bibinfo{person}{Sharon~S McKool}.} \bibinfo{year}{2009}\natexlab{}.
\newblock \showarticletitle{Assessing creativity using the consensual assessment technique}.
\newblock In \bibinfo{booktitle}{\emph{Handbook of research on assessment technologies, methods, and applications in higher education}}. \bibinfo{publisher}{IGI Global}, \bibinfo{pages}{65--77}.
\newblock


\bibitem[Beaty and Johnson(2021)]%
        {beaty2021automating}
\bibfield{author}{\bibinfo{person}{Roger~E Beaty} {and} \bibinfo{person}{Dan~R Johnson}.} \bibinfo{year}{2021}\natexlab{}.
\newblock \showarticletitle{Automating creativity assessment with SemDis: An open platform for computing semantic distance}.
\newblock \bibinfo{journal}{\emph{Behavior research methods}} \bibinfo{volume}{53}, \bibinfo{number}{2} (\bibinfo{year}{2021}), \bibinfo{pages}{757--780}.
\newblock


\bibitem[Bernstein et~al\mbox{.}(2010)]%
        {bernstein2010soylent}
\bibfield{author}{\bibinfo{person}{Michael~S Bernstein}, \bibinfo{person}{Greg Little}, \bibinfo{person}{Robert~C Miller}, \bibinfo{person}{Bj{\"o}rn Hartmann}, \bibinfo{person}{Mark~S Ackerman}, \bibinfo{person}{David~R Karger}, \bibinfo{person}{David Crowell}, {and} \bibinfo{person}{Katrina Panovich}.} \bibinfo{year}{2010}\natexlab{}.
\newblock \showarticletitle{Soylent: a word processor with a crowd inside}. In \bibinfo{booktitle}{\emph{Proceedings of the 23nd annual ACM symposium on User interface software and technology}}. \bibinfo{pages}{313--322}.
\newblock


\bibitem[Biggs and Collis(1982)]%
        {biggs1982psychological}
\bibfield{author}{\bibinfo{person}{John~B Biggs} {and} \bibinfo{person}{Kevin~F Collis}.} \bibinfo{year}{1982}\natexlab{}.
\newblock \showarticletitle{The psychological structure of creative writing}.
\newblock \bibinfo{journal}{\emph{Australian Journal of Education}} \bibinfo{volume}{26}, \bibinfo{number}{1} (\bibinfo{year}{1982}), \bibinfo{pages}{59--70}.
\newblock


\bibitem[Boardman(1992)]%
        {boardman1992narrative}
\bibfield{author}{\bibinfo{person}{Michael~M Boardman}.} \bibinfo{year}{1992}\natexlab{}.
\newblock \bibinfo{booktitle}{\emph{Narrative Innovation and Incoherence: Ideology in Defoe, Goldsmith, Austen, Eliot, and Hemingway}}.
\newblock \bibinfo{publisher}{Duke University Press}.
\newblock


\bibitem[Bourdeau et~al\mbox{.}(2020)]%
        {10.1145/3313831.3376495}
\bibfield{author}{\bibinfo{person}{Simon Bourdeau}, \bibinfo{person}{Annemarie Lesage}, \bibinfo{person}{B\'{e}atrice Couturier~Caron}, {and} \bibinfo{person}{Pierre-Majorique L\'{e}ger}.} \bibinfo{year}{2020}\natexlab{}.
\newblock \showarticletitle{When Design Novices and LEGO® Meet: Stimulating Creative Thinking for Interface Design}. In \bibinfo{booktitle}{\emph{Proceedings of the 2020 CHI Conference on Human Factors in Computing Systems}} (Honolulu, HI, USA) \emph{(\bibinfo{series}{CHI '20})}. \bibinfo{publisher}{Association for Computing Machinery}, \bibinfo{address}{New York, NY, USA}, \bibinfo{pages}{1–14}.
\newblock
\showISBNx{9781450367080}
\urldef\tempurl%
\url{https://doi.org/10.1145/3313831.3376495}
\showDOI{\tempurl}


\bibitem[Burroway et~al\mbox{.}(2019)]%
        {burroway2019writing}
\bibfield{author}{\bibinfo{person}{Janet Burroway}, \bibinfo{person}{Elizabeth Stuckey-French}, {and} \bibinfo{person}{Ned Stuckey-French}.} \bibinfo{year}{2019}\natexlab{}.
\newblock \bibinfo{booktitle}{\emph{Writing fiction: A guide to narrative craft}}.
\newblock \bibinfo{publisher}{University of Chicago Press}.
\newblock


\bibitem[Campe and Weber(2014)]%
        {campe2014rethinking}
\bibfield{author}{\bibinfo{person}{R{\"u}diger Campe} {and} \bibinfo{person}{Julia Weber}.} \bibinfo{year}{2014}\natexlab{}.
\newblock \bibinfo{booktitle}{\emph{Rethinking Emotion: Interiority and Exteriority in Premodern, Modern, and Contemporary Thought}}. Vol.~\bibinfo{volume}{15}.
\newblock \bibinfo{publisher}{Walter de Gruyter GmbH \& Co KG}.
\newblock


\bibitem[Chakrabarty et~al\mbox{.}(2023)]%
        {chakrabarty2023creativity}
\bibfield{author}{\bibinfo{person}{Tuhin Chakrabarty}, \bibinfo{person}{Vishakh Padmakumar}, \bibinfo{person}{Faeze Brahman}, {and} \bibinfo{person}{Smaranda Muresan}.} \bibinfo{year}{2023}\natexlab{}.
\newblock \showarticletitle{Creativity Support in the Age of Large Language Models: An Empirical Study Involving Emerging Writers}.
\newblock \bibinfo{journal}{\emph{arXiv preprint arXiv:2309.12570}} (\bibinfo{year}{2023}).
\newblock


\bibitem[Chen et~al\mbox{.}(2023b)]%
        {chen2023chatgpt}
\bibfield{author}{\bibinfo{person}{Lingjiao Chen}, \bibinfo{person}{Matei Zaharia}, {and} \bibinfo{person}{James Zou}.} \bibinfo{year}{2023}\natexlab{b}.
\newblock \showarticletitle{How is ChatGPT's behavior changing over time?}
\newblock \bibinfo{journal}{\emph{arXiv preprint arXiv:2307.09009}} (\bibinfo{year}{2023}).
\newblock


\bibitem[Chen et~al\mbox{.}(2023a)]%
        {chen2023next}
\bibfield{author}{\bibinfo{person}{Xiang'Anthony' Chen}, \bibinfo{person}{Jeff Burke}, \bibinfo{person}{Ruofei Du}, \bibinfo{person}{Matthew~K Hong}, \bibinfo{person}{Jennifer Jacobs}, \bibinfo{person}{Philippe Laban}, \bibinfo{person}{Dingzeyu Li}, \bibinfo{person}{Nanyun Peng}, \bibinfo{person}{Karl~DD Willis}, \bibinfo{person}{Chien-Sheng Wu}, {et~al\mbox{.}}} \bibinfo{year}{2023}\natexlab{a}.
\newblock \showarticletitle{Next steps for human-centered generative AI: A technical perspective}.
\newblock \bibinfo{journal}{\emph{arXiv preprint arXiv:2306.15774}} (\bibinfo{year}{2023}).
\newblock


\bibitem[Chen et~al\mbox{.}(2023c)]%
        {chen2023ambient}
\bibfield{author}{\bibinfo{person}{Zexin Chen}, \bibinfo{person}{Eric Zhou}, \bibinfo{person}{Kenneth Eaton}, \bibinfo{person}{Xiangyu Peng}, {and} \bibinfo{person}{Mark Riedl}.} \bibinfo{year}{2023}\natexlab{c}.
\newblock \showarticletitle{Ambient Adventures: Teaching ChatGPT on Developing Complex Stories}.
\newblock \bibinfo{journal}{\emph{arXiv preprint arXiv:2308.01734}} (\bibinfo{year}{2023}).
\newblock


\bibitem[Chung et~al\mbox{.}(2021)]%
        {10.1145/3461778.3462050}
\bibfield{author}{\bibinfo{person}{John Joon~Young Chung}, \bibinfo{person}{Shiqing He}, {and} \bibinfo{person}{Eytan Adar}.} \bibinfo{year}{2021}\natexlab{}.
\newblock \showarticletitle{The Intersection of Users, Roles, Interactions, and Technologies in Creativity Support Tools}. In \bibinfo{booktitle}{\emph{Proceedings of the 2021 ACM Designing Interactive Systems Conference}} (Virtual Event, USA) \emph{(\bibinfo{series}{DIS '21})}. \bibinfo{publisher}{Association for Computing Machinery}, \bibinfo{address}{New York, NY, USA}, \bibinfo{pages}{1817–1833}.
\newblock
\showISBNx{9781450384766}
\urldef\tempurl%
\url{https://doi.org/10.1145/3461778.3462050}
\showDOI{\tempurl}


\bibitem[Clark et~al\mbox{.}(2021)]%
        {clark-etal-2021-thats}
\bibfield{author}{\bibinfo{person}{Elizabeth Clark}, \bibinfo{person}{Tal August}, \bibinfo{person}{Sofia Serrano}, \bibinfo{person}{Nikita Haduong}, \bibinfo{person}{Suchin Gururangan}, {and} \bibinfo{person}{Noah~A. Smith}.} \bibinfo{year}{2021}\natexlab{}.
\newblock \showarticletitle{All That{'}s {`}Human{'} Is Not Gold: Evaluating Human Evaluation of Generated Text}. In \bibinfo{booktitle}{\emph{Proceedings of the 59th Annual Meeting of the Association for Computational Linguistics and the 11th International Joint Conference on Natural Language Processing (Volume 1: Long Papers)}}. \bibinfo{publisher}{Association for Computational Linguistics}, \bibinfo{address}{Online}, \bibinfo{pages}{7282--7296}.
\newblock
\urldef\tempurl%
\url{https://doi.org/10.18653/v1/2021.acl-long.565}
\showDOI{\tempurl}


\bibitem[Clark(2008)]%
        {clark2008writing}
\bibfield{author}{\bibinfo{person}{Roy~Peter Clark}.} \bibinfo{year}{2008}\natexlab{}.
\newblock \bibinfo{booktitle}{\emph{Writing tools: 55 essential strategies for every writer}}.
\newblock \bibinfo{publisher}{Little, Brown Spark}.
\newblock


\bibitem[Currie(1990)]%
        {currie1990nature}
\bibfield{author}{\bibinfo{person}{Gregory Currie}.} \bibinfo{year}{1990}\natexlab{}.
\newblock \bibinfo{booktitle}{\emph{The nature of fiction}}.
\newblock \bibinfo{publisher}{Cambridge University Press}.
\newblock


\bibitem[Doty(2014)]%
        {doty2014art}
\bibfield{author}{\bibinfo{person}{Mark Doty}.} \bibinfo{year}{2014}\natexlab{}.
\newblock \bibinfo{booktitle}{\emph{The art of description: World into word}}.
\newblock \bibinfo{publisher}{Graywolf Press}.
\newblock


\bibitem[Fishelov(1990)]%
        {fishelov1990types}
\bibfield{author}{\bibinfo{person}{David Fishelov}.} \bibinfo{year}{1990}\natexlab{}.
\newblock \showarticletitle{Types of character, characteristics of types}.
\newblock \bibinfo{journal}{\emph{Style}} (\bibinfo{year}{1990}), \bibinfo{pages}{422--439}.
\newblock


\bibitem[Flower and Hayes(1981)]%
        {flower1981cognitive}
\bibfield{author}{\bibinfo{person}{Linda Flower} {and} \bibinfo{person}{John~R Hayes}.} \bibinfo{year}{1981}\natexlab{}.
\newblock \showarticletitle{A cognitive process theory of writing}.
\newblock \bibinfo{journal}{\emph{College composition and communication}} \bibinfo{volume}{32}, \bibinfo{number}{4} (\bibinfo{year}{1981}), \bibinfo{pages}{365--387}.
\newblock


\bibitem[Forster(1927)]%
        {forster1927aspects}
\bibfield{author}{\bibinfo{person}{Edward~Morgan Forster}.} \bibinfo{year}{1927}\natexlab{}.
\newblock \bibinfo{booktitle}{\emph{Aspects of the Novel}}.
\newblock \bibinfo{publisher}{Harcourt, Brace}.
\newblock


\bibitem[Fountain(2012)]%
        {fountain2012cliches}
\bibfield{author}{\bibinfo{person}{Nigel Fountain}.} \bibinfo{year}{2012}\natexlab{}.
\newblock \bibinfo{booktitle}{\emph{Clich{\'e}s: Avoid them like the plague}}.
\newblock \bibinfo{publisher}{Michael O'Mara Books}.
\newblock


\bibitem[Friedman(1955)]%
        {friedman1955point}
\bibfield{author}{\bibinfo{person}{Norman Friedman}.} \bibinfo{year}{1955}\natexlab{}.
\newblock \showarticletitle{Point of view in fiction: the development of a critical concept}.
\newblock \bibinfo{journal}{\emph{PMlA}} \bibinfo{volume}{70}, \bibinfo{number}{5} (\bibinfo{year}{1955}), \bibinfo{pages}{1160--1184}.
\newblock


\bibitem[Gao et~al\mbox{.}(2023)]%
        {gao2023human}
\bibfield{author}{\bibinfo{person}{Mingqi Gao}, \bibinfo{person}{Jie Ruan}, \bibinfo{person}{Renliang Sun}, \bibinfo{person}{Xunjian Yin}, \bibinfo{person}{Shiping Yang}, {and} \bibinfo{person}{Xiaojun Wan}.} \bibinfo{year}{2023}\natexlab{}.
\newblock \showarticletitle{Human-like summarization evaluation with chatgpt}.
\newblock \bibinfo{journal}{\emph{arXiv preprint arXiv:2304.02554}} (\bibinfo{year}{2023}).
\newblock


\bibitem[Gero et~al\mbox{.}(2023)]%
        {gero2023social}
\bibfield{author}{\bibinfo{person}{Katy~Ilonka Gero}, \bibinfo{person}{Tao Long}, {and} \bibinfo{person}{Lydia~B Chilton}.} \bibinfo{year}{2023}\natexlab{}.
\newblock \showarticletitle{Social Dynamics of AI Support in Creative Writing}. In \bibinfo{booktitle}{\emph{Proceedings of the 2023 CHI Conference on Human Factors in Computing Systems}} (Hamburg, Germany) \emph{(\bibinfo{series}{CHI '23})}. \bibinfo{publisher}{Association for Computing Machinery}, \bibinfo{address}{New York, NY, USA}, Article \bibinfo{articleno}{245}, \bibinfo{numpages}{15}~pages.
\newblock
\showISBNx{9781450394215}
\urldef\tempurl%
\url{https://doi.org/10.1145/3544548.3580782}
\showDOI{\tempurl}


\bibitem[Goldfarb-Tarrant et~al\mbox{.}(2020)]%
        {goldfarb-tarrant-etal-2020-content}
\bibfield{author}{\bibinfo{person}{Seraphina Goldfarb-Tarrant}, \bibinfo{person}{Tuhin Chakrabarty}, \bibinfo{person}{Ralph Weischedel}, {and} \bibinfo{person}{Nanyun Peng}.} \bibinfo{year}{2020}\natexlab{}.
\newblock \showarticletitle{Content Planning for Neural Story Generation with Aristotelian Rescoring}. In \bibinfo{booktitle}{\emph{Proceedings of the 2020 Conference on Empirical Methods in Natural Language Processing (EMNLP)}}. \bibinfo{publisher}{Association for Computational Linguistics}, \bibinfo{address}{Online}, \bibinfo{pages}{4319--4338}.
\newblock
\urldef\tempurl%
\url{https://doi.org/10.18653/v1/2020.emnlp-main.351}
\showDOI{\tempurl}


\bibitem[Guilford(1967)]%
        {guilford1967nature}
\bibfield{author}{\bibinfo{person}{Joy~Paul Guilford}.} \bibinfo{year}{1967}\natexlab{}.
\newblock \showarticletitle{The nature of human intelligence.}
\newblock  (\bibinfo{year}{1967}).
\newblock


\bibitem[H{\"a}m{\"a}l{\"a}inen et~al\mbox{.}(2023)]%
        {hamalainen2023evaluating}
\bibfield{author}{\bibinfo{person}{Perttu H{\"a}m{\"a}l{\"a}inen}, \bibinfo{person}{Mikke Tavast}, {and} \bibinfo{person}{Anton Kunnari}.} \bibinfo{year}{2023}\natexlab{}.
\newblock \showarticletitle{Evaluating large language models in generating synthetic hci research data: a case study}. In \bibinfo{booktitle}{\emph{Proceedings of the 2023 CHI Conference on Human Factors in Computing Systems}}. \bibinfo{pages}{1--19}.
\newblock


\bibitem[Holland(2009)]%
        {holland2009literature}
\bibfield{author}{\bibinfo{person}{Norman~Norwood Holland}.} \bibinfo{year}{2009}\natexlab{}.
\newblock \bibinfo{booktitle}{\emph{Literature and the Brain}}.
\newblock \bibinfo{publisher}{PsyArt Foundation}.
\newblock


\bibitem[Ippolito et~al\mbox{.}(2022)]%
        {ippolito2022creative}
\bibfield{author}{\bibinfo{person}{Daphne Ippolito}, \bibinfo{person}{Ann Yuan}, \bibinfo{person}{Andy Coenen}, {and} \bibinfo{person}{Sehmon Burnam}.} \bibinfo{year}{2022}\natexlab{}.
\newblock \showarticletitle{Creative writing with an ai-powered writing assistant: Perspectives from professional writers}.
\newblock \bibinfo{journal}{\emph{arXiv preprint arXiv:2211.05030}} (\bibinfo{year}{2022}).
\newblock


\bibitem[Jameson(1991)]%
        {jameson1991postmodernism}
\bibfield{author}{\bibinfo{person}{Fredric Jameson}.} \bibinfo{year}{1991}\natexlab{}.
\newblock \bibinfo{booktitle}{\emph{Postmodernism, or, the cultural logic of late capitalism}}.
\newblock \bibinfo{publisher}{Duke university press}.
\newblock


\bibitem[Karpinska et~al\mbox{.}(2021)]%
        {karpinska-etal-2021-perils}
\bibfield{author}{\bibinfo{person}{Marzena Karpinska}, \bibinfo{person}{Nader Akoury}, {and} \bibinfo{person}{Mohit Iyyer}.} \bibinfo{year}{2021}\natexlab{}.
\newblock \showarticletitle{The Perils of Using {M}echanical {T}urk to Evaluate Open-Ended Text Generation}. In \bibinfo{booktitle}{\emph{Proceedings of the 2021 Conference on Empirical Methods in Natural Language Processing}}. \bibinfo{publisher}{Association for Computational Linguistics}, \bibinfo{address}{Online and Punta Cana, Dominican Republic}, \bibinfo{pages}{1265--1285}.
\newblock
\urldef\tempurl%
\url{https://doi.org/10.18653/v1/2021.emnlp-main.97}
\showDOI{\tempurl}


\bibitem[Kaufman et~al\mbox{.}(2008)]%
        {kaufman2008essentials}
\bibfield{author}{\bibinfo{person}{James~C Kaufman}, \bibinfo{person}{Jonathan~A Plucker}, {and} \bibinfo{person}{John Baer}.} \bibinfo{year}{2008}\natexlab{}.
\newblock \bibinfo{booktitle}{\emph{Essentials of creativity assessment}}.
\newblock \bibinfo{publisher}{John Wiley \& Sons}.
\newblock


\bibitem[Kittur et~al\mbox{.}(2013)]%
        {kittur2013future}
\bibfield{author}{\bibinfo{person}{Aniket Kittur}, \bibinfo{person}{Jeffrey~V Nickerson}, \bibinfo{person}{Michael Bernstein}, \bibinfo{person}{Elizabeth Gerber}, \bibinfo{person}{Aaron Shaw}, \bibinfo{person}{John Zimmerman}, \bibinfo{person}{Matt Lease}, {and} \bibinfo{person}{John Horton}.} \bibinfo{year}{2013}\natexlab{}.
\newblock \showarticletitle{The future of crowd work}. In \bibinfo{booktitle}{\emph{Proceedings of the 2013 conference on Computer supported cooperative work}}. \bibinfo{pages}{1301--1318}.
\newblock


\bibitem[Kochis(2007)]%
        {kochis2007baxter}
\bibfield{author}{\bibinfo{person}{Maria Kochis}.} \bibinfo{year}{2007}\natexlab{}.
\newblock \showarticletitle{Baxter, Charles. The Art of Subtext: Beyond Plot}.
\newblock \bibinfo{journal}{\emph{Library Journal}} \bibinfo{volume}{132}, \bibinfo{number}{14} (\bibinfo{year}{2007}), \bibinfo{pages}{135--136}.
\newblock


\bibitem[Laban et~al\mbox{.}(2023)]%
        {laban2023llms}
\bibfield{author}{\bibinfo{person}{Philippe Laban}, \bibinfo{person}{Wojciech Kry{\'s}ci{\'n}ski}, \bibinfo{person}{Divyansh Agarwal}, \bibinfo{person}{Alexander~R Fabbri}, \bibinfo{person}{Caiming Xiong}, \bibinfo{person}{Shafiq Joty}, {and} \bibinfo{person}{Chien-Sheng Wu}.} \bibinfo{year}{2023}\natexlab{}.
\newblock \showarticletitle{LLMs as Factual Reasoners: Insights from Existing Benchmarks and Beyond}.
\newblock \bibinfo{journal}{\emph{arXiv preprint arXiv:2305.14540}} (\bibinfo{year}{2023}).
\newblock


\bibitem[Lee et~al\mbox{.}(2022)]%
        {lee2022coauthor}
\bibfield{author}{\bibinfo{person}{Mina Lee}, \bibinfo{person}{Percy Liang}, {and} \bibinfo{person}{Qian Yang}.} \bibinfo{year}{2022}\natexlab{}.
\newblock \showarticletitle{CoAuthor: Designing a Human-AI Collaborative Writing Dataset for Exploring Language Model Capabilities}. In \bibinfo{booktitle}{\emph{Proceedings of the 2022 CHI Conference on Human Factors in Computing Systems}} (New Orleans, LA, USA) \emph{(\bibinfo{series}{CHI '22})}. \bibinfo{publisher}{Association for Computing Machinery}, \bibinfo{address}{New York, NY, USA}, Article \bibinfo{articleno}{388}, \bibinfo{numpages}{19}~pages.
\newblock
\showISBNx{9781450391573}
\urldef\tempurl%
\url{https://doi.org/10.1145/3491102.3502030}
\showDOI{\tempurl}


\bibitem[Liu et~al\mbox{.}(2023)]%
        {liu2023gpteval}
\bibfield{author}{\bibinfo{person}{Yang Liu}, \bibinfo{person}{Dan Iter}, \bibinfo{person}{Yichong Xu}, \bibinfo{person}{Shuohang Wang}, \bibinfo{person}{Ruochen Xu}, {and} \bibinfo{person}{Chenguang Zhu}.} \bibinfo{year}{2023}\natexlab{}.
\newblock \showarticletitle{Gpteval: Nlg evaluation using gpt-4 with better human alignment}.
\newblock \bibinfo{journal}{\emph{arXiv preprint arXiv:2303.16634}} (\bibinfo{year}{2023}).
\newblock


\bibitem[Madaan et~al\mbox{.}(2023)]%
        {madaan2023self}
\bibfield{author}{\bibinfo{person}{Aman Madaan}, \bibinfo{person}{Niket Tandon}, \bibinfo{person}{Prakhar Gupta}, \bibinfo{person}{Skyler Hallinan}, \bibinfo{person}{Luyu Gao}, \bibinfo{person}{Sarah Wiegreffe}, \bibinfo{person}{Uri Alon}, \bibinfo{person}{Nouha Dziri}, \bibinfo{person}{Shrimai Prabhumoye}, \bibinfo{person}{Yiming Yang}, {et~al\mbox{.}}} \bibinfo{year}{2023}\natexlab{}.
\newblock \showarticletitle{Self-refine: Iterative refinement with self-feedback}.
\newblock \bibinfo{journal}{\emph{arXiv preprint arXiv:2303.17651}} (\bibinfo{year}{2023}).
\newblock


\bibitem[Matell and Jacoby(1971)]%
        {doi:10.1177/001316447103100307}
\bibfield{author}{\bibinfo{person}{Michael~S. Matell} {and} \bibinfo{person}{Jacob Jacoby}.} \bibinfo{year}{1971}\natexlab{}.
\newblock \showarticletitle{Is There an Optimal Number of Alternatives for Likert Scale Items? Study I: Reliability and Validity}.
\newblock \bibinfo{journal}{\emph{Educational and Psychological Measurement}} \bibinfo{volume}{31}, \bibinfo{number}{3} (\bibinfo{year}{1971}), \bibinfo{pages}{657--674}.
\newblock
\urldef\tempurl%
\url{https://doi.org/10.1177/001316447103100307}
\showDOI{\tempurl}
\showeprint{https://doi.org/10.1177/001316447103100307}


\bibitem[Mayers(2007)]%
        {mayers2007re}
\bibfield{author}{\bibinfo{person}{Tim Mayers}.} \bibinfo{year}{2007}\natexlab{}.
\newblock \bibinfo{booktitle}{\emph{(Re) Writing craft: composition, creative writing, and the future of English studies}}.
\newblock \bibinfo{publisher}{University of Pittsburgh Pre}.
\newblock


\bibitem[McIntyre et~al\mbox{.}(2003)]%
        {mcintyre2003individual}
\bibfield{author}{\bibinfo{person}{Faye~S McIntyre}, \bibinfo{person}{Robert~E Hite}, {and} \bibinfo{person}{Mary~Kay Rickard}.} \bibinfo{year}{2003}\natexlab{}.
\newblock \showarticletitle{Individual characteristics and creativity in the marketing classroom: Exploratory insights}.
\newblock \bibinfo{journal}{\emph{Journal of Marketing Education}} \bibinfo{volume}{25}, \bibinfo{number}{2} (\bibinfo{year}{2003}), \bibinfo{pages}{143--149}.
\newblock


\bibitem[Mirowski et~al\mbox{.}(2023)]%
        {mirowski2023cowriting}
\bibfield{author}{\bibinfo{person}{Piotr Mirowski}, \bibinfo{person}{Kory~W. Mathewson}, \bibinfo{person}{Jaylen Pittman}, {and} \bibinfo{person}{Richard Evans}.} \bibinfo{year}{2023}\natexlab{}.
\newblock \showarticletitle{Co-Writing Screenplays and Theatre Scripts with Language Models: Evaluation by Industry Professionals}. In \bibinfo{booktitle}{\emph{Proceedings of the 2023 CHI Conference on Human Factors in Computing Systems}} (Hamburg, Germany) \emph{(\bibinfo{series}{CHI '23})}. \bibinfo{publisher}{Association for Computing Machinery}, \bibinfo{address}{New York, NY, USA}, Article \bibinfo{articleno}{355}, \bibinfo{numpages}{34}~pages.
\newblock
\showISBNx{9781450394215}
\urldef\tempurl%
\url{https://doi.org/10.1145/3544548.3581225}
\showDOI{\tempurl}


\bibitem[Murray(2012)]%
        {murray2012craft}
\bibfield{author}{\bibinfo{person}{Donald~M Murray}.} \bibinfo{year}{2012}\natexlab{}.
\newblock \bibinfo{booktitle}{\emph{The craft of revision}}.
\newblock \bibinfo{publisher}{Cengage Learning}.
\newblock


\bibitem[Nebeling et~al\mbox{.}(2016)]%
        {nebeling2016wearwrite}
\bibfield{author}{\bibinfo{person}{Michael Nebeling}, \bibinfo{person}{Alexandra To}, \bibinfo{person}{Anhong Guo}, \bibinfo{person}{Adrian~A de Freitas}, \bibinfo{person}{Jaime Teevan}, \bibinfo{person}{Steven~P Dow}, {and} \bibinfo{person}{Jeffrey~P Bigham}.} \bibinfo{year}{2016}\natexlab{}.
\newblock \showarticletitle{WearWrite: Crowd-assisted writing from smartwatches}. In \bibinfo{booktitle}{\emph{Proceedings of the 2016 CHI conference on human factors in computing systems}}. \bibinfo{pages}{3834--3846}.
\newblock


\bibitem[Nichols et~al\mbox{.}(2020)]%
        {10.1145/3424636.3426903}
\bibfield{author}{\bibinfo{person}{Eric Nichols}, \bibinfo{person}{Leo Gao}, {and} \bibinfo{person}{Randy Gomez}.} \bibinfo{year}{2020}\natexlab{}.
\newblock \showarticletitle{Collaborative Storytelling with Large-Scale Neural Language Models}. In \bibinfo{booktitle}{\emph{Proceedings of the 13th ACM SIGGRAPH Conference on Motion, Interaction and Games}} (Virtual Event, SC, USA) \emph{(\bibinfo{series}{MIG '20})}. \bibinfo{publisher}{Association for Computing Machinery}, \bibinfo{address}{New York, NY, USA}, Article \bibinfo{articleno}{17}, \bibinfo{numpages}{10}~pages.
\newblock
\showISBNx{9781450381710}
\urldef\tempurl%
\url{https://doi.org/10.1145/3424636.3426903}
\showDOI{\tempurl}


\bibitem[Nussbaum(1997)]%
        {nussbaum1997poetic}
\bibfield{author}{\bibinfo{person}{Martha Nussbaum}.} \bibinfo{year}{1997}\natexlab{}.
\newblock \bibinfo{booktitle}{\emph{Poetic justice: The literary imagination and public life}}.
\newblock \bibinfo{publisher}{Beacon Press}.
\newblock


\bibitem[OpenAI(2022)]%
        {ChatGPT}
\bibfield{author}{\bibinfo{person}{OpenAI}.} \bibinfo{year}{2022}\natexlab{}.
\newblock \showarticletitle{ChatGT: Optimizing language models for dialogue.}
\newblock  (\bibinfo{year}{2022}).
\newblock
\urldef\tempurl%
\url{https://openai.com/blog/chatgpt/}
\showURL{%
\tempurl}


\bibitem[OpenAI(2023)]%
        {OpenAI2023GPT4TR}
\bibfield{author}{\bibinfo{person}{OpenAI}.} \bibinfo{year}{2023}\natexlab{}.
\newblock \showarticletitle{GPT-4 Technical Report}.
\newblock \bibinfo{journal}{\emph{ArXiv}}  \bibinfo{volume}{abs/2303.08774} (\bibinfo{year}{2023}).
\newblock


\bibitem[Phelan(1996)]%
        {phelan1996narrative}
\bibfield{author}{\bibinfo{person}{James Phelan}.} \bibinfo{year}{1996}\natexlab{}.
\newblock \bibinfo{booktitle}{\emph{Narrative as rhetoric: Technique, audiences, ethics, ideology}}.
\newblock \bibinfo{publisher}{Ohio State University Press}.
\newblock


\bibitem[Plucker et~al\mbox{.}(2010)]%
        {plucker2010assessment}
\bibfield{author}{\bibinfo{person}{Jonathan~A Plucker}, \bibinfo{person}{Matthew~C Makel}, {and} \bibinfo{person}{Meihua Qian}.} \bibinfo{year}{2010}\natexlab{}.
\newblock \showarticletitle{Assessment of creativity}.
\newblock \bibinfo{journal}{\emph{The Cambridge handbook of creativity}} (\bibinfo{year}{2010}), \bibinfo{pages}{48--73}.
\newblock


\bibitem[Rajani et~al\mbox{.}(2023)]%
        {rajani2023llm_labels}
\bibfield{author}{\bibinfo{person}{Nazneen Rajani}, \bibinfo{person}{Nathan Lambert}, \bibinfo{person}{Sheon Han}, \bibinfo{person}{Jean Wang}, \bibinfo{person}{Osvald Nitski}, \bibinfo{person}{Edward Beeching}, {and} \bibinfo{person}{Lewis Tunstall}.} \bibinfo{year}{2023}\natexlab{}.
\newblock \showarticletitle{Can foundation models label data like humans?}
\newblock \bibinfo{journal}{\emph{Hugging Face Blog}} (\bibinfo{year}{2023}).
\newblock
\newblock
\shownote{https://huggingface.co/blog/llm-leaderboard}.


\bibitem[Rashkin et~al\mbox{.}(2020)]%
        {rashkin-etal-2020-plotmachines}
\bibfield{author}{\bibinfo{person}{Hannah Rashkin}, \bibinfo{person}{Asli Celikyilmaz}, \bibinfo{person}{Yejin Choi}, {and} \bibinfo{person}{Jianfeng Gao}.} \bibinfo{year}{2020}\natexlab{}.
\newblock \showarticletitle{{P}lot{M}achines: Outline-Conditioned Generation with Dynamic Plot State Tracking}. In \bibinfo{booktitle}{\emph{Proceedings of the 2020 Conference on Empirical Methods in Natural Language Processing (EMNLP)}}. \bibinfo{publisher}{Association for Computational Linguistics}, \bibinfo{address}{Online}, \bibinfo{pages}{4274--4295}.
\newblock
\urldef\tempurl%
\url{https://doi.org/10.18653/v1/2020.emnlp-main.349}
\showDOI{\tempurl}


\bibitem[Rodr{\'\i}guez(2008)]%
        {rodriguez2008problem}
\bibfield{author}{\bibinfo{person}{Alicia Rodr{\'\i}guez}.} \bibinfo{year}{2008}\natexlab{}.
\newblock \showarticletitle{The ‘problem’of creative writing: using grading rubrics based on narrative theory as solution}.
\newblock \bibinfo{journal}{\emph{New Writing}} \bibinfo{volume}{5}, \bibinfo{number}{3} (\bibinfo{year}{2008}), \bibinfo{pages}{167--177}.
\newblock


\bibitem[Roemmele and Gordon(2018a)]%
        {roemmele-gordon-2018-linguistic}
\bibfield{author}{\bibinfo{person}{Melissa Roemmele} {and} \bibinfo{person}{Andrew Gordon}.} \bibinfo{year}{2018}\natexlab{a}.
\newblock \showarticletitle{Linguistic Features of Helpfulness in Automated Support for Creative Writing}. In \bibinfo{booktitle}{\emph{Proceedings of the First Workshop on Storytelling}}. \bibinfo{publisher}{Association for Computational Linguistics}, \bibinfo{address}{New Orleans, Louisiana}, \bibinfo{pages}{14--19}.
\newblock
\urldef\tempurl%
\url{https://doi.org/10.18653/v1/W18-1502}
\showDOI{\tempurl}


\bibitem[Roemmele and Gordon(2018b)]%
        {roemmele2018automated}
\bibfield{author}{\bibinfo{person}{Melissa Roemmele} {and} \bibinfo{person}{Andrew~S Gordon}.} \bibinfo{year}{2018}\natexlab{b}.
\newblock \showarticletitle{Automated assistance for creative writing with an rnn language model}. In \bibinfo{booktitle}{\emph{Proceedings of the 23rd international conference on intelligent user interfaces companion}}. \bibinfo{pages}{1--2}.
\newblock


\bibitem[Silvia et~al\mbox{.}(2008)]%
        {silvia2008assessing}
\bibfield{author}{\bibinfo{person}{Paul~J Silvia}, \bibinfo{person}{Beate~P Winterstein}, \bibinfo{person}{John~T Willse}, \bibinfo{person}{Christopher~M Barona}, \bibinfo{person}{Joshua~T Cram}, \bibinfo{person}{Karl~I Hess}, \bibinfo{person}{Jenna~L Martinez}, {and} \bibinfo{person}{Crystal~A Richard}.} \bibinfo{year}{2008}\natexlab{}.
\newblock \showarticletitle{Assessing creativity with divergent thinking tasks: exploring the reliability and validity of new subjective scoring methods.}
\newblock \bibinfo{journal}{\emph{Psychology of Aesthetics, Creativity, and the Arts}} \bibinfo{volume}{2}, \bibinfo{number}{2} (\bibinfo{year}{2008}), \bibinfo{pages}{68}.
\newblock


\bibitem[Stiennon et~al\mbox{.}(2020)]%
        {stiennon2020learning}
\bibfield{author}{\bibinfo{person}{Nisan Stiennon}, \bibinfo{person}{Long Ouyang}, \bibinfo{person}{Jeffrey Wu}, \bibinfo{person}{Daniel Ziegler}, \bibinfo{person}{Ryan Lowe}, \bibinfo{person}{Chelsea Voss}, \bibinfo{person}{Alec Radford}, \bibinfo{person}{Dario Amodei}, {and} \bibinfo{person}{Paul~F Christiano}.} \bibinfo{year}{2020}\natexlab{}.
\newblock \showarticletitle{Learning to summarize with human feedback}.
\newblock \bibinfo{journal}{\emph{Advances in Neural Information Processing Systems}}  \bibinfo{volume}{33} (\bibinfo{year}{2020}), \bibinfo{pages}{3008--3021}.
\newblock


\bibitem[Thomas(2006)]%
        {thomas2006general}
\bibfield{author}{\bibinfo{person}{David~R Thomas}.} \bibinfo{year}{2006}\natexlab{}.
\newblock \showarticletitle{A general inductive approach for analyzing qualitative evaluation data}.
\newblock \bibinfo{journal}{\emph{American journal of evaluation}} \bibinfo{volume}{27}, \bibinfo{number}{2} (\bibinfo{year}{2006}), \bibinfo{pages}{237--246}.
\newblock


\bibitem[Torrance(1966)]%
        {torrance1966torrance}
\bibfield{author}{\bibinfo{person}{Ellis~Paul Torrance}.} \bibinfo{year}{1966}\natexlab{}.
\newblock \bibinfo{booktitle}{\emph{Torrance tests of creative thinking: Norms-technical manual: Verbal tests, forms a and b: Figural tests, forms a and b}}.
\newblock \bibinfo{publisher}{Personal Press, Incorporated}.
\newblock


\bibitem[Trisnayanti et~al\mbox{.}(2019)]%
        {trisnayanti2019development}
\bibfield{author}{\bibinfo{person}{Y Trisnayanti}, \bibinfo{person}{A Khoiri}, \bibinfo{person}{Miterianifa Miterianifa}, {and} \bibinfo{person}{HD Ayu}.} \bibinfo{year}{2019}\natexlab{}.
\newblock \showarticletitle{Development of Torrance test creativity thinking (TTCT) instrument in science learning}. In \bibinfo{booktitle}{\emph{AIP Conference Proceedings}}, Vol.~\bibinfo{volume}{2194}. AIP Publishing.
\newblock


\bibitem[Vaezi and Rezaei(2019)]%
        {vaezi2019development}
\bibfield{author}{\bibinfo{person}{Maryam Vaezi} {and} \bibinfo{person}{Saeed Rezaei}.} \bibinfo{year}{2019}\natexlab{}.
\newblock \showarticletitle{Development of a rubric for evaluating creative writing: a multi-phase research}.
\newblock \bibinfo{journal}{\emph{New Writing}} \bibinfo{volume}{16}, \bibinfo{number}{3} (\bibinfo{year}{2019}), \bibinfo{pages}{303--317}.
\newblock


\bibitem[Veselovsky et~al\mbox{.}(2023)]%
        {veselovsky2023artificial}
\bibfield{author}{\bibinfo{person}{Veniamin Veselovsky}, \bibinfo{person}{Manoel~Horta Ribeiro}, {and} \bibinfo{person}{Robert West}.} \bibinfo{year}{2023}\natexlab{}.
\newblock \showarticletitle{Artificial Artificial Artificial Intelligence: Crowd Workers Widely Use Large Language Models for Text Production Tasks}.
\newblock \bibinfo{journal}{\emph{arXiv preprint arXiv:2306.07899}} (\bibinfo{year}{2023}).
\newblock


\bibitem[Viswanathan et~al\mbox{.}(2023)]%
        {viswanathan2023large}
\bibfield{author}{\bibinfo{person}{Vijay Viswanathan}, \bibinfo{person}{Kiril Gashteovski}, \bibinfo{person}{Carolin Lawrence}, \bibinfo{person}{Tongshuang Wu}, {and} \bibinfo{person}{Graham Neubig}.} \bibinfo{year}{2023}\natexlab{}.
\newblock \showarticletitle{Large Language Models Enable Few-Shot Clustering}.
\newblock \bibinfo{journal}{\emph{arXiv preprint arXiv:2307.00524}} (\bibinfo{year}{2023}).
\newblock


\bibitem[Ward et~al\mbox{.}(1999)]%
        {ward1999creative}
\bibfield{author}{\bibinfo{person}{Thomas~B Ward}, \bibinfo{person}{Steven~M Smith}, {and} \bibinfo{person}{Ronald~A Finke}.} \bibinfo{year}{1999}\natexlab{}.
\newblock \showarticletitle{Creative cognition}.
\newblock \bibinfo{journal}{\emph{Handbook of creativity}}  \bibinfo{volume}{189} (\bibinfo{year}{1999}), \bibinfo{pages}{212}.
\newblock


\bibitem[Wei et~al\mbox{.}(2022)]%
        {wei2022chain}
\bibfield{author}{\bibinfo{person}{Jason Wei}, \bibinfo{person}{Xuezhi Wang}, \bibinfo{person}{Dale Schuurmans}, \bibinfo{person}{Maarten Bosma}, \bibinfo{person}{Fei Xia}, \bibinfo{person}{Ed Chi}, \bibinfo{person}{Quoc~V Le}, \bibinfo{person}{Denny Zhou}, {et~al\mbox{.}}} \bibinfo{year}{2022}\natexlab{}.
\newblock \showarticletitle{Chain-of-thought prompting elicits reasoning in large language models}.
\newblock \bibinfo{journal}{\emph{Advances in Neural Information Processing Systems}}  \bibinfo{volume}{35} (\bibinfo{year}{2022}), \bibinfo{pages}{24824--24837}.
\newblock


\bibitem[Weigle(2002)]%
        {weigle2002assessing}
\bibfield{author}{\bibinfo{person}{Sara~Cushing Weigle}.} \bibinfo{year}{2002}\natexlab{}.
\newblock \bibinfo{booktitle}{\emph{Assessing writing}}.
\newblock \bibinfo{publisher}{Cambridge University Press}.
\newblock


\bibitem[Yang et~al\mbox{.}(2023)]%
        {yang2022doc}
\bibfield{author}{\bibinfo{person}{Kevin Yang}, \bibinfo{person}{Dan Klein}, \bibinfo{person}{Nanyun Peng}, {and} \bibinfo{person}{Yuandong Tian}.} \bibinfo{year}{2023}\natexlab{}.
\newblock \showarticletitle{{DOC}: Improving Long Story Coherence With Detailed Outline Control}. In \bibinfo{booktitle}{\emph{Proceedings of the 61st Annual Meeting of the Association for Computational Linguistics (Volume 1: Long Papers)}}. \bibinfo{publisher}{Association for Computational Linguistics}, \bibinfo{address}{Toronto, Canada}, \bibinfo{pages}{3378--3465}.
\newblock
\urldef\tempurl%
\url{https://doi.org/10.18653/v1/2023.acl-long.190}
\showDOI{\tempurl}


\bibitem[Yang et~al\mbox{.}(2022)]%
        {yang2022re3}
\bibfield{author}{\bibinfo{person}{Kevin Yang}, \bibinfo{person}{Yuandong Tian}, \bibinfo{person}{Nanyun Peng}, {and} \bibinfo{person}{Dan Klein}.} \bibinfo{year}{2022}\natexlab{}.
\newblock \showarticletitle{Re3: Generating Longer Stories With Recursive Reprompting and Revision}. In \bibinfo{booktitle}{\emph{Proceedings of the 2022 Conference on Empirical Methods in Natural Language Processing}}. \bibinfo{publisher}{Association for Computational Linguistics}, \bibinfo{address}{Abu Dhabi, United Arab Emirates}, \bibinfo{pages}{4393--4479}.
\newblock
\urldef\tempurl%
\url{https://doi.org/10.18653/v1/2022.emnlp-main.296}
\showDOI{\tempurl}


\bibitem[Yao et~al\mbox{.}(2019)]%
        {10.1609/aaai.v33i01.33017378}
\bibfield{author}{\bibinfo{person}{Lili Yao}, \bibinfo{person}{Nanyun Peng}, \bibinfo{person}{Ralph Weischedel}, \bibinfo{person}{Kevin Knight}, \bibinfo{person}{Dongyan Zhao}, {and} \bibinfo{person}{Rui Yan}.} \bibinfo{year}{2019}\natexlab{}.
\newblock \showarticletitle{Plan-and-Write: Towards Better Automatic Storytelling}. In \bibinfo{booktitle}{\emph{Proceedings of the Thirty-Third AAAI Conference on Artificial Intelligence and Thirty-First Innovative Applications of Artificial Intelligence Conference and Ninth AAAI Symposium on Educational Advances in Artificial Intelligence}} (Honolulu, Hawaii, USA) \emph{(\bibinfo{series}{AAAI'19/IAAI'19/EAAI'19})}. \bibinfo{publisher}{AAAI Press}, Article \bibinfo{articleno}{906}, \bibinfo{numpages}{8}~pages.
\newblock
\showISBNx{978-1-57735-809-1}
\urldef\tempurl%
\url{https://doi.org/10.1609/aaai.v33i01.33017378}
\showDOI{\tempurl}


\bibitem[Yuan et~al\mbox{.}(2022)]%
        {yuan2022wordcraft}
\bibfield{author}{\bibinfo{person}{Ann Yuan}, \bibinfo{person}{Andy Coenen}, \bibinfo{person}{Emily Reif}, {and} \bibinfo{person}{Daphne Ippolito}.} \bibinfo{year}{2022}\natexlab{}.
\newblock \showarticletitle{Wordcraft: story writing with large language models}. In \bibinfo{booktitle}{\emph{27th International Conference on Intelligent User Interfaces}}. \bibinfo{pages}{841--852}.
\newblock


\bibitem[Zhou et~al\mbox{.}(2022)]%
        {zhou2022large}
\bibfield{author}{\bibinfo{person}{Yongchao Zhou}, \bibinfo{person}{Andrei~Ioan Muresanu}, \bibinfo{person}{Ziwen Han}, \bibinfo{person}{Keiran Paster}, \bibinfo{person}{Silviu Pitis}, \bibinfo{person}{Harris Chan}, {and} \bibinfo{person}{Jimmy Ba}.} \bibinfo{year}{2022}\natexlab{}.
\newblock \showarticletitle{Large language models are human-level prompt engineers}.
\newblock \bibinfo{journal}{\emph{arXiv preprint arXiv:2211.01910}} (\bibinfo{year}{2022}).
\newblock


\end{thebibliography}

\appendix % Added this so that Appendix Sections are denoted as A.1, A.2, etc.

\newpage
\newpage

\section{Appendix} \label{appendix}

\subsection{NewYorker Data for evaluation}

\begin{figure}[!ht]
\small
\centering
\includegraphics[width=0.4\textwidth]{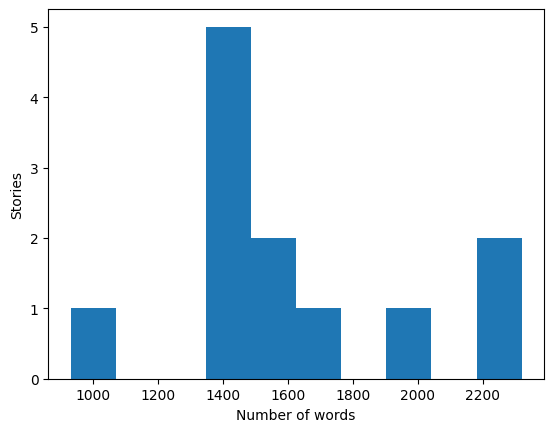}
\caption{\label{lengthdist} Distribution of word count of stories in our test set}
\end{figure}

Table \ref{teststories} shows the data used for conducting our evaluation. The 12 stories shown are taken from The New Yorker and summarized into single-sentence plots. These stories come from highly established literary experts acting as an upper bound for what it means to be creative. These stories span complex themes.

\begin{table*}[!ht]
\centering
\small
\def\arraystretch{1.35}
\begin{tabular}{|l|}
\hline
\begin{tabular}[c]{@{}l@{}}Write a New Yorker-style story given the plot below. Make sure it is atleast \textbf{\color{blue}\{\{word\_count\}\}} words. Directly start with the\\ story, do not say things like `Here's the story {[}...{]}:\end{tabular}                                                                                                                                                                                            \\ \hline\hline
\begin{tabular}[c]{@{}l@{}}You wrote the story I gave you below. I requested a story with \textbf{\color{blue}\{\{word\_count\}\}} words, but the story only has\\ \textbf{\color{blue}\{\{current\_word\_count\}\}} words. Can you rewrite the story to make it longer, and closer to the \textbf{\color{blue}\{\{word\_count\}\}} word target\\ I gave you. Directly start with the story, do not say things like `Here's the story {[}...{]}:`\\ \\ Current story: \{\{story\}\}\end{tabular} \\ \hline
\end{tabular}
\vspace{2ex}
\caption{\label{promptstory}Prompt to write the initial story (Row1) vs Prompt to rewrite the initial story to be longer. word\_count represents the number of words in the human written story on a given plot (P) while current\_word\_count represents the number of words in the LLM generated story on the same plot (P)}
\end{table*}

\begin{table*}[!ht]
\def\arraystretch{1.15}
\small
\begin{tabular}{|l|l|}
\hline
Story                                    & Plot                                                                                                                                                                                                                                                                                                                                                                                                                                                                                                                                   \\ \hline
\href{https://www.newyorker.com/books/flash-fiction/a-triangle}{A Triangle}                               & \begin{tabular}[c]{@{}l@{}}An observer becomes entranced by a seemingly ordinary couple on the street, follows them home, and then \\watches them from outside in the rising floodwaters, drawing an eerie connection between the woman and\\ a discarded, burned chair they'd noticed earlier.\end{tabular}                                                                                                                                                                    \\ \hline\hline
\href{https://www.newyorker.com/books/flash-fiction/barbara-detroit-1966}{\begin{tabular}[c]{@{}l@{}}Barbara\\ Detroit,1966\end{tabular}}                    & \begin{tabular}[c]{@{}l@{}}On Feb 12, 1966, a heavily pregnant woman named Barbara experienced a shocking incident in her synagogue\\in Southfield, Detroit, where a young man shot and killed the renowned Rabbi Adler before turning the gun\\ on himself, and though Barbara tried to reach the shooter, she was swept away by the fleeing crowd.\end{tabular}                                                                              \\ \hline\hline
\href{https://www.newyorker.com/books/flash-fiction/beyond-nature}{Beyond Nature}                           & \begin{tabular}[c]{@{}l@{}}A solitary man walking in a remote mountainous region comes across a car crash, and stays by the side\\ of the lifeless female victim, narrating stories of his past and reflecting on the impermanence of \\events and life itself, while awaiting emergency services amidst the looming presence of wilderness.\end{tabular}                                                                                                                \\ \hline\hline
\href{https://www.newyorker.com/books/flash-fiction/certain-european-movies}{\begin{tabular}[c]{@{}l@{}}Certain European\\ Movies\end{tabular}}                  & \begin{tabular}[c]{@{}l@{}}Two individuals, at a residency together, navigate the complexity of their ephemeral relationship during\\ their final beach trip, framed by misadventures, subtle tensions, unspoken desires, and looming departures.\end{tabular}                                                                                                                                                                                   \\ \hline\hline
\href{https://www.newyorker.com/books/flash-fiction/keys}{Keys}                                     & \begin{tabular}[c]{@{}l@{}}Daniel, struggling with recurring dreams of his ex-wife Rachel and a mysterious unused flat, eventually \\discusses them with his current partner Isabel, sparking various reflections and conversations about their\\ past relationships, until a real-life discovery of old keys triggers a nostalgic memory and helps him find a\\ way to reconnect with his present relationship through canoeing.\end{tabular}                                     \\ \hline\hline
\href{https://www.newyorker.com/books/flash-fiction/listening-for-the-click}{\begin{tabular}[c]{@{}l@{}}Listening For\\ the Click\end{tabular}}                  & \begin{tabular}[c]{@{}l@{}}Navigating a complex social landscape, the protagonist experiences a series of complex relationships \\and emotional turmoil in a student environment, and engages in self-discovery and self-reflection as she\\ interacts with the characters Carl, Martin, Lizzy, and Johan, resulting in a journey of introspection,\\ betrayal, love, and personal growth.\end{tabular}                                                          \\ \hline\hline
\href{https://www.newyorker.com/magazine/2023/05/15/maintenance-hvidovre-fiction-olga-ravn}{\begin{tabular}[c]{@{}l@{}}Maintenance,\\ Hvidovre\end{tabular}}                   & \begin{tabular}[c]{@{}l@{}}A woman experiences a disorienting night in a maternity ward where she encounters other similarly \\disoriented new mothers, leading to an uncanny mix-up where she leaves the hospital with a baby \\that she realizes is not her own, yet accepts the situation with an inexplicable sense of happiness.\end{tabular}                                                                                                  \\ \hline\hline
\href{https://www.newyorker.com/magazine/2022/11/14/returns}{Returns}                                  & \begin{tabular}[c]{@{}l@{}}The narrator visits their elderly mother in her small town, spending a day with her that is filled with \\nostalgia, conversation, and old habits, only to return a month later after her hospitalization due to\\ a sunstroke, finding remnants of their last visit.\end{tabular}                                                                                                                                                                      \\ \hline\hline
\href{https://www.newyorker.com/books/flash-fiction/the-facade-renovation-thats-going-well}{\begin{tabular}[c]{@{}l@{}}The Facade \\Renovation\\ That’s Going Well\end{tabular}} & \begin{tabular}[c]{@{}l@{}}An academic faculty housed in a building with a critical waterproofing layer missing experiences a series\\ of disruptive and problematic construction repairs, causing tension, inconvenience, and health concerns\\ among the tenants, ultimately leading to resignation and endurance in hopes of better future circumstances.\end{tabular}                                                        \\ \hline\hline
\href{https://www.newyorker.com/books/flash-fiction/the-kingdom-that-failed}{\begin{tabular}[c]{@{}l@{}}The Kingdom\\ That Failed\end{tabular}}                  & \begin{tabular}[c]{@{}l@{}}The narrator recounts their college friendship with the seemingly flawless Q, and after a decade apart, \\they accidentally cross paths at a pool, where the narrator anonymously observes Q's failed attempt to \\let down a woman about a work-related issue, demonstrating that Q, too, has his share of difficulties.\end{tabular}                                                                                                \\ \hline\hline
\href{https://www.newyorker.com/magazine/2022/06/13/trash }{Trash}                                    & \begin{tabular}[c]{@{}l@{}}A woman unexpectedly marries the son of a successful, ambitious woman named Miss Emily, finding both \\acceptance and critique from her mother-in-law as she navigates this new relationship and confronts the \\stark contrasts between her former life as a supermarket cashier and her new life as part of a well-off family.\end{tabular}                                                                                                            \\ \hline\hline
\href{https://www.newyorker.com/culture/personal-history/the-last-dance-with-my-dad}{\begin{tabular}[c]{@{}l@{}}The Last Dance\\ with my Dad \end{tabular}}               & \begin{tabular}[c]{@{}l@{}}A young teenager recounts her experiences of fitting into her father's gay lifestyle, highlighted by a\\ seven-day cruise with hundreds of gay men, where she experienced acceptance and connection, had her\\ first genuine interaction with a  boy, and shared a last dance with her terminally ill father.\end{tabular}                                                                                                       \\ \hline
\end{tabular}
\vspace{2ex}
\caption{\label{teststories} Expert-written short stories from the New Yorker along with their human-verified GPT4 generated summary as plots that are included as part of our test data for Creativity Evaluation}
\end{table*}

\subsection{Expert Perception on the TTCW tests}

\begin{figure*}[!ht]
    \centering
     \includegraphics[width=\textwidth]{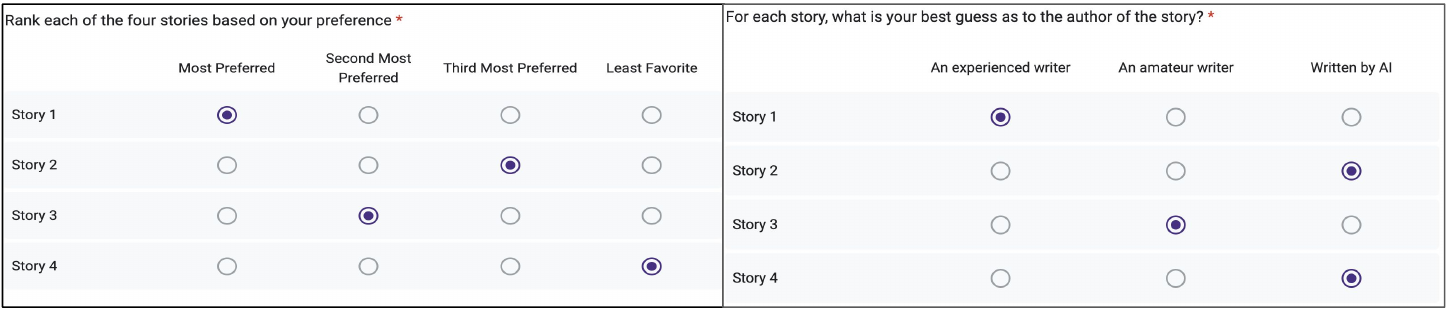}
    \caption{\label{relev} Relative Evaluation by Creative Writing Experts within a given group of four stories}
\end{figure*}

\begin{table*}[!ht]
\small
\centering
\begin{tabular}{|l|l|}
\hline
E5 & \begin{tabular}[c]{@{}l@{}}It was a pretty effective rubric! I'm used to being more subjective in my work -- did you like a story? Did it connect with \\you? Did it make sense? Why or why not? It was often challenging to break it down into more regimented segments \\like the rubric asked for -- but I do think that it allowed me to express the subjective feelings in a pretty thorough and\\ structured way!\end{tabular}                                                                                                                                                                 \\ \hline
E3 & \begin{tabular}[c]{@{}l@{}}As for the rubric, I thought it was quite thorough. There were some categories where I would say the story didn’t ``pass,"\\ but which were excellent. This happened often with the categories about multiple points of view, and innovative\\ structure and form. Overall, I think the rubric was helpful in helping me think about the different aspects of storytelling.\end{tabular}                                                                                                                                                                                 \\ \hline
E4 & \begin{tabular}[c]{@{}l@{}}I thought the rubric felt pretty thorough; the only part I felt could be added was that suggestion about consistency in\\ voice \& diction!\end{tabular}                                                                                                                                                         \\ \hline
E2 & \begin{tabular}[c]{@{}l@{}}The rubric seemed great to me! It’s however hard to talk about something like pacing without talking about scene and \\summary, for instance. Or the difference between originality of thought and originality in theme/content—wouldn’t the \\latter make up the former and vice/versa? But it is also comprehensive and I can see the merits of this sort of repetition in\\ teasing out a fuller picture of things\end{tabular} \\ \hline
E1 & \begin{tabular}[c]{@{}l@{}}I thought the rubric was pretty good tbh. I think there is overlap in some of the different elements, like "language \\proficiency \& literary devices" and "originality in thought." it's tricky to use words like "satisfying" and "sophisticated" \\when assessing art, but there's always going to be a great deal of subjectivity in these matters.I think that voice is a crucial \\aspect of high-quality writing that is being overlooked by the rubric, and one that greatly informs how I as a reader\\ evaluate 
and appreciate literary writing.\end{tabular} \\ \hline
\end{tabular}
\vspace{2ex}
\caption{\label{expertfeedbackrubric}Expert perception and feedback on the TTCW tests they conducted as part of our data collection.}
\end{table*}

Since the experts listed in Table \ref{creativeexperts} were not involved in designing the rubric but evaluated several stories based on the rubric we asked them their \textit{overall thought about the rubric and any potentially crucial test we missed out on that they use to discriminate between good and bad writing}.As can be seen in Table \ref{expertfeedbackrubric} in Appendix overall almost every expert agreed on the thorough and effective nature of our rubric. Many of them agreed on the fact that our rubric helped them to think about different aspects of storytelling in a more structured way. One of the difficult things about coming up with a rubric for creativity is ensuring coverage. Even though our rubric covers most aspects of creative writing, some experts such as E1 and E4 emphasized on the utility of \textbf{Consistency of Voice and Diction} as a measurable test. In E4's words \textit{``Inconsistent voice and diction are sometimes/often notable in stories that aren't very good, and when voice \& diction are used beautifully, it enhances a story considerably"}. E1 similarly exclaimed \textit{``One of the most meaningful aspects of high-quality literary writing is voice, which conveys qualities of proficiency, artistry, personality, and identity."}. We hope future work can adapt this as a meaningful test in addition to the tests covered in our rubric. Finally, some of the tests from our rubric can have potential overlaps as pointed out by E2. This is further corroborated by the similar numbers for \textit{Narrative Pacing} and \textit{Scenes vs Exposition} suggesting a strong correlation between the two.
\begin{table*}[!ht]
\small
\centering
\begin{tabular}{|l|l|l|}
\hline
Test & Passing Stories & Failing Stories \\ \hline
\begin{tabular}[c]{@{}l@{}}Originality in\\ Form\end{tabular} & \begin{tabular}[c]{@{}l@{}}Inventive techniques like time jumping, varied \\ perspectives, unconventional punctuation, and\\ delayed revelation of key information\end{tabular} & \begin{tabular}[c]{@{}l@{}}Conventional and linear in its form, language, \\ and narrative, with occasional attempts at \\ innovation that do not significantly contribute to \\ its overall originality or creativity\end{tabular} \\ \hline
\begin{tabular}[c]{@{}l@{}}Originality in\\ Thought\end{tabular} & \begin{tabular}[c]{@{}l@{}}Fresh language, unique plot and characters, subtle\\ emotional resonance, and inventive metaphors. Minor \\ familiar elements, but do not undermine the overall \\ sense of imagination and distinctiveness\end{tabular} & \begin{tabular}[c]{@{}l@{}}Stories relies heavily on cliches \& tired tropes.\\ Language does not feel fresh or original with \\ narrative arc following a predictable trajectory.\\ Metaphors, descriptions, and overall premise \\ cover familiar ground that lacks novelty or nuance\end{tabular} \\ \hline
\begin{tabular}[c]{@{}l@{}}Originality in\\ Theme/Content\end{tabular} & \begin{tabular}[c]{@{}l@{}}Unconventional, dreamlike exploration of emotions\\ such as love and loss, evoking empathy and reflection\\ through its distinct main character perspective, \\ eschewing simplistic meanings for ambiguity, and \\ valuing open-ended questions over singular messages,\\ thus providing a unique reading experience compared\\ to conventional stories.\end{tabular} & \begin{tabular}[c]{@{}l@{}}Disjointed narrative, underdeveloped themes, \\ inconsistent tone, vaguely defined characters, and\\ abrupt context shifts, lack depth and fail to provide \\ substantive insight or originality to the reader.\end{tabular} \\ \hline\hline
\begin{tabular}[c]{@{}l@{}}World Building\\ and Setting\end{tabular} & \begin{tabular}[c]{@{}l@{}}Strategic use of concrete, specific sensory details from\\ a particular character’s perspective balances narrative\\ momentum, making a fictional world feel real, vivid\\ and immersive for readers. Thoughtful depiction of\\ everyday objects, and idiosyncratic elements within\\ narrative and dialogue to balance exposition with \\ vivid scene-setting, creating authenticity and realism \\ that serves the plot and characters\end{tabular} & \begin{tabular}[c]{@{}l@{}}Fictional world is not always convincingly \\established through sensory details and language. \\Stories rely too heavily on overwrought imagery\\ and figurative language without grounding \\the reader in a tangible reality.\end{tabular} \\ \hline
\begin{tabular}[c]{@{}l@{}}Character\\ Development\end{tabular} & \begin{tabular}[c]{@{}l@{}}Fully realized characters with contradictions, \\ motivations, and backstories that make them\\ feel lifelike. Flatter, less developed characters\\ that feel appropriate for the narrative goals \\ and style is not necessarily a weakness\end{tabular} & \begin{tabular}[c]{@{}l@{}}Characters not well rounded. easily resorting to \\stereotypes. Predictable arcs not making them\\memorable. Actions or motivations unclear leading \\to disconnect\end{tabular} \\ \hline
\begin{tabular}[c]{@{}l@{}}Rhetorical\\ Complexity\end{tabular} & \begin{tabular}[c]{@{}l@{}}Rich subtext that emerges through contrasts between\\ characters and settings. Omissions that let readers \\ fill in meaning, metaphors with layered significance, \\ implicit characterization, ambiguous endings, \\ intertextual references, and rich language that goes\\ beyond straightforward storytelling\end{tabular} & \begin{tabular}[c]{@{}l@{}}Straightforward storytelling, lack of complexity,\\ nuance, and ambiguity.The absence of context and \\ ironic tensions, overtelling instead of showing, \\ Use of expository dialogue, blunt symbolism, \\ predictable characters, and definitive endings, all \\of which limit interpretation and undermine\\ their potential resonance\end{tabular} \\ \hline
\end{tabular}
\vspace{2ex}
\caption{\label{expertexpl2}Common themes and issues found in expert explanations for tests focusing on TTCW-Originality and TTCW-Elaboration}
\vspace{-8ex}
\end{table*}

\subsection{Common themes in TTCW of Originality and Elaboration}

Table~\ref{expertexpl2} provides common themes and issues found in the expert explanations for each TTCW in the Originality and Elaboration dimensions.

\subsection{Example LLM-generated and expert-written explanations for a TTCW assessment} 

\begin{table*}[!ht]
\centering
\small
\def\arraystretch{1.35}
\begin{tabular}{|l|l|}
\hline
GPT4 & \begin{tabular}[c]{@{}l@{}}The story incorporates numerous instances of subtext that serve to further develop the characters' internal emotions \\ and relationships, as well as accentuate the atmospheric tension and undertones of longing. The subtext, which is\\ subtly weaved into surface-level interactions, actions, and observations, is eloquently crafted and does not feel \\ forced, enriching both the narrative depth and setting.\\ One key passage with evident subtext is when he is in the ocean, swimming and jumping off the rock, while she\\ observes from the towel. The surface-level interpretation suggests a casual, relaxed beach day. However, the \\ subtext is rife with allusions of emotional undercurrents in their interaction. His unawareness of her going \\ into the water parallels the emotional disconnect between them, insinuating the vast, metaphorical space that \\ separates their respective experiences and understanding of their relationship.\\ Another instance of subtext occurs when he takes a photo of her with her swimsuit top off, indicating his attraction\\ to her. But the meaning deepens when we understand that he will delete the photo before going home, implicitly\\ revealing his guilt or fear of being discovered, and again highlighting the transient, secret nature of their relationship.\\ The line, "As in certain European movies" suggests the fleeting, impermanent element of their relationship and the\\ eventual abruptness and intensity of feelings that often feature in said films. The 'European movies' serve as a \\ metaphor for the unexpected and sudden end of their encounter. \\ In these examples, the subtext does not feel forced but instead flows naturally from the characters and \\ circumstances, subtly conveying deeper meanings that heighten both the narrative tension and emotional depth.\\ \\ So Yes.\end{tabular} \\ \hline
E3   & \begin{tabular}[c]{@{}l@{}}There is rich subtext, as the main character seems continually conflicted about whether she wants to be where she is, \\ doing what she is doing. On the surface, she is carefree, riding to the beach with the guy she met, skipping the ceramics\\ and the museum, and whatever else. And yet, she is unhappy and unsatisfied, longing for a beer, imagining that if their\\ relationship continued they would only hate each other. This tension is maintained throughout the story.\end{tabular}                                                                                                                                                                                                                                                                                                               \\ \hline
E1   & \begin{tabular}[c]{@{}l@{}}This piece has an iceberg of subtext floating underneath it. The entire story is conveyed through the successful \\ integration of subtext and text. The interactions between the protagonist and the man (Did you see me jump of the \\ rock? No, she hadn't.Did he notice she had gone in the water too, that her hair was dripping? No, he hadn't.)convey\\ a profound disconnect that causes the reader to wonder why the protagonist continues to suffer the presence of this\\ man she clearly disdains and seems to view as an incompetent man-child.\end{tabular}
               \\ \hline
E7   & \begin{tabular}[c]{@{}l@{}}Yes!!!!! Again, the idea of the story was fairly simple (the inevitability of age, parting, change), but it was illustrated\\ in a way that felt inspiring re: questioning how these ideas relate and resonate throughout our own lives. It was really \\ beautiful and I was left feeling changed at the end of it :)\end{tabular}                                                                                                                                                                                                                                                            \\ \hline
\end{tabular}
\vspace{2ex}
\caption{\label{llmvsexpertexpl}LLM explanation vs expert explanation for Rhetorical Complexity}
\end{table*}

In Table~\ref{llmvsexpertexpl}, we show examples of explanations that experts wrote in conjunction with a binary TTCW assessment they made on a story, as well as the corresponding LLM-generated explanations.

\subsection{Can non-experts administer TTCW tests?}

Recruiting experts for data annotation purposes is challenging, and costly, and must consider the time constraint put on the experts. Prior work has shown the potential of crowd-sourcing (through platforms such as Amazon Mechanical Turk) and the ability of non-experts to accomplish complex tasks as a crowd \cite{kittur2013future}, when following an appropriate workflow that iterates and validates the work on individual non-experts. Some prior work has even shown the validity of crowd-based feedback for writing tasks \cite{bernstein2010soylent,nebeling2016wearwrite}. 

In this work, we chose to rely on experts for annotation, to maximize the validity of our experiments, and confirm whether experts with domain knowledge would reach satisfying agreement levels when evaluating stories with TTCW. Future work can leverage our open-sourced annotations to explore whether non-experts correlate with experts when performing TTCW evaluation, which could lead to more cost-effective TTCW evaluation.

\subsection{Prompts for TTCW} \label{allprompts}

All the instructions shown to creative writing experts and LLMs are given in the tables below.

\begin{table*}[!ht]
\centering
\small
\begin{tabular}{|l|l|}
\hline
\begin{tabular}[c]{@{}l@{}}Expert \\ Measure\end{tabular}               & Does the manipulation of time in terms of compression or stretching feel appropriate and balanced?                                                                                                                                    \\ \hline
\begin{tabular}[c]{@{}l@{}}Expanded\\ Expert\\ Measure (M)\end{tabular} & \begin{tabular}[c]{@{}l@{}}`Compression/stretching of time' in fiction writing, also known as pacing, refers to the manipulation of time in \\storytelling for dramatic effect, pacing, or other narrative purposes. Essentially, it's about controlling the perceived \\speed and rhythm at which a story unfolds.\\ \\

Compression of time refers to when events that take a long time (hours, days, weeks, or even years) are summarized \\or condensed into a brief narrative span. For example, a writer might compress several years of a character's life \\into a few paragraphs to quickly convey important changes or developments.\\ \\

On the other hand, stretching of time is when a brief moment or event is drawn out over pages or chapters. It's often \\used to create suspense, emphasize details, or delve deeper into a character's thoughts and feelings. For example, \\the few seconds it takes for a dropped glass to hit the floor might be stretched out with detailed descriptions of the\\ action, reactions, and thoughts of characters involved.\\ \\

Storytime refers to the time within the world of the story, while real-world time refers to the time it takes for the \\reader to read the story. A skilled writer can manipulate the relationship between these two to affect the pacing of \\the narrative, either speeding it up (compression) or slowing it down (stretching). This technique plays a crucial role \\in shaping the reader's experience and engagement with the story.\end{tabular} \\ \hline
\begin{tabular}[c]{@{}l@{}}Human\\ Instruction\end{tabular}             & \begin{tabular}[c]{@{}l@{}}\{\{M\}\}\\ \\ Based on the story that you just read, answer the following question.\\ \textit{\color{blue}Does the manipulation of time in terms of compression or stretching feel appropriate and balanced?}\\ -Yes \\ -No \\\\ Reasoning : \end{tabular}                                                                       \\ \hline
\begin{tabular}[c]{@{}l@{}}LLM\\ Instruction\end{tabular}               & \begin{tabular}[c]{@{}l@{}}\{\{M\}\}\\ \\ Given the story above, list out the scenes in the story in which time compression or time stretching is used, and \\argue for each whether it is successfully implemented.  Then overall, give your reasoning about the question below \\and give an answer to it between 'Yes' or 'No' only \\ \\ \textit{\color{blue} Q) Does the manipulation of time in terms of compression or stretching feel appropriate and balanced?}\end{tabular}                                                                                                                                                                                                                    \\ \hline
\end{tabular}
\vspace{2ex}
\caption{\label{prompting}TTCW Fluency1 (Narrative Pacing) }
\vspace{-5ex}
\end{table*}

% ==================================================

\begin{table*}[!ht]
\centering
\small
\begin{tabular}{|l|l|}
\hline
\begin{tabular}[c]{@{}l@{}}Expert \\ Measure\end{tabular}               & \begin{tabular}[c]{@{}l@{}}Does the story have an appropriate balance between scene and summary/exposition or it relies on one\\ of the elements heavily compared to the other?  \end{tabular}                                                                                                                                  \\ \hline
\begin{tabular}[c]{@{}l@{}}Expanded\\ Expert\\ Measure (M)\end{tabular} & \begin{tabular}[c]{@{}l@{}}'Scene' and 'summary/exposition' are two crucial elements of narrative storytelling, and balancing them \\appropriately is an important skill in fiction writing.\\ \\ 

A 'scene' is a moment in the story that is dramatized in real-time. Scenes are usually vivid and engaging, often \\featuring character interaction, dialogue, and action. They are the building blocks of the plot, and through them, \\the story unfolds.\\ \\ 

'Summary' or 'exposition', on the other hand, involves summarizing events or providing information. Instead of \\unfolding in real time, \\summaries compress time and tell the reader what happened. Exposition provides \\necessary background information, like character history, setting details, or prior events. \\ \\ 

A good writer knows when to use scenes to make the story come alive, show character development, or increase \\tension. They also know when to use summary or exposition to move the story forward, fill in background \\information, or bridge gaps between important scenes. \\ \\ 

The right balance between scene and summary/exposition can vary depending on the story, but in general, it's \\essential for maintaining a good pace, keeping the reader engaged, and delivering necessary information. \\A story with too many scenes and not enough summary might feel overwhelming or slow, while a story with \\too much exposition and not enough scenes could feel dry and unengaging.\end{tabular} \\ \hline
\begin{tabular}[c]{@{}l@{}}Human\\ Instruction\end{tabular}             & \begin{tabular}[c]{@{}l@{}}\{\{M\}\}\\ \\ Based on the story that you just read, answer the following question.\\ \textit{\color{blue} Does the story have an appropriate balance between scene and summary/exposition or it relies on one of the elements} \\\textit{\color{blue}heavily compared to the other?} \\ -Yes \\ -No \\\\ Reasoning : \end{tabular}    
\\ \hline
\begin{tabular}[c]{@{}l@{}}LLM\\ Instruction\end{tabular}               & \begin{tabular}[c]{@{}l@{}}\{\{M\}\}\\ \\ Given the story above, answer the following question. Please first explain your reasoning step by step \\and then given an answer between 'Yes' or 'No' only \\ \\ \textit{\color{blue} Does the story have an appropriate balance between scene and summary/exposition or it relies on one of the elements} \\\textit{\color{blue}heavily compared to the other?}\end{tabular}                                                                                                                                                                                                                    \\ \hline
\end{tabular}
\vspace{2ex}
\caption{\label{prompting}TTCW Fluency2 (Scene vs Exposition) }
\vspace{-5ex}
\end{table*}

% ==================================================

\begin{table*}[!ht]
\centering
\small
\begin{tabular}{|l|l|}
\hline
\begin{tabular}[c]{@{}l@{}}Expert \\ Measure\end{tabular}               & Does the story make sophisticated use of idiom or metaphor or literary allusion?                                                                                                                                     \\ \hline
\begin{tabular}[c]{@{}l@{}}Expanded\\ Expert\\ Measure (M)\end{tabular} & \begin{tabular}[c]{@{}l@{}}`Idiom' refers to phrases or expressions that have a figurative, or sometimes literal, meaning that is \\comprehensible to a particular group of people. These can be cultural, regional, or specific to a certain group or \\profession.Sophisticated use of idiom suggests that the writer is skillfully using these expressions to lend \\authenticity to character voices or to convey specific meanings in a concise way.\\\\

`Metaphor' is a figure of speech that describes an object or action in a way that isn't literally true, but helps explain\\ an idea or make a comparison. Sophisticated use of metaphor suggests the
writer could create impactful, original \\comparisons that reveal deeper insights about themes,
characters, or situations in the story.\\\\

`Literary allusion' refers to a brief and indirect reference to a person, place, thing or idea of
historical, cultural,\\ literary, or political significance. It does not describe in detail the person or thing to which it refers. A sophisticated\\ use of literary allusion implies the writer can effectively incorporate these references to enhance the depth\\ and resonance of the story. They can provide contextual richness, evoke a specific tone, or draw parallels between\\ the narrative and the work alluded to.\\\\

Overall, when a writer uses these techniques well, they add depth, interest, and nuanced \\meaning
to their work. It allows for a richer reading experience, where the literal events are \\imbued with deeper symbolic or thematic significance.\end{tabular} \\ \hline
\begin{tabular}[c]{@{}l@{}}Human\\ Instruction\end{tabular}             & \begin{tabular}[c]{@{}l@{}}\{\{M\}\}\\ \\ Based on the story that you just read, answer the following question.\\ \textit{\color{blue}Does the story make sophisticated use of idiom or metaphor or literary allusion?}\\ -Yes \\ -No \\\\ Reasoning: \end{tabular}                                                                       \\ \hline
\begin{tabular}[c]{@{}l@{}}LLM\\ Instruction\end{tabular}               & \begin{tabular}[c]{@{}l@{}}\{\{M\}\}\\ \\ Given the story above, please list out all the metaphors, idioms and literary allusions, and for each decide \\whether it is successful vs it feels forced or too easy.  Then overall, give your reasoning about the question \\below and give an answer to it between 'Yes' or 'No' only\\ \\ \textit{\color{blue} Q) Does the story make sophisticated use of idiom or metaphor or literary allusion?}\end{tabular}                                                                                                                                                                                                                    \\ \hline
\end{tabular}
\vspace{2ex}
\caption{\label{prompting}TTCW Fluency3 (Language Proficiency \& Literary Devices) }
\vspace{-5ex}
\end{table*}

% ==================================================

\begin{table*}[!ht]
\centering
\small
\begin{tabular}{|l|l|}
\hline
\begin{tabular}[c]{@{}l@{}}Expert \\ Measure\end{tabular}               & Does the end of the story feel natural and earned, as opposed to arbitrary or abrupt?                                                                                                                                    \\ \hline
\begin{tabular}[c]{@{}l@{}}Expanded\\ Expert\\ Measure (M)\end{tabular} & \begin{tabular}[c]{@{}l@{}}If the writer ends the piece simply because they are 'tired of writing', the conclusion might feel abrupt, disjointed, \\or unfulfilling to the reader. It suggests a rushed ending, where plot threads might be left unresolved and character \\arcs incomplete.\\ \\ 

Conversely, if the writer concludes because they've reached `the moment the entire piece has been leading readers \\towards', it implies a well-considered and purposeful ending. The events, character development, and themes \\throughout the story have built towards this climactic moment, providing a satisfying resolution to the reader.\\ \\ 

A strong ending offers a sense of closure, ties up the central conflicts or questions of the story, and generally \\leaves the reader feeling that the narrative journey was worthwhile and complete.\end{tabular} \\ \hline
\begin{tabular}[c]{@{}l@{}}Human\\ Instruction\end{tabular}             & \begin{tabular}[c]{@{}l@{}}\{\{M\}\}\\ \\ Based on the story that you just read, answer the following question.\\ \textit{\color{blue}Does the end of the story feel natural and earned, as opposed to arbitrary or abrupt?}\\ -Yes \\ -No \\\\ Reasoning : \end{tabular}                                                                       \\ \hline
\begin{tabular}[c]{@{}l@{}}LLM\\ Instruction\end{tabular}               & \begin{tabular}[c]{@{}l@{}}\{\{M\}\}\\ \\ Given the story above, answer the following question. Please first explain your reasoning step by step \\ and then given an answer between 'Yes' or 'No' only\\ \\ \textit{\color{blue} Q) Does the end of the story feel natural and earned, as opposed to arbitrary or abrupt?}\end{tabular}                                                                                                                                                                                                                    \\ \hline
\end{tabular}
\vspace{2ex}
\caption{\label{prompting}TTCW Fluency4 (Narrative Ending) }
\vspace{-5ex}
\end{table*}

% ==================================================

\begin{table*}[!ht]
\centering
\small
\begin{tabular}{|l|l|}
\hline
\begin{tabular}[c]{@{}l@{}}Expert \\ Measure\end{tabular}               & Do the different elements of the story work together to form a unified, engaging, and satisfying whole?                                                                                                                                     \\ \hline
\begin{tabular}[c]{@{}l@{}}Expanded\\ Expert\\ Measure (M)\end{tabular} & \begin{tabular}[c]{@{}l@{}}A well-crafted story usually follows a logical path, where the events in the beginning set up the middle, which then\\ logically leads to the end. Every scene, character action, and piece of dialogue should serve the story and propel it \\forward. Well-written stories have an underlying the unity that binds the elements together. The themes, plotlines, \\character arcs, and other elements of the story interweave to create a harmonious whole. A story with 'disorder'\\ might feel disjointed, with characters, scenes, etc that don't connect or contribute to the overall narrative.\end{tabular} \\ \hline
\begin{tabular}[c]{@{}l@{}}Human\\ Instruction\end{tabular}             & \begin{tabular}[c]{@{}l@{}}\{\{M\}\}\\ \\ Based on the story that you just read, answer the following question.\\ \textit{\color{blue}Do the different elements of the story work together to form a unified, engaging, and satisfying whole?}\\ -Yes \\ -No \\\\ Reasoning : \end{tabular}                                                                       \\ \hline
\begin{tabular}[c]{@{}l@{}}LLM\\ Instruction\end{tabular}               & \begin{tabular}[c]{@{}l@{}}\{\{M\}\}\\ \\ Given the story above, answer the following question. Please first explain your reasoning step by step and then \\give an answer between 'Yes' or 'No' only\\ \\ \textit{\color{blue} Q) Do the different elements of the story work together to form a unified, engaging, and satisfying whole?}\end{tabular}                                                                                                                                                                                                                                 \\ \hline
\end{tabular}
\vspace{2ex}
\caption{\label{prompting}TTCW Fluency5 (Understandability \& Coherence) }
\vspace{-5ex}
\end{table*}

% ==================================================

\begin{table*}[!ht]
\centering
\small
\begin{tabular}{|l|l|}
\hline
\begin{tabular}[c]{@{}l@{}}Expert \\ Measure\end{tabular}               & \begin{tabular}[c]{@{}l@{}}Does the story provide diverse perspectives, and if there are unlikeable characters, are their perspectives \\presented convincingly and accurately? \end{tabular}                                                                                                                                     \\ \hline
\begin{tabular}[c]{@{}l@{}}Expanded\\ Expert\\ Measure (M)\end{tabular} & \begin{tabular}[c]{@{}l@{}}A good writer can convincingly and accurately depict a wide range of character viewpoints, including those of\\ characters who may be morally ambiguous, difficult, or otherwise unappealing.\\ \\ 

This can involve diving into the mindset of characters who may act or think in ways that the reader, or even \\the writer, finds objectionable or repugnant. It involves understanding their motivations, their beliefs, and the \\reasons behind their actions, and then conveying these elements in a way that is believable and consistent.\\ \\ 

The purpose of doing so is not to justify or endorse these perspectives, but rather to create complex, three-\\dimensional characters who contribute to the richness and depth of the story. This can also serve to \\challenge the reader, provoke thought, and provide insights into different aspects of the human experience.\end{tabular} \\ \hline
\begin{tabular}[c]{@{}l@{}}Human\\ Instruction\end{tabular}             & \begin{tabular}[c]{@{}l@{}}\{\{M\}\}\\ \\ Based on the story that you just read, answer the following question.\\ \textit{\color{blue}Does the story provide diverse perspectives, and if there are unlikeable characters, are their perspectives presented} \\ \textit{\color{blue}convincingly and accurately?}\\ -Yes \\ -No \\\\ Reasoning : \end{tabular}                                                                       \\ \hline
\begin{tabular}[c]{@{}l@{}}LLM\\ Instruction\end{tabular}               & \begin{tabular}[c]{@{}l@{}}\{\{M\}\}\\ \\ Given the story above, answer the following question. Please first explain your reasoning step by step and then \\give an answer between 'Yes' or 'No' only\\ \\ \textit{\color{blue} Q) Does the story provide diverse perspectives, and if there are unlikeable characters, are their perspectives presented}\\\textit{\color{blue} convincingly and accurately?}\end{tabular}                                                                                                                                                                                                                                 \\ \hline
\end{tabular}
\vspace{2ex}
\caption{\label{prompting}TTCW Flexibility1 (Perspective \& Voice Flexibility) }
\vspace{-5ex}
\end{table*}

% ==================================================

\begin{table*}[!ht]
\centering
\small
\begin{tabular}{|l|l|}
\hline
\begin{tabular}[c]{@{}l@{}}Expert \\ Measure\end{tabular}               & \begin{tabular}[c]{@{}l@{}}Does the story achieve a good balance between interiority and exteriority, in a way that feels \\emotionally flexible? \end{tabular}                                                                                                                                     \\ \hline
\begin{tabular}[c]{@{}l@{}}Expanded\\ Expert\\ Measure (M)\end{tabular} & \begin{tabular}[c]{@{}l@{}}`Emotional flexibility' is asking whether the piece of writing effectively balances action and introspection, and \\if it portrays a broad and realistic spectrum of emotions.\\ \\

`Exteriority' refers to the observable actions, behaviors, or dialogue of a character, and the physical or visible \\aspects of the setting, plot, and conflicts.\\ \\

`Interiority', on the other hand, pertains to the inner life of a character — their thoughts, feelings, memories, \\and subjective experiences.\\ \\

A balance between these two aspects is crucial in creating well-rounded characters and compelling narratives. \\If a piece is too heavy on exteriority, it may feel shallow or lack emotional depth. If it leans too much on\\ interiority, it could become overly introspective and potentially lose the momentum of the plot.
\end{tabular} \\ \hline
\begin{tabular}[c]{@{}l@{}}Human\\ Instruction\end{tabular}             & \begin{tabular}[c]{@{}l@{}}\{\{M\}\}\\ \\ Based on the story that you just read, answer the following question.\\ \textit{\color{blue}Does the story achieve a good balance between interiority and exteriority, in a way that feels emotionally flexible?}\\ -Yes \\ -No \\\\ Reasoning : \end{tabular}                                                                       \\ \hline
\begin{tabular}[c]{@{}l@{}}LLM\\ Instruction\end{tabular}               & \begin{tabular}[c]{@{}l@{}}\{\{M\}\}\\ \\ Given the story above, answer the following question. Please first explain your reasoning step by step and \\then give an answer between 'Yes' or 'No' only\\ \\ \textit{\color{blue}Q) Does the story achieve a good balance between interiority and exteriority, in a way that feels} \\\textit{\color{blue}emotionally flexible?}\end{tabular}                                                                                                                                                                                                                                 \\ \hline
\end{tabular}
\vspace{2ex}
\caption{\label{prompting}TTCW Flexibility2 (Emotional Flexibility) }
\vspace{-5ex}
\end{table*}

% ==================================================

\begin{table*}[!ht]
\centering
\small
\begin{tabular}{|l|l|}
\hline
\begin{tabular}[c]{@{}l@{}}Expert \\ Measure\end{tabular}               & \begin{tabular}[c]{@{}l@{}}Does the story contain turns that are both surprising and appropriate? \end{tabular}                                                                                                                                     \\ \hline
\begin{tabular}[c]{@{}l@{}}Expanded\\ Expert\\ Measure (M)\end{tabular} & \begin{tabular}[c]{@{}l@{}}`Surprising': This refers to the element of unpredictability in a narrative. A good story often has plot twists, \\character developments, or thematic revelations that surprise the reader, subverting their expectations in a \\thrilling way.It's about keeping readers engaged and curious, never fully knowing what's going to happen next.\\ \\ 

`Appropriate': Despite the surprises and twists, the turns in the story must also make sense within the established \\context of the story's universe, its characters, and its themes. This means that even though an event might be \\surprising, it should feel appropriate or fitting in hindsight. It shouldn't feel like the writer has broken the rules \\they've set up, or made a character behave inconsistently without reason, simply for the sake of shock value.\\ \\ 

So when someone wonders if a writer can make turns that are 'both surprising and appropriate', they're asking \\if the writer can strike this balance between unexpectedness and coherence, keeping the reader on their toes \\while maintaining a believable, satisfying narrative. \end{tabular} \\ \hline
\begin{tabular}[c]{@{}l@{}}Human\\ Instruction\end{tabular}             & \begin{tabular}[c]{@{}l@{}}\{\{M\}\}\\ \\ Based on the story that you just read, answer the following question.\\ \textit{\color{blue}Does the story contain turns that are both surprising and appropriate?}\\ -Yes \\ -No \\\\ Reasoning: \end{tabular}                                                                       \\ \hline
\begin{tabular}[c]{@{}l@{}}LLM\\ Instruction\end{tabular}               & \begin{tabular}[c]{@{}l@{}}\{\{M\}\}\\ \\ Given the story above, list each element in the story that is intended to be surprising. For each, decide whether the\\ surprising element remains appropriate with respect to the entire story. Then overall, give your reasoning \\about the question below and give an answer to it between 'Yes' or 'No' only\\ \\ \textit{\color{blue} Q) Does the story contain turns that are both surprising and appropriate?}\end{tabular}                                                                                                                                                                                                                                 \\ \hline
\end{tabular}
\vspace{2ex}
\caption{\label{prompting}TTCW Flexibility3 (Structural Flexibility) }
\vspace{-5ex}
\end{table*}

% ==================================================

\begin{table*}[!ht]
\centering
\small
\begin{tabular}{|l|l|}
\hline
\begin{tabular}[c]{@{}l@{}}Expert \\ Measure\end{tabular}               & \begin{tabular}[c]{@{}l@{}}Will an average reader of this story obtain a unique and original idea from reading it? \end{tabular}                                                                                                                                     \\ \hline
\begin{tabular}[c]{@{}l@{}}Expanded\\ Expert\\ Measure (M)\end{tabular} & \begin{tabular}[c]{@{}l@{}}If a story is good, the reader gains new insights, perspectives, or knowledge from it. This doesn't necessarily\\ mean factual information but could relate to a deeper understanding of human nature, cultural insights,\\ unique viewpoints, or even the exploration of new ideas and themes. Essentially, it's about what\\ the reader takes away from the story beyond just the plot.\\ \\ 

A good story has lasting impacts on its readers and the society. It is meant to entertain, inform, provoke \\thought, challenge beliefs, provide comfort, or raise awareness on specific issues.
 \end{tabular} \\ \hline
\begin{tabular}[c]{@{}l@{}}Human\\ Instruction\end{tabular}             & \begin{tabular}[c]{@{}l@{}}\{\{M\}\}\\ \\ Based on the story that you just read, answer the following question.\\ \textit{\color{blue}Will an average reader of this story obtain a unique and original idea from reading it?}\\ -Yes \\ -No \\\\ Reasoning : \end{tabular}                                                                       \\ \hline
\begin{tabular}[c]{@{}l@{}}LLM\\ Instruction\end{tabular}               & \begin{tabular}[c]{@{}l@{}}\{\{M\}\}\\ \\ Given the story above, list out elements that are unique takeaways of this story for the reader. Then overall, \\give your reasoning about the question below and give an answer to it between 'Yes' or 'No' only\\ \\ \textit{\color{blue} Q) Will an average reader of this story obtain a unique and original idea from reading it?}\end{tabular}                                                                                                                                                                                                                                 \\ \hline
\end{tabular}
\vspace{2ex}
\caption{\label{prompting}TTCW Originality1 (Originality in Theme and Content) }
\vspace{-3ex}
\end{table*}

% ==================================================

\begin{table*}[!ht]
\centering
\small
\begin{tabular}{|l|l|}
\hline
\begin{tabular}[c]{@{}l@{}}Expert \\ Measure\end{tabular}               & \begin{tabular}[c]{@{}l@{}}Is the story an original piece of writing without any cliches?\end{tabular}                                                                                                                                     \\ \hline
\begin{tabular}[c]{@{}l@{}}Expanded\\ Expert\\ Measure (M)\end{tabular} & \begin{tabular}[c]{@{}l@{}}A cliche is an idea, expression, character, or plot that has been overused to the point of losing its original \\meaning or impact. They often become predictable and uninteresting for the reader. Originality suggests\\ that the piece isn't cliche.

 \end{tabular} \\ \hline
\begin{tabular}[c]{@{}l@{}}Human\\ Instruction\end{tabular}             & \begin{tabular}[c]{@{}l@{}}\{\{M\}\}\\ \\ Based on the story that you just read, answer the following question.\\ \textit{\color{blue}Is the story an original piece of writing without any cliches?}\\ -Yes \\ -No \\\\ Reasoning: \end{tabular}                                                                       \\ \hline
\begin{tabular}[c]{@{}l@{}}LLM\\ Instruction\end{tabular}               & \begin{tabular}[c]{@{}l@{}}\{\{M\}\}\\ \\ Given the story above, are there any cliches in the story? If so, list out all the elements in this story that \\are cliche. Then overall, give your reasoning if the piece is negatively impacted by the cliches and give \\an answer to the question below between 'Yes' or 'No' only\\ \\ \textit{\color{blue} Q) Is the story an original piece of writing without any cliches?}\end{tabular}                                                                                                                                                                                                                                 \\ \hline
\end{tabular}
\vspace{2ex}
\caption{\label{prompting}TTCW Originality2 (Originality in Thought) }
\vspace{-5ex}
\end{table*}

% ==================================================

\begin{table*}[!ht]
\centering
\small
\begin{tabular}{|l|l|}
\hline
\begin{tabular}[c]{@{}l@{}}Expert \\ Measure\end{tabular}               & \begin{tabular}[c]{@{}l@{}}Does the story show originality in its form?\end{tabular}                                                                                                                                     \\ \hline
\begin{tabular}[c]{@{}l@{}}Expanded\\ Expert\\ Measure (M)\end{tabular} & \begin{tabular}[c]{@{}l@{}}When someone says that a piece of fiction 'shows an innovative use of form/structure', they're referring to\\ how the writer has chosen to shape and organize the story in an unusual, original, or inventive way. This could \\involve a variety of different elements, including:\\ \\ 

Narrative Structure: This could include unconventional timelines (e.g. a non-linear story, a story told in reverse)\\, multiple perspectives or narrators, or unusual narrative voices (e.g. a story told from the perspective of an \\inanimate object).\\ \\ 

Format: This could be something as simple as using unconventional punctuation or capitalization, or as complex \\as telling a story through a series of letters, diary entries, newspaper clippings, or other documents. In recent years,\\ some authors have even experimented with using social media posts or text messages as a form of narrative structure.\\ \\ 

Genre Hybridity: Combining elements from different genres or sub-genres in unexpected ways can also be seen\\ as an innovative use of form such as Horror-Mystery or Comic Fantasy.\\ \\ 

Plot Structure: Deviating from traditional plot structures such as three-act structure, or following them in unexpected\\ ways.For example, telling a story without a clear resolution, incorporating multiple climaxes or using revelation as a \\device where a surprising, and often shocking, development occurs that was previously kept hidden from the \\characters and/or the audience. It's typically designed to provide new context for interpreting what has previously \\occurred in the story. \\ \\ 

Language and Style: Innovative use of form can also come in the form of unique use of language, style, or \\even typography, such as concrete poetry or writing that visually represents its subject matter on the page.\\ \\ 

The goal of this innovation is often to provide a fresh reader experience, challenge conventional reading\\ expectations, or to create a deeper or more complex exploration of the story's themes.

 \end{tabular} \\ \hline
\begin{tabular}[c]{@{}l@{}}Human\\ Instruction\end{tabular}             & \begin{tabular}[c]{@{}l@{}}\{\{M\}\}\\ \\ Based on the story that you just read, answer the following question.\\ \textit{\color{blue}Does the story show originality in its form?}\\ -Yes \\ -No \\\\ Reasoning: \end{tabular}                                                                       \\ \hline
\begin{tabular}[c]{@{}l@{}}LLM\\ Instruction\end{tabular}               & \begin{tabular}[c]{@{}l@{}}\{\{M\}\}\\ \\ Given the story and the devices mentioned above, list each device used with a short explanation of whether it is \\successful or not. Then overall, give your reasoning about the question below and give an answer to it\\ between 'Yes' or 'No' only\\ \\ \textit{\color{blue} Q) Does the story show originality in its form?}\end{tabular}                                                                                                                                                                                                                                 \\ \hline
\end{tabular}
\vspace{2ex}
\caption{\label{prompting}TTCW Originality3 (Originality in Form) }
\vspace{-5ex}
\end{table*}

% ==================================================

\begin{table*}[!ht]
\centering
\small
\begin{tabular}{|l|l|}
\hline
\begin{tabular}[c]{@{}l@{}}Expert \\ Measure\end{tabular}               & \begin{tabular}[c]{@{}l@{}}Does each character in the story feel developed at the appropriate complexity level, ensuring that no character \\feels like they are present simply to satisfy a plot requirement?\end{tabular}                                                                                                                                     \\ \hline
\begin{tabular}[c]{@{}l@{}}Expanded\\ Expert\\ Measure (M)\end{tabular} & \begin{tabular}[c]{@{}l@{}} A `flat character' is typically a minor character who is not thoroughly developed or who does not undergo \\significant change or growth throughout the story. They often embody or represent a single trait or idea, \\and they're used to advance the plot or highlight certain qualities in other characters.\\ \\ 

A `complex character', also known as a round character, has depth in feelings and passions, has a variety \\of traits of a real human being, and evolves over time. They have their strengths, weaknesses, \\and they learn from their experiences. They tend to be more engaging to the reader, as they mirror \\the complexity of real people.\\ \\ 

In good stories, authors take a character who initially appears to be one-dimensional or stereotypical (flat) and \\add depth to them. This could be done by revealing more about their backstory, introducing unexpected traits \\or motivations, or having them grow and change in response to the events of the story. \\This transformation from a flat to a complex character can make the narrative more engaging and believable.
 \end{tabular} \\ \hline
\begin{tabular}[c]{@{}l@{}}Human\\ Instruction\end{tabular}             & \begin{tabular}[c]{@{}l@{}}\{\{M\}\}\\ \\ Based on the story that you just read, answer the following question.\\  \textit{\color{blue} Q) Does each character in the story feel developed at the appropriate complexity level, ensuring that no character} \\ \textit{\color{blue}feels like they are present simply to satisfy a plot requirement?}\\ -Yes \\ -No \\\\ Reasoning: \end{tabular}                                                                       \\ \hline
\begin{tabular}[c]{@{}l@{}}LLM\\ Instruction\end{tabular}               & \begin{tabular}[c]{@{}l@{}}\{\{M\}\}\\ \\ Given the story above, list each character and the level of development. Then overall, give your reasoning \\about the question below and give an answer to it between 'Yes' or 'No' only\\ \\ 
 \textit{\color{blue} Q) Does each character in the story feel developed at the appropriate complexity level, ensuring that no character} \\ \textit{\color{blue}feels like they are present simply to satisfy a plot requirement?}\end{tabular}                                                                                                                                                                                                                                 \\ \hline
\end{tabular}
\vspace{2ex}
\caption{\label{prompting}TTCW Elaboration2 (Character Development) }
\vspace{-5ex}
\end{table*}

% ==================================================

\begin{table*}[!ht]
\centering
\small
\begin{tabular}{|l|l|}
\hline
\begin{tabular}[c]{@{}l@{}}Expert \\ Measure\end{tabular}               & \begin{tabular}[c]{@{}l@{}}Are there passages in the story that involve subtext and when there is subtext, does it enrich the story's setting \\or does it feel forced?\end{tabular}                                                                                                                                     \\ \hline
\begin{tabular}[c]{@{}l@{}}Expanded\\ Expert\\ Measure (M)\end{tabular} & \begin{tabular}[c]{@{}l@{}} `Surface' level: This is the most apparent and straightforward level of a story. It includes the visible actions, \\explicit dialogue, and clear descriptions. This is what literally happens in the plot: the characters' actions, events, \\and the apparent consequences.\\ \\ 

`Subtext' level: This is the underlying or implicit meaning that isn't directly stated but can be inferred from \\the characters'  actions, dialogue, and other elements of the story. Subtext often reveals deeper truths about \\characters, themes, or the overall message of the piece. It could be a hidden motive, an unstated\\ emotion, a cultural commentary, or a symbolic meaning.\\ \\ 

For example, in a conversation between two characters, the surface text might be polite and cordial, but the \\subtext \\discerned from the characters' nonverbal cues, previous interactions, or the context of their conversation\\ — could suggest tension or hostility.\\ \\ 

Effective fiction often operates on both levels. The surface text keeps the reader engaged with the plot and \\characters, while the subtext provides depth, complexity, and additional layers of interpretation, \\contributing to a richer and more rewarding reading experience.
 \end{tabular} \\ \hline
\begin{tabular}[c]{@{}l@{}}Human\\ Instruction\end{tabular}             & \begin{tabular}[c]{@{}l@{}}\{\{M\}\}\\ \\ Based on the story that you just read, answer the following question.\\  \textit{\color{blue} Q) Are there passages in the story that involve subtext and when there is subtext, does it enrich the story's setting} \\ \textit{\color{blue} or does it feel forced?}\\ -Yes \\ -No \\\\ Reasoning: \end{tabular}                                                                       \\ \hline
\begin{tabular}[c]{@{}l@{}}LLM\\ Instruction\end{tabular}               & \begin{tabular}[c]{@{}l@{}}\{\{M\}\}\\ \\ Given the story above, answer the following question. Please first explain your reasoning step by step \\and then give an answer between 'Yes' or 'No' only\\ \\ 
 \textit{\color{blue} Q)Are there passages in the story that involve subtext and when there is subtext, does it enrich the story's setting} \\ \textit{\color{blue} or does it feel forced?}\end{tabular}                                                                                                                                                                                                                                 \\ \hline
\end{tabular}
\vspace{2ex}
\caption{\label{prompting}TTCW Elaboration3 (Rhetorical Complexity) }
\vspace{-5ex}
\end{table*}

% ==================================================

%TC:endignore
\end{document}
\endinput

%%
%% End of file `sample-authordraft.tex'.